\documentclass[10pt,twocolumn,letterpaper]{article}

\usepackage{ijcb}
\usepackage{times}
\usepackage{epsfig}
\usepackage{graphicx}
\usepackage{amsmath}
\usepackage{amssymb}

\usepackage{ijcb}
\usepackage{subcaption}
\usepackage{enumitem}
\setlist{nosep} 

\usepackage[pagebackref=true,breaklinks=true,colorlinks,bookmarks=false]{hyperref}

\ijcbfinalcopy 

\ifijcbfinal\fi
\begin{document}

\title{Haven't I Seen You Before?\\Assessing Identity Leakage in Synthetic Irises}

\author{Patrick Tinsley\\
{\tt\small ptinsley@nd.edu}
~\\
\and
Adam Czajka\\
{\tt\small aczajka@nd.edu}\\
University of Notre Dame, IN 46556, USA\\
\and
Patrick Flynn\\
{\tt\small flynn@nd.edu}\\
~
}

\newcommand{\teaser}{
{
   \begin{center}
        \vskip5mm
        \centering
            \begin{minipage}{\textwidth}
                \centering
                \includegraphics[width=.8\textwidth]{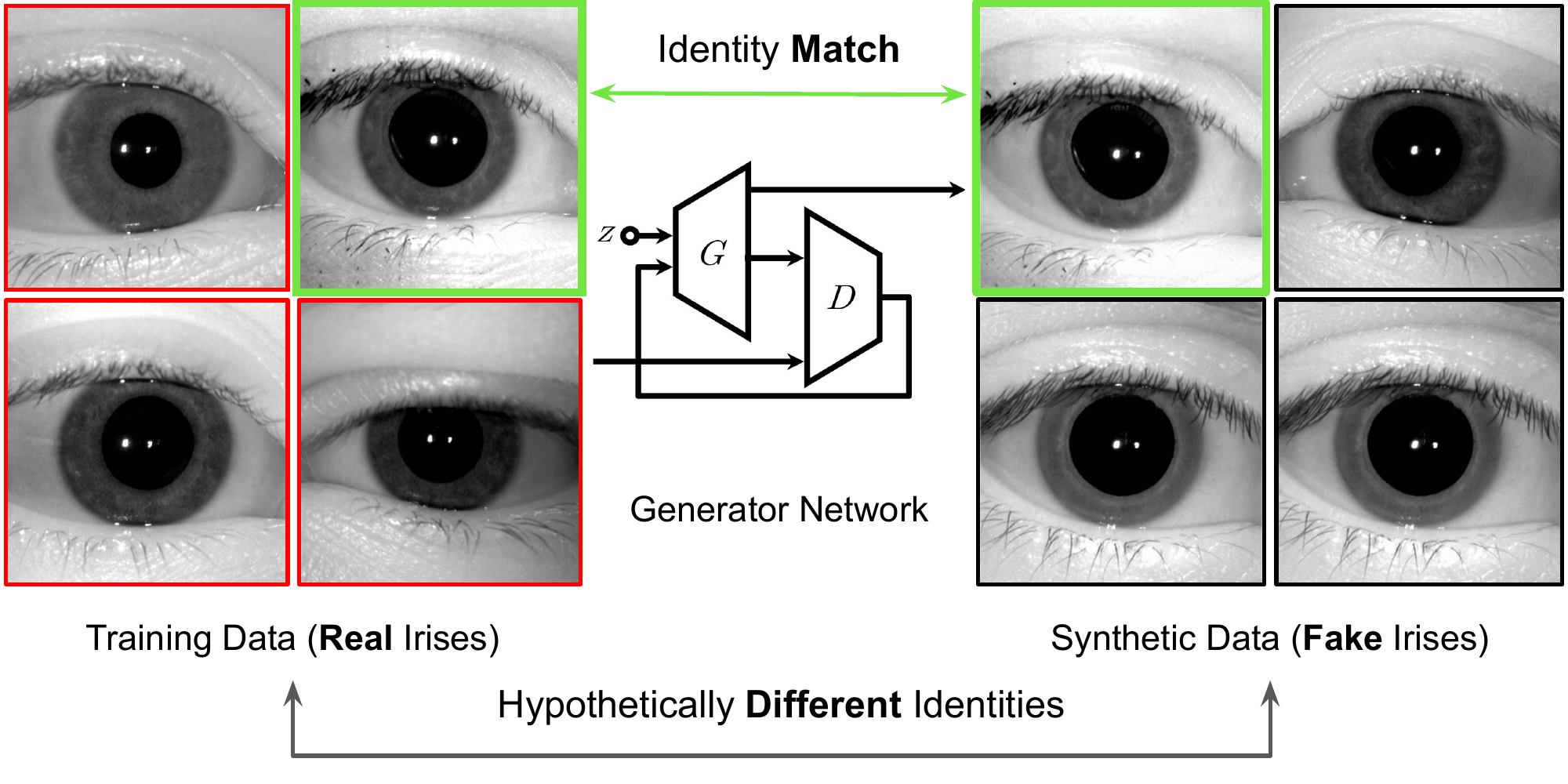}
            \end{minipage}%
            \captionof{figure}{Demonstration of identity leakage in iris synthesis: one of the real training samples (left of green arrow) is re-constructed almost perfectly (right of arrow) by the trained generative model.}
            \label{fig:teaser}
    \end{center}
}
}

\maketitle
\thispagestyle{empty}


\begin{abstract}

Generative Adversarial Networks (GANs) have proven to be a preferred method of synthesizing fake images of objects, such as faces, animals, and automobiles. It is not surprising these models can also generate ISO-compliant, yet synthetic iris images, which can be used to augment training data for iris matchers and liveness detectors. In this work, we trained one of the most recent GAN models (StyleGAN3~\cite{karras2021alias}) to generate fake iris images with two primary goals: (i) to understand the GAN's ability to produce ``never-before-seen'' irises, and (ii) to investigate the phenomenon of identity leakage as a function of the GAN's training time. Previous work has shown that personal biometric data can inadvertently flow from training data into synthetic samples, raising a privacy concern for subjects who accidentally appear in the training dataset. This paper presents analysis for three different iris matchers at varying points in the GAN training process to diagnose where and when authentic training samples are in jeopardy of leaking through the generative process. Our results show that while \textit{most} synthetic samples do not show signs of identity leakage, a handful of generated samples match authentic (training) samples nearly perfectly, with consensus across all matchers. In order to prioritize privacy, security, and trust in the machine learning model development process, the research community must strike a delicate balance between the benefits of using synthetic data and the corresponding threats against privacy from potential identity leakage.

\end{abstract}


\begin{figure}[t]
    \centering
    \includegraphics[width=\columnwidth]{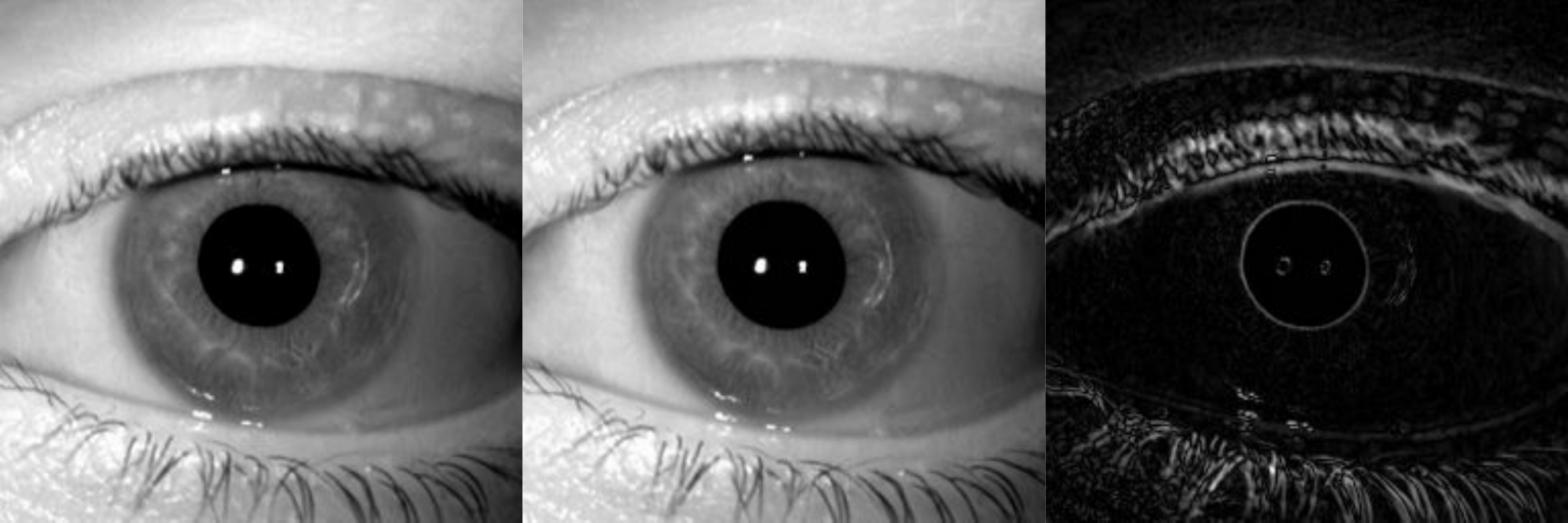}
    \caption{Training sample (left), synthetic sample (middle), and the marginal difference between the two (right), showing an instance of near-perfect identity leakage.}
    \label{fig:diff}
\end{figure}

\section{Introduction}

Ever since Goodfellow \etal introduced the generative adversarial network (GAN) framework~\cite{goodfellow2014generative}, novel image synthesis has received massive amounts of attention in the computer vision and biometric communities. Over the years, improvements have steadily been made towards producing ``fake'' images that can pass as real. Beyond the entertainment value associated with synthesizing objects (faces, animals, cityscapes, etc.) that ``do not exist,'' \cite{thisxdoesnotexist} there is also a practical demand for novel image synthesis for training both presentation attack detection (PAD) models and biometric recognition models.

Though there are many GAN architectures, one architecture that has seen sustained popularity over the years has been StyleGAN~\cite{karras2019style}. For all versions to date, out-of-the-box StyleGAN comes with a pre-trained network to generate a virtually infinite number of face images of non-existent people. One in possession of large-enough sets of training examples (in the order of tens of thousands) can train StyleGAN to generate other biometric traits of ``non-existing'' subjects, for instance, irises.

A well-known deficiency of machine learning training processes is the tendency to overfit training data. In the case of generative models, model overfitting can manifest as \textit{identity leakage}, where information about real training samples inadvertently ``leaks'' into the fake, synthetic samples that are supposed to be unrelated to real exemplars used in training. Figures \ref{fig:teaser} and \ref{fig:diff} highlight one such instance of this leakage, wherein a synthetic sample matches (nearly perfectly) to a real sample seen during training. If overfitting does occur and identity-salient information can spill into the synthesis process, there exists a privacy concern for subjects who appear in the training data. As the machine learning community continues to work toward trustworthy AI, it must strike a fine balance between increasing model performance and preserving the privacy of training subjects. The models must not remember ``too much'' \cite{too_much}.

In this work, we train NVIDIA's StyleGAN3 architecture on a dataset of $\sim$80.000 iris images, and investigate how far (from the biometric matching point of view) the generated samples fall when compared to real samples used to train this model. Three iris recognition approaches are employed to assess whether or not outputs of the trained model unintentionally divulge identity information during the synthesis process: HDBSIF~\cite{czajka2019domain}, USITv3~\cite{USIT3}, and VeriEye, a commercial iris matcher from Neurotechnology~\cite{verieye}. {\bf Our experiments show that:} 
\begin{itemize}
    \item[(a)] our trained model is capable of generating realistically-looking and ISO-compliant \cite{ISO_19794_6_2011} iris images that are correctly processed by the selected matchers;
    \item[(b)] while \textit{most} of the synthetically-generated samples do not match with training samples, the model spontaneously generates iris images that are considered a match with those in the training set by all matchers (= identity leakage);
    \item[(c)] rather than global shifts in impostor distributions (fake identities vs real identities), the identity leakage materializes in a form of long tails of these impostor distributions.
\end{itemize}

In this paper we additionally explore the dynamics of this phenomenon, analyzing the matching results at different stages of training. The source codes (forked StyleGAN3 GitHub repository with installation instructions), and the trained StyleGAN3 model generating synthetic iris images, are offered along with this paper for reproducibility purposes, iris data enhancement, and to facilitate further exploration of this topic by other colleagues.

\section{Related Work}
\label{Related Work}

\subsection{Biometric Image Synthesis}

Similar to synthetic face generation, the process of generating novel irises carries practical importance in the field of biometric re-identification. In the case of face synthesis, there are several popular open-source architectures available to generate photo-realistic faces, such as ProGAN~\cite{karras2017progressive}, StyleGAN~\cite{karras2019style}, and StarGAN~\cite{choi2018stargan}. 

For the iris domain, there are fewer existing works that attempt to construct texturally-believable ``fake'' iris samples. In their \textit{Iris-GAN} work~\cite{minaee2018iris}, Minaee \etal train a convolutional DC-GAN, introduced by Radford \etal in~\cite{radford2015unsupervised}, on two iris datasets: CASIAv3 and IIIT-Delhi. Similar to this work, the authors plot evolving iris images as a function of training time (epoch), with the final model eventually achieving an Fr\'{e}chet Inception Distance (FID) consistent with real images. 

In a similar work~\cite{zou2018generation}, Zou \etal expand on Zhu's CycleGAN framework~\cite{zhu2017unpaired} to propose 4DCycle-GAN. However, the primary goal of this work is to generate spoofed, fake iris images wearing textured
contact lenses for better PAD training. Their results show that the inclusion of their synthetic samples in the training data increased performance in spoof detection across several iris PAD models.

There are also methods for iris synthesis that are not GAN-based. In their work~\cite{cui2004iris}, Cui \etal augment coarse PCA-generated irises with finer features from a super-resolution step to generate realistic-looking iris images. Wei \etal~\cite{wei2008synthesis} suggest using iris patches and patch-based sampling to mix and match iris prototypes (coarse representations) with finer intra-class textures.

Multi-resolution approaches~\cite{samavati2007local, wecker2010multiresolution} decompose real iris images into lower resolution sub-components, which are then re-compiled randomly to synthesize realistically-looking novel samples. Makthal \etal employ a deterministic (and hence computationally inexpensive) approach to large scale iris generation involving a learned probability distribution from Markov Random Fields~\cite{makthal2005synthesis}.

Finally, Venugopalan \etal introduce an iris synthesis procedure that generates spoofed iris images given an iris code~\cite{venugopalan2011generate}.

\subsection{Identity Leakage}

The concept of identity leakage in GANs inherently suggests that models are capable of memorizing samples (or parts of samples) seen in the training process, and subsequently recreating them in the synthesis process. Though the term often relates to classification models, this ``overfitting'' is also an artifact in generative models, and has seen increased scrutiny in recent literature, both theoretically and practically~\cite{yeom2018privacy, liu2018generative, nagarajan2018theoretical}.

Feng \etal~\cite{feng2021gans} explore how dataset size relates to this exact phenomenon of overfitting/identity leakage under the verbiage of ``GAN replication.'' The authors find that, with both BigGAN~\cite{brock2018large} and StyleGAN2~\cite{karras2020analyzing}, the number of images in the training dataset has a direct correlation with the quality and diversity of synthesized images.

Webster \etal and Song \etal explore this undesirable feature of models to ``remember too much''~\cite{webster2019detecting, song2017machine}. The former suggests latent recovery as a way to detect overfitting/leakage by attempting to recreate target images from train and test splits. They find that leakage is evident in hybrid loss-based GANs and non-GAN-based methods. The latter work offers results from several targeted adversarial methods on overfit white- and black-box models for face recognition and image classification.

GANs also display evidence of identity leakage in the case of incremental or continual learning, as suggested in \cite{cong2020gan}, where Cong \etal intentionally imbue GANs with lifelong memories, modulating ``style'' sequentially atop an already well-behaved GAN architecture.

Related to the concept of identity leakage is the field of inverse biometrics, which focuses on attempts to invert biometric templates into the corresponding samples from which they came. The reconstruction of source identity from associated with the generated template obviously raises privacy concerns, and the current state of the art is documented by Gomez \etal in~\cite{gomez2020reversing}. Furthermore, the call and answer for trustworthy AI models supports development of non-invertible GANs, especially in the privacy-centric field of biometrics, as described by Jain \etal~\cite{jain2021biometrics}.

Identity leakage has been studied for StyleGAN2~\cite{karras2020analyzing} by Tinsley \etal~\cite{tinsley2021face}, who conduct identity leakage experiments in the context of face synthesis and recognition, claiming that leakage may occur depending on which face matching model is used to compare real and generated samples. Compared to Tinsley's work, this papers differs primarily in terms of biometric modality (face $\rightarrow$ iris). Furthermore, this work extends~\cite{tinsley2021face} to explore identity leakage as a function of training time. As a longitudinal study across training time, these experiments can be used to suggest potential early stopping points in future training runs to avoid generator overfitting.

Concluding this section, the literature suggests that the identity leakage in approaches synthesizing iris images has not yet been explored. Hence, this paper -- by delivering somewhat provocative results -- may open an interesting research area related to trustworthy application of iris generative models.

\begin{figure}[t]
    \centering
    \includegraphics[width=\columnwidth]{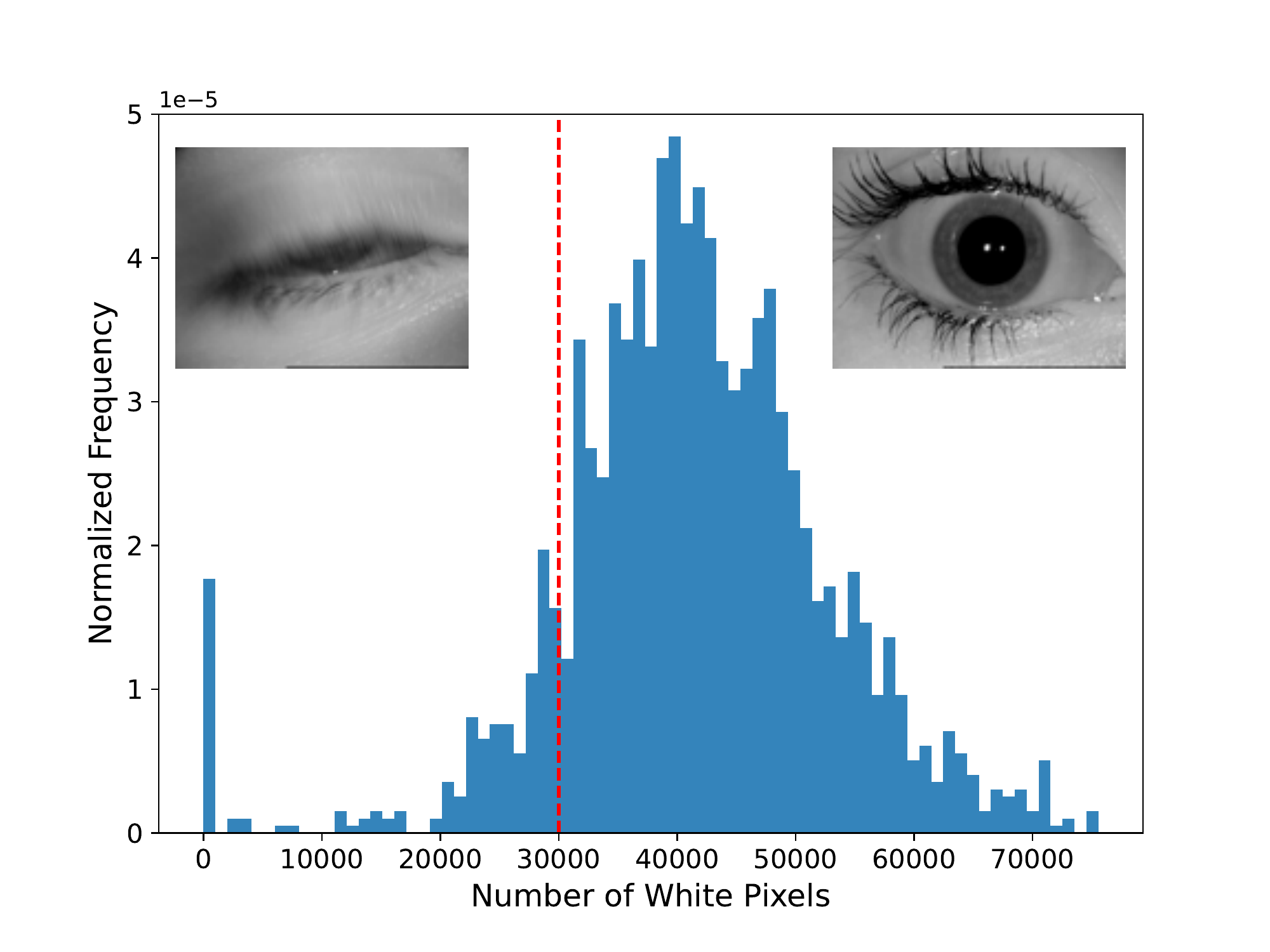}
    \caption{Training data curation step: number of white pixels per iris segmentation mask is helpful to detect images of blinking eyes. Images left of the dashed line were discarded from training due to lack of sufficiently-large iris texture regions in corresponding iris image.}
    \label{fig:whitePixels}
\end{figure}

\section{Data}
\label{Data}

The dataset used to train our StyleGAN3 generator network is a subset of a larger dataset collected for iris liveness detection \cite{czajka2015pupil} and volumetric iris segmentation \cite{Kinnison_ICB_2019}. The original dataset is comprised of iris-focused videos from 26 subjects (52 total irises) showing pupil size changes under visual light stimuli. At the 15 second mark in each video, a light is aimed into the subject's eyes, resulting in rapid pupil contraction, followed shortly thereafter by pupil dilation. In total, this collection of video data contains 117,717 grayscale images at a resolution of 768 $\times$ 576. This specific data has been selected due to large controlled pupil dilation difference among samples from the same eyes, leading to increased within-class variance.

Before initiating the training process, we filtered the dataset to only include images where the subject was not in the process of blinking. This was accomplished by filtering out images with minimal iris texture visible (based on segmentation mask), as shown in Fig. \ref{fig:whitePixels}. Secondly, given that (i) StyleGAN requires square training images and (ii) we want centered iris images, we removed all iris images that could not be center-cropped around the pupil center at a size of 512 $\times$ 512. This procedure removed all images with an iris located too close to the image border. Although the training irises are centrally located, the StyleGAN3 model was trained with a translationally-equivariant backbone, meaning we could still generate off-center iris images without having to retrain the model.

After the filtering and pre-processing steps, the training dataset contained 39,517 pupil-centered images from 47 irises at resolution 512$ \times$ 512. Due to low spatial correlation within the iris patterns, mirrored augmentations (along the left-right dimension) were included, effectively doubling the size of our training data.

\begin{figure}[t]
    \centering
    \includegraphics[width=\columnwidth]{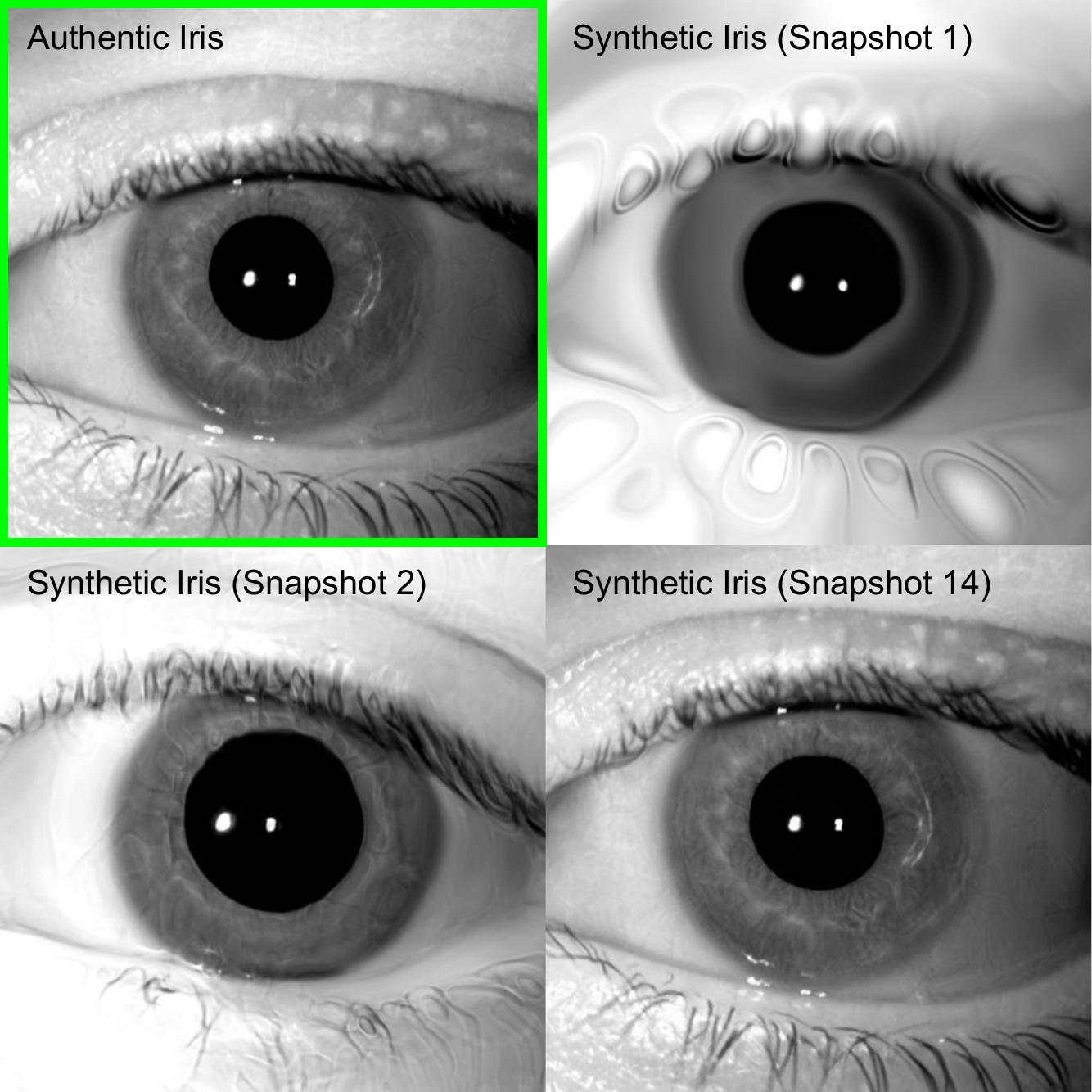}
    \caption{(Top-left) genuine iris seen during training; (top-right) synthetic iris from \textit{snapshot 1} with bubble artifacts; (bottom-left) synthetic iris from \textit{snapshot 2} with slightly finer details; (bottom-right) synthetic iris from \textit{snapshot 14} that matches synthetic iris (top-left) almost perfectly. All synthetic irises come from the same random seed.}
    \label{fig:seed159}
\end{figure}

\section{Methodology \& Experiments} 
\label{Methodology}

After curating our dataset to work with the native StyleGAN3 training regime, we configured and initiated our training process according to our hardware and dataset specifications. In total, our model (with a \verb+stylegan3-t+ base configuration) was trained for 4.56 million iterations via infinite random sampling of images (original and augmented) with a batch size of 32, as suggested by StyleGAN authors.
Once we were qualitatively satisfied with the quality of the synthesized samples and quantitatively satisfied with the Fr\'{e}chet Inception Distance between training and synthetic images of~36.6, the training process was terminated to avoid further overfitting.

Throughout the training process, model snapshots and metrics were logged every 80,000 training iterations. With a total of 4.8 million iterations, the training process produced 60 model snapshots, with which we could generate images. Of the 60 total snapshots, 14 of these (spaced at intervals of 320k iterations with an initial 80k iteration offset) were used to generate images for later experiments to study identity leakage as a function of training iteration\footnote{Training iteration, training time, and snapshot number are inherently related throughout the training process, and are thus used interchangeably in this work.}. There is an 80k iteration offset since synthetic samples at snapshot 0 are iris-less random noise \ie the model has not yet learned how to generate images that show irises. As such, {\it snapshot 1} corresponds to the model snapshot at which point 80k randomly sampled images (with replacement) have been seen during training, {\it snapshot 2} corresponds to 400k images seen, etc.

For each of the 14 model snapshots of interest, we generated 200 random iris images using seeds 0 through 199. Direct outputs from StyleGAN3 synthesis were 512 $\times$ 512 (the same resolution as the training images). In order to make the generated iris images closer to ISO-compliant samples and thus properly extract iris templates, every synthetic sample was padded with gray bars on the left and right side in order to restore the original 4:3 aspect ratio. Such samples were then resized to (640 $\times$ 480) and passed to the template extraction step.

\begin{figure*}[t]
    \centering
    \begin{subfigure}{\columnwidth}
        \centering
        \includegraphics[width=\columnwidth]{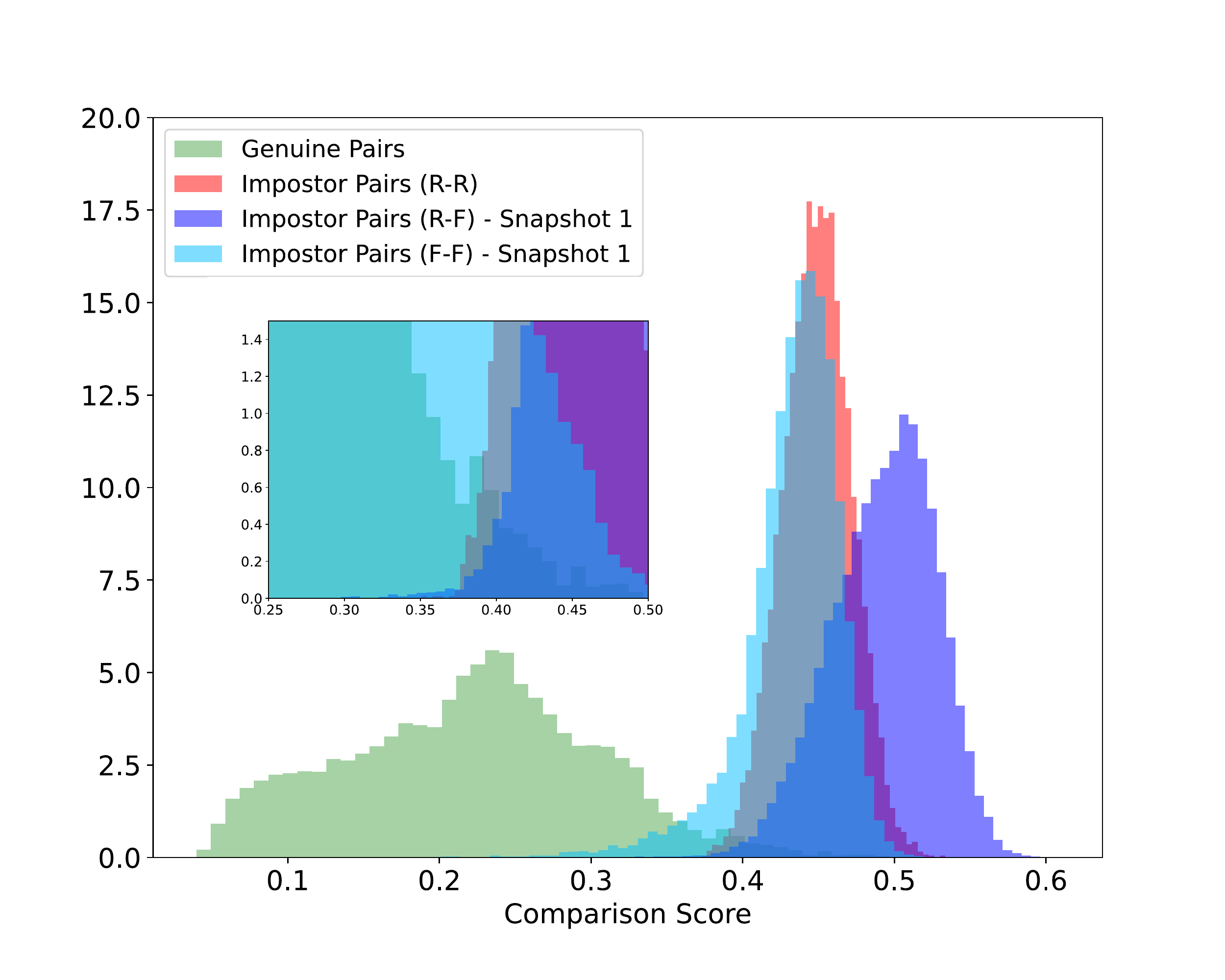}
        \caption{HDBSIF}
        \label{fig:hdbsif-snap1}
    \end{subfigure}
    \hfill
    \begin{subfigure}{\columnwidth}
        \centering
        \includegraphics[width=\columnwidth]{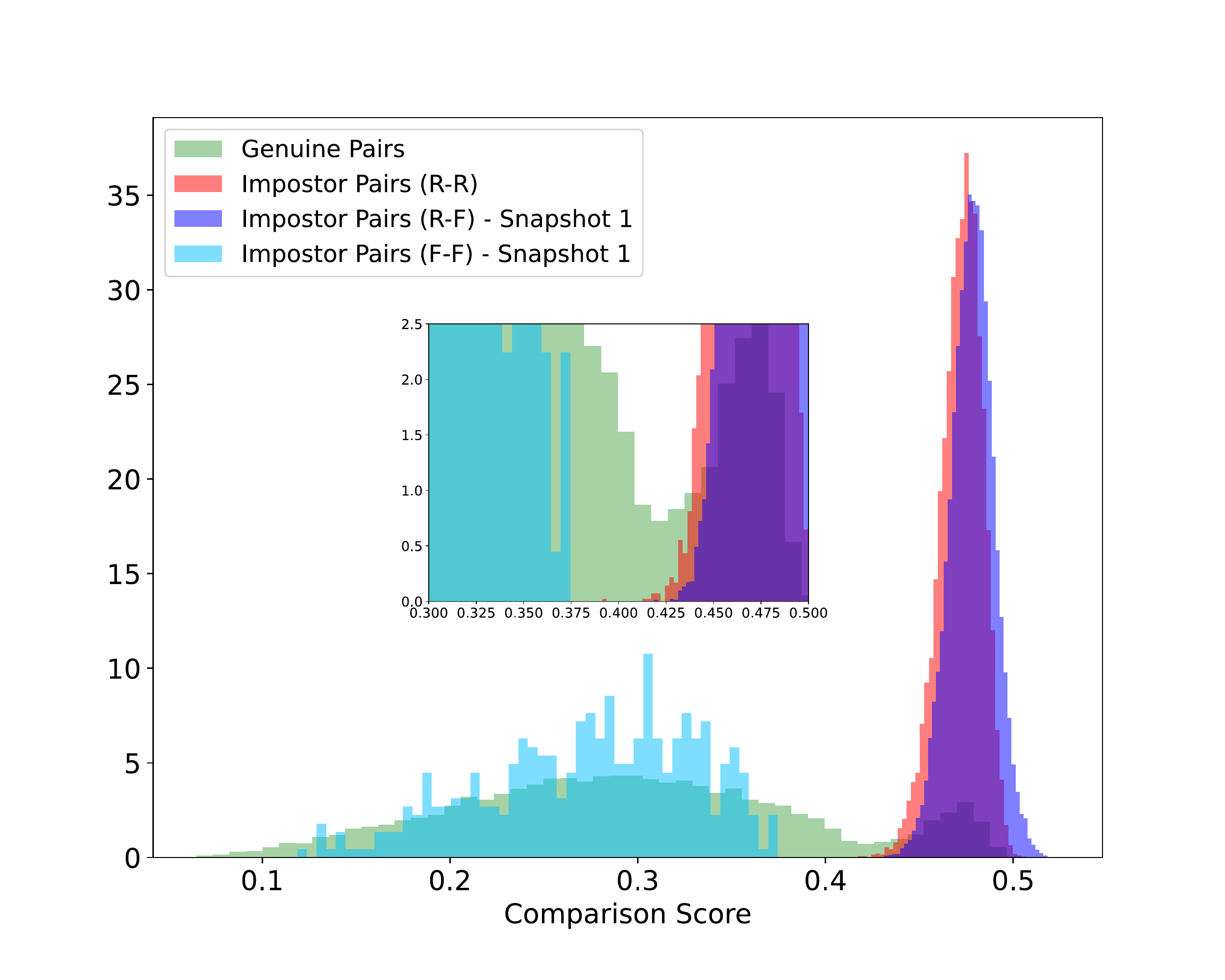}
        \caption{USIT3}
        \label{fig:usit-snap1}
    \end{subfigure}
    \caption{Comparison score distributions for the HDBSIF and USIT3 matchers for images generated at {\it snapshot 1}.}
    \label{fig:snap1}
\end{figure*}

Iris templates were consequently extracted for all images (real and fake). Three different iris code extractors were used to comparatively match irises: the domain-specific binarized statistical image feature extractor (\textbf{HD-BSIF}) from~\cite{czajka2019domain}, an iris toolkit from the University of Salzburg (\textbf{USIT3})~\cite{USIT3}, and \textbf{VeriEye} from Neurotechnology~\cite{verieye}. Features from the first two extractors are distance-based, meaning large comparison scores represent different irises. The commercial matcher VeriEye is similarity-based, meaning large comparison scores represent similar irises.

After iris templates had been extracted, four types of match scores were generated for each matcher: 
\begin{enumerate}
    \item \textbf{Genuine Pairs}, showing match scores for pairs of real images of the same iris.
    \item \textbf{Impostor R-R Pairs}, showing match scores for pairs of real ({\bf R}) images of two different irises. 
    \item \textbf{Impostor R-F Pairs}, showing match scores for pairs of images where one is real ({\bf R}; seen during training) and the other is fake ({\bf F}; generated after training). 
    \item \textbf{Impostor F-F Pairs}, showing match scores for pairs of images where both images are fake ({\bf F}).
\end{enumerate}

All four score distribution types were calculated across all three matchers for snapshots of interest during the training process, and are discussed in Section \ref{Results}.

\section{Results} 
\label{Results}

The results quantitatively and qualitatively differ depending on the training time. Hence, we grouped them into three training time windows and provide separate commentary on each one: after the first 80k iterations of training ({\it snapshot 1}; Sec. \ref{sec:snap1}), after 400k training iterations ({\it snapshot 2}; Sec. \ref{sec:snap2}), and after 4.56M training iterations ({\it snapshot 14}; Sec. \ref{sec:snap4}). 
Analyses of snapshots 3 through 13 are not listed in this work but can be rightfully treated as gradual interpolations from \textit{snapshot 2} (showing some evidence of leakage) to \textit{snapshot 14} (showing extensive evidence of leakage). Tinsley \etal discuss how to observe identity leakage by comparing the shifts in impostor score distributions calculated between different real faces and between real and synthetic faces. We follow this approach for irises.

\subsection{Snapshot 1: After 80k Training Iterations}
\label{sec:snap1}

Images generated at {\it snapshot 1} are easily recognizable as fake, as seen in Figure \ref{fig:seed159} (top-right). The irises demonstrate ``water bubble'' artifacts and lack detail in iris texture. As a commercial product, the VeriEye matcher has a built-in quality control mechanism that runs before template generation. At {\it snapshot 1}, VeriEye rejected all 200 fakes as low-quality samples, and consequently did not generate templates.

Without this pre-feature extraction quality assessment, the HDBSIF and USIT3 methods were able to generate iris templates from the bubble-ridden samples. Comparison scores between these extracted templates were used to plot the four distributions described in Section \ref{Methodology}. In Figure \ref{fig:snap1}, we notice a shift to the \textit{right} for the impostor R-F distributions. Namely,
real versus fake (R-F) image pairs produce larger match scores than the (R-R) pairs, suggesting real irises from different subjects are \textit{closer} to each other than real irises are to the fakes generated by StyleGAN3 model. This may be caused by an eye rotation mechanism that does not function well for very low-quality iris textures present in R-F matching pairs. Given the lack of defined iris texture, this identity ``obfuscation'' is to be expected.

Figure \ref{fig:snap1} also shows impostor F-F distributions for HDBSIF and USIT3, which are shifted \textit{left} towards genuine scores, implying a high degree of similarity among the synthetic images towards the beginning of the training process. We hypothesize that the lack of fine details in iris texture (used to discern identity) is responsible for this shift.

\begin{figure*}[t]
    \centering
    \begin{subfigure}{.3\textwidth}
        \centering
        \includegraphics[width=\textwidth]{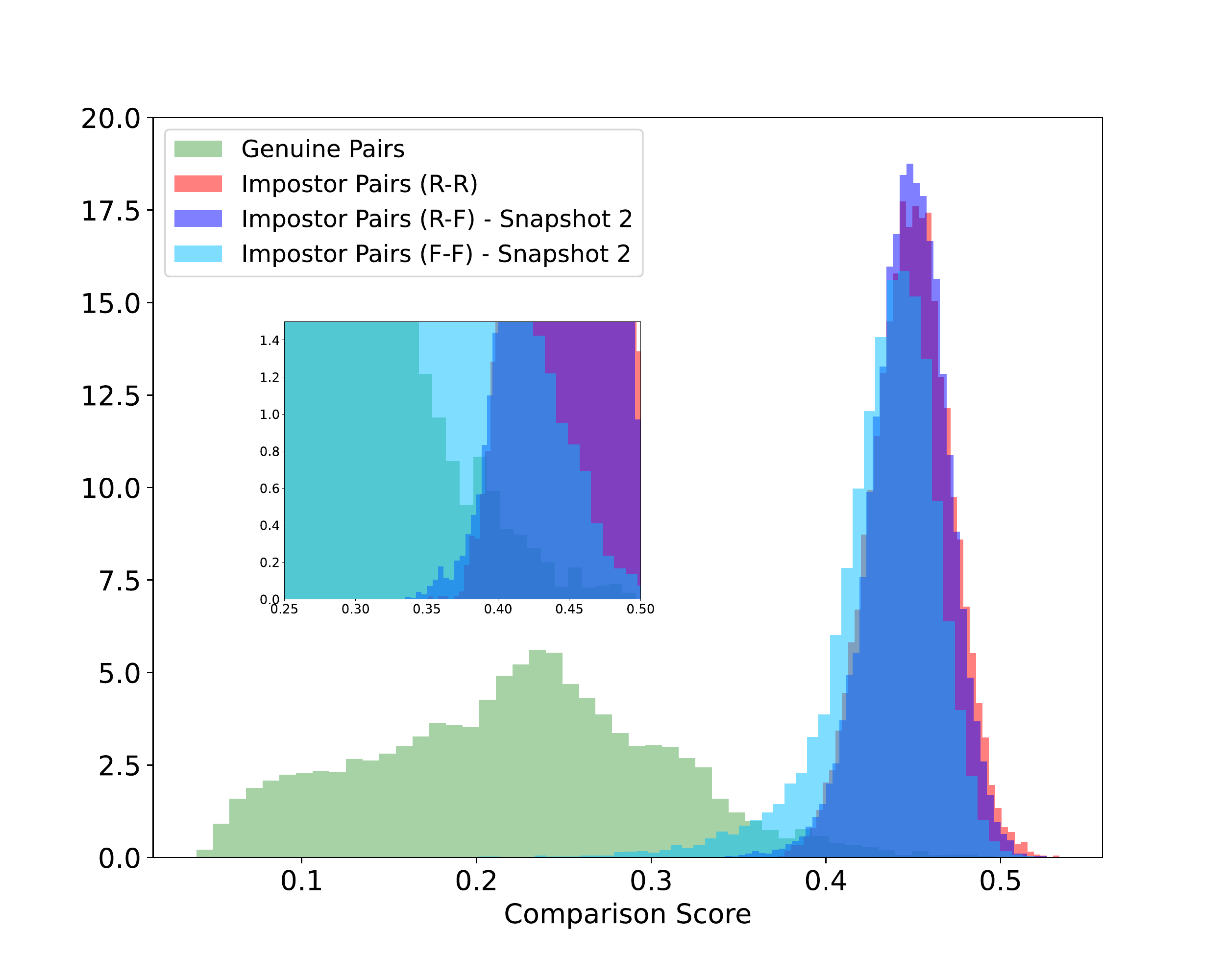}
        \caption{HDBSIF}
        \label{fig:hdbsif-snap2}
    \end{subfigure}
    \begin{subfigure}{.3\textwidth}
        \centering
        \includegraphics[width=\textwidth]{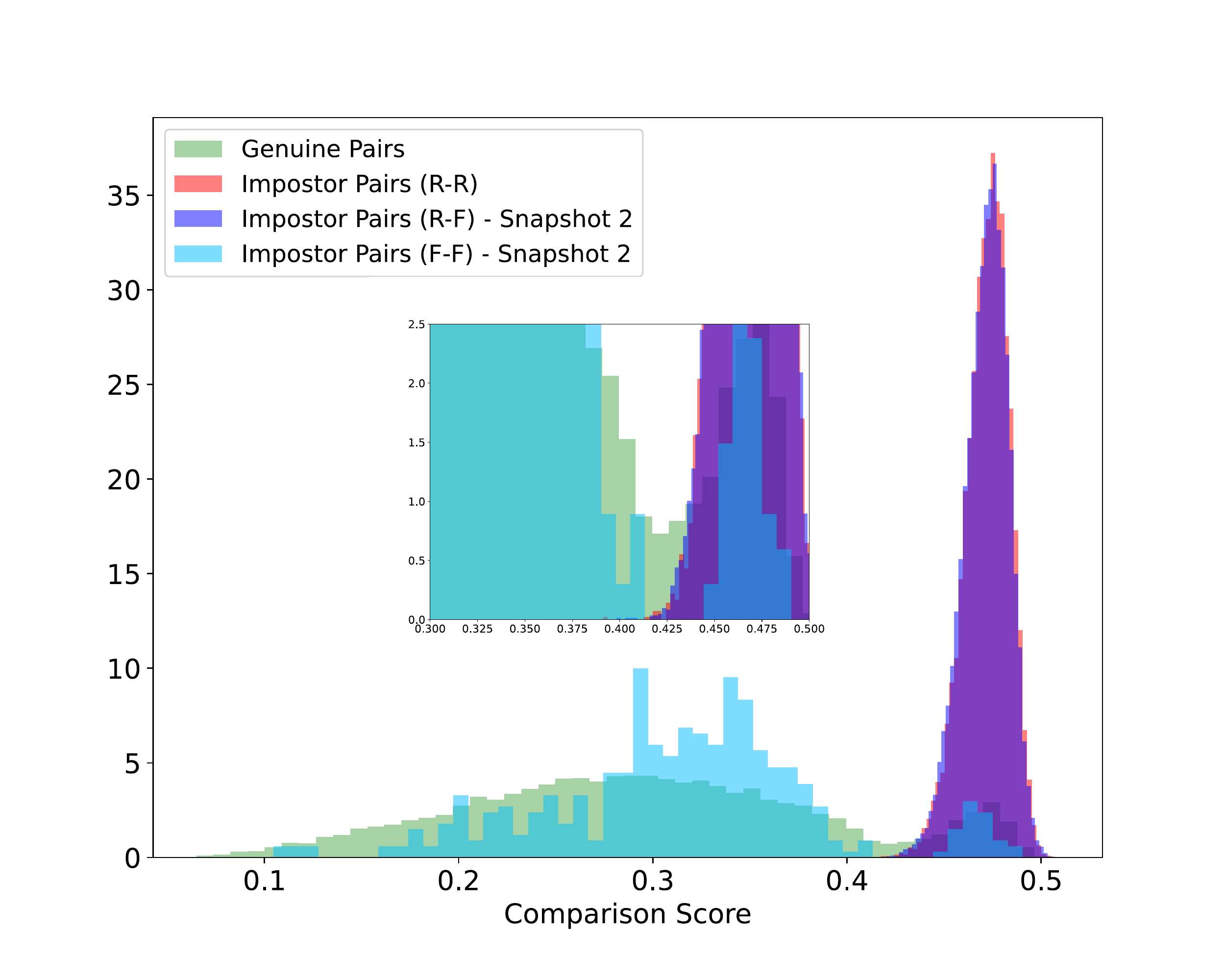}
        \caption{USIT3}
        \label{fig:usit-snap2}
    \end{subfigure}
    \begin{subfigure}{0.3\textwidth}
        \centering
        \includegraphics[width=\textwidth]{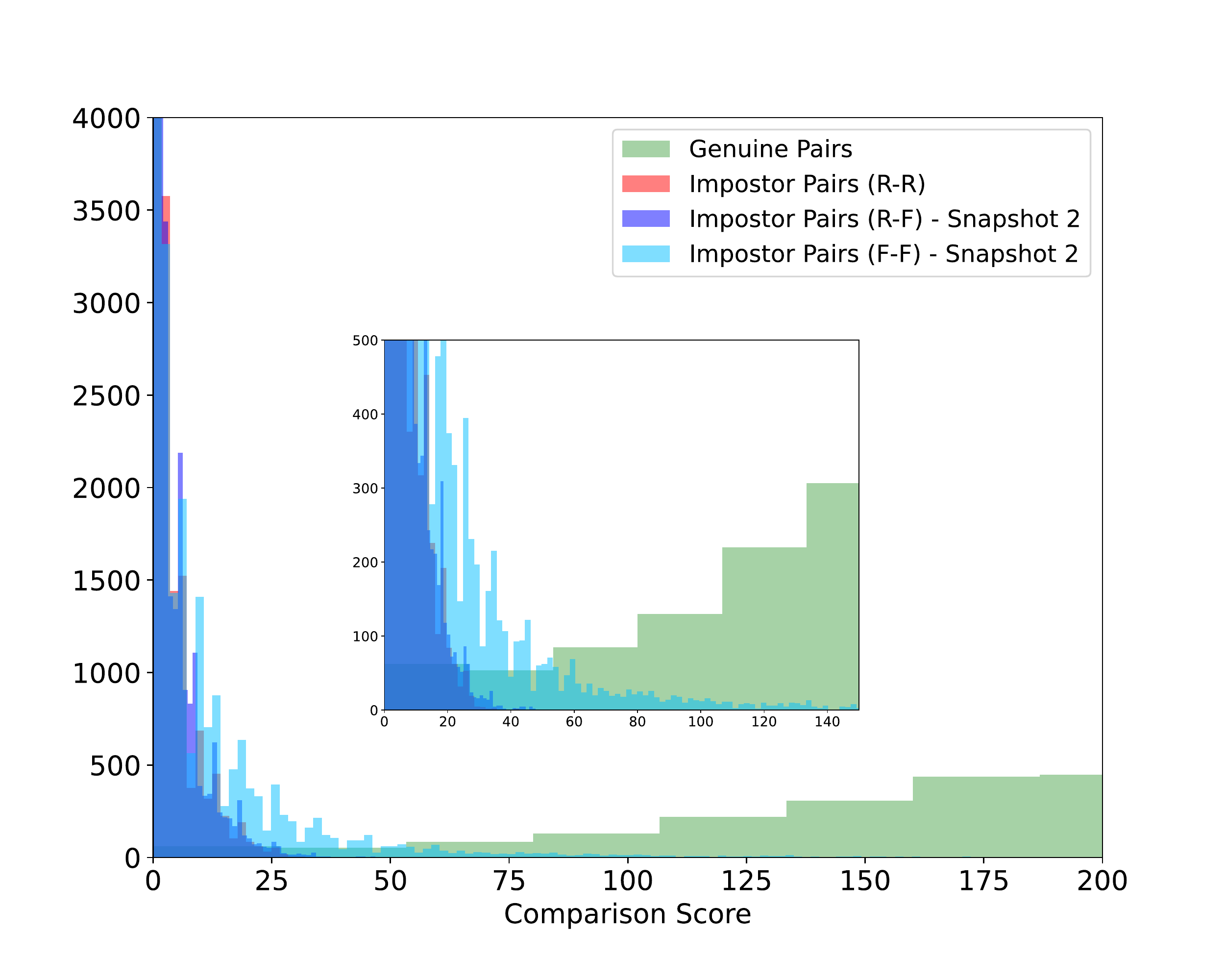}
        \caption{VeriEye}
        \label{fig:verieye-snap2}
    \end{subfigure}
    
    \caption{Match score distributions for HDBSIF, USIT3, and VeriEye results for images generated at \textit{snapshot 2}.}
    \label{fig:snap2}
\end{figure*}

\begin{figure*}
    \centering
    \begin{subfigure}{.3\textwidth}
        \centering
        \includegraphics[width=\textwidth]{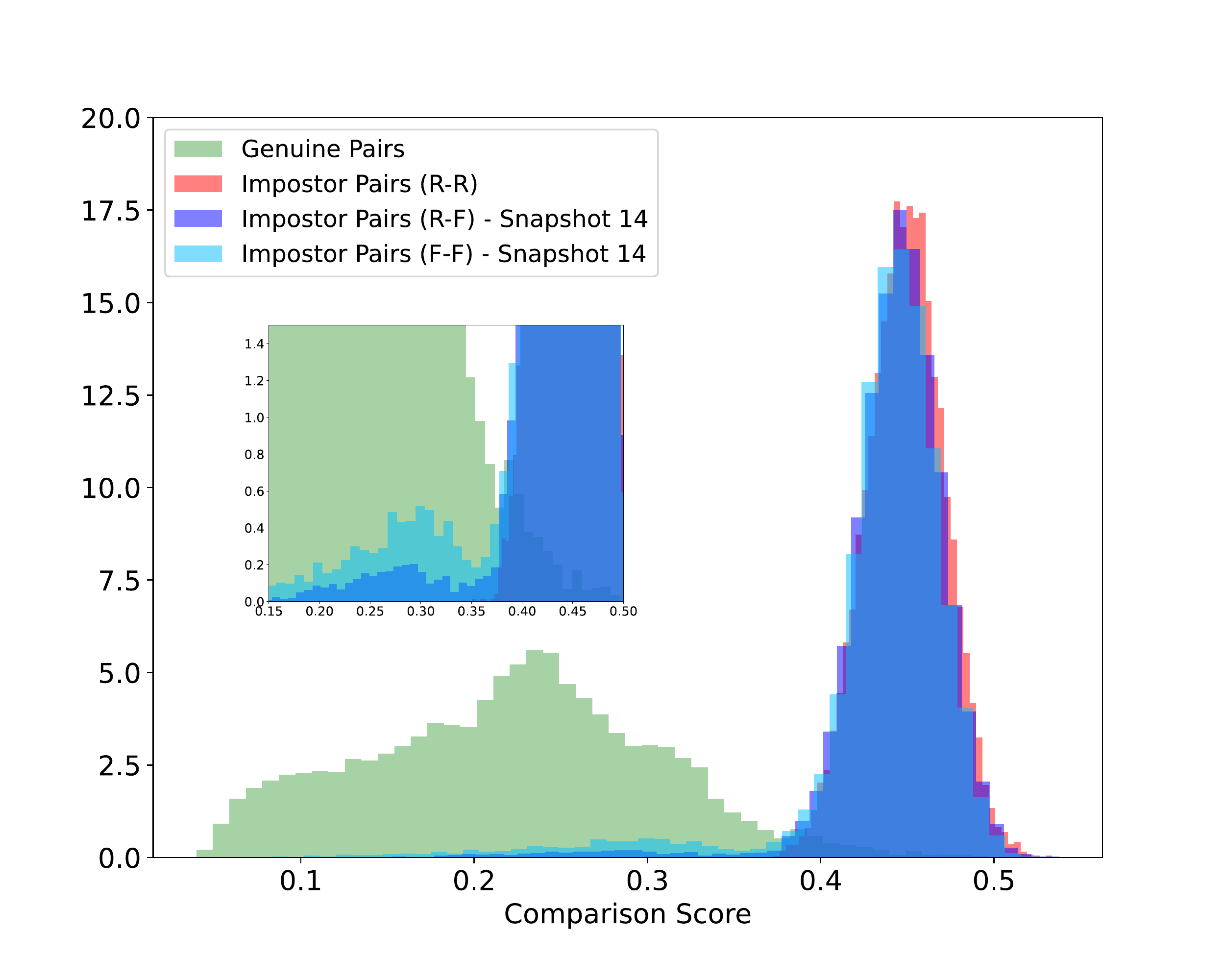}
        \caption{HDBSIF}
        \label{fig:hdbsif-snap14}
    \end{subfigure}
    \begin{subfigure}{.3\textwidth}
        \centering
        \includegraphics[width=\textwidth]{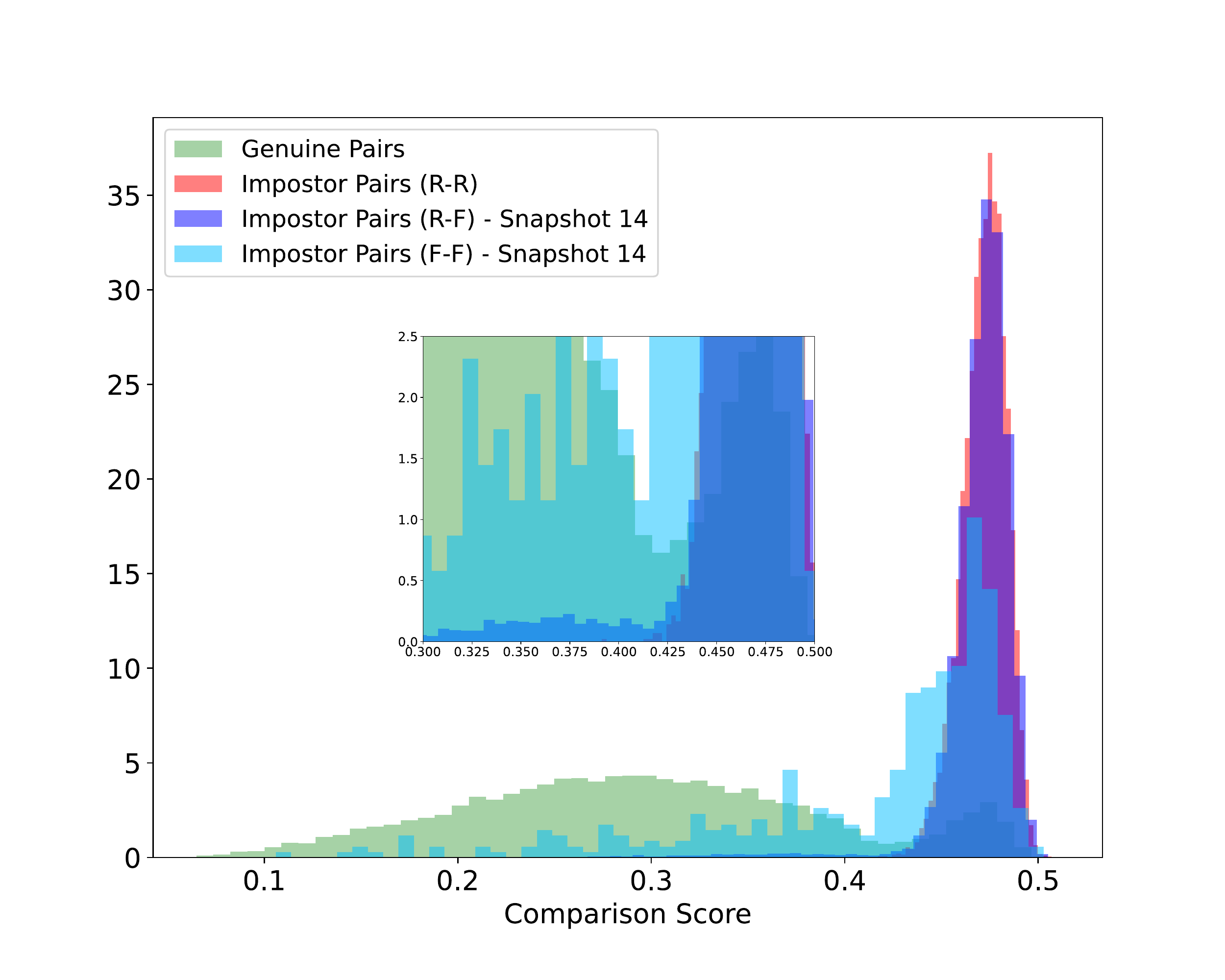}
        \caption{USIT3}
        \label{fig:usit-snap14}
    \end{subfigure}
    \begin{subfigure}{0.3\textwidth}
        \centering
        \includegraphics[width=\textwidth]{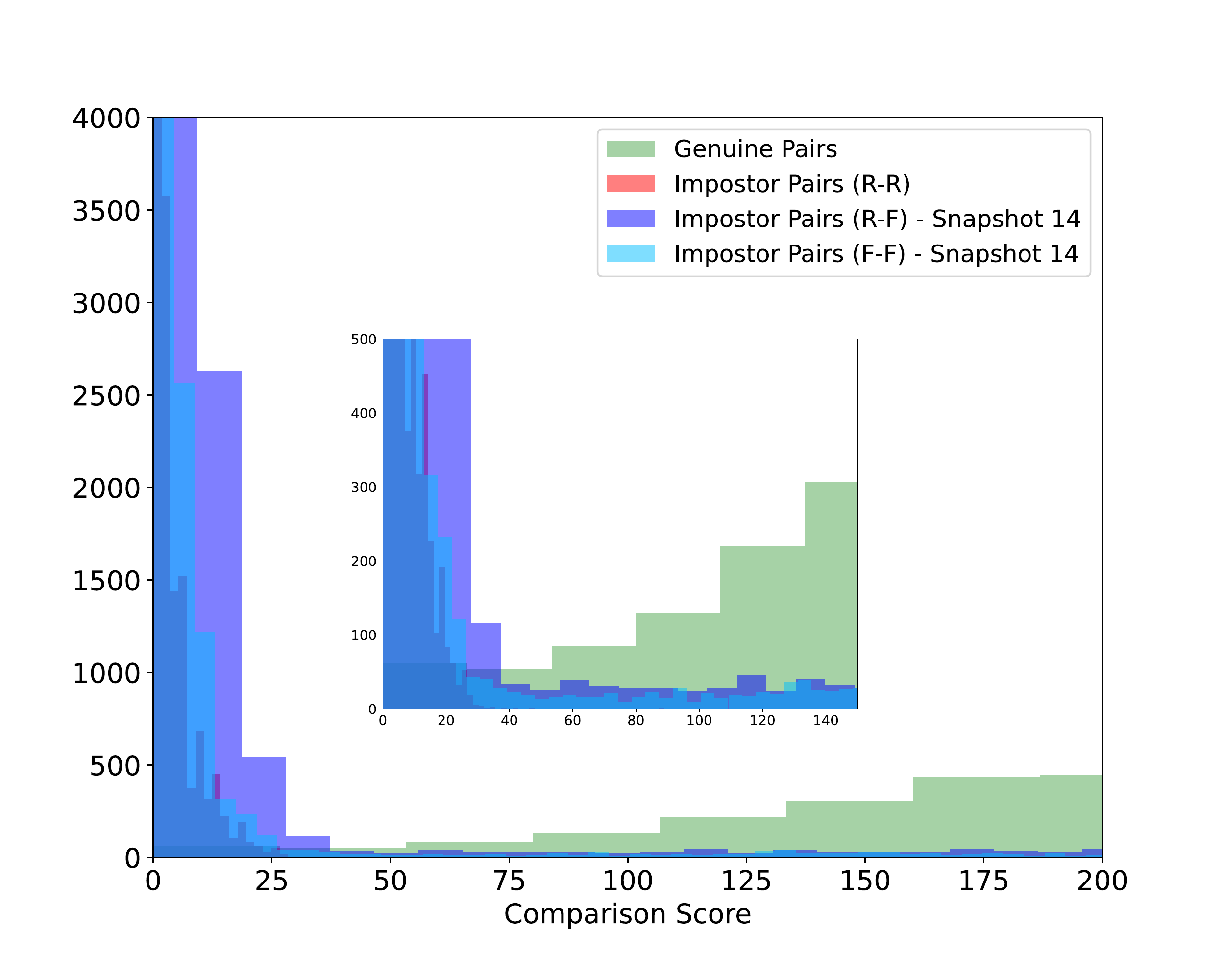}
        \caption{VeriEye}
        \label{fig:verieye-snap14}
    \end{subfigure}
    
    \caption{Match score distributions for HDBSIF, USIT3, and VeriEye results for images generated at \textit{snapshot 14}.}
    \label{fig:snap14}
\end{figure*}

\begin{figure*}
    \centering
    \begin{subfigure}{.3\textwidth}
        \centering
        \includegraphics[width=\textwidth]{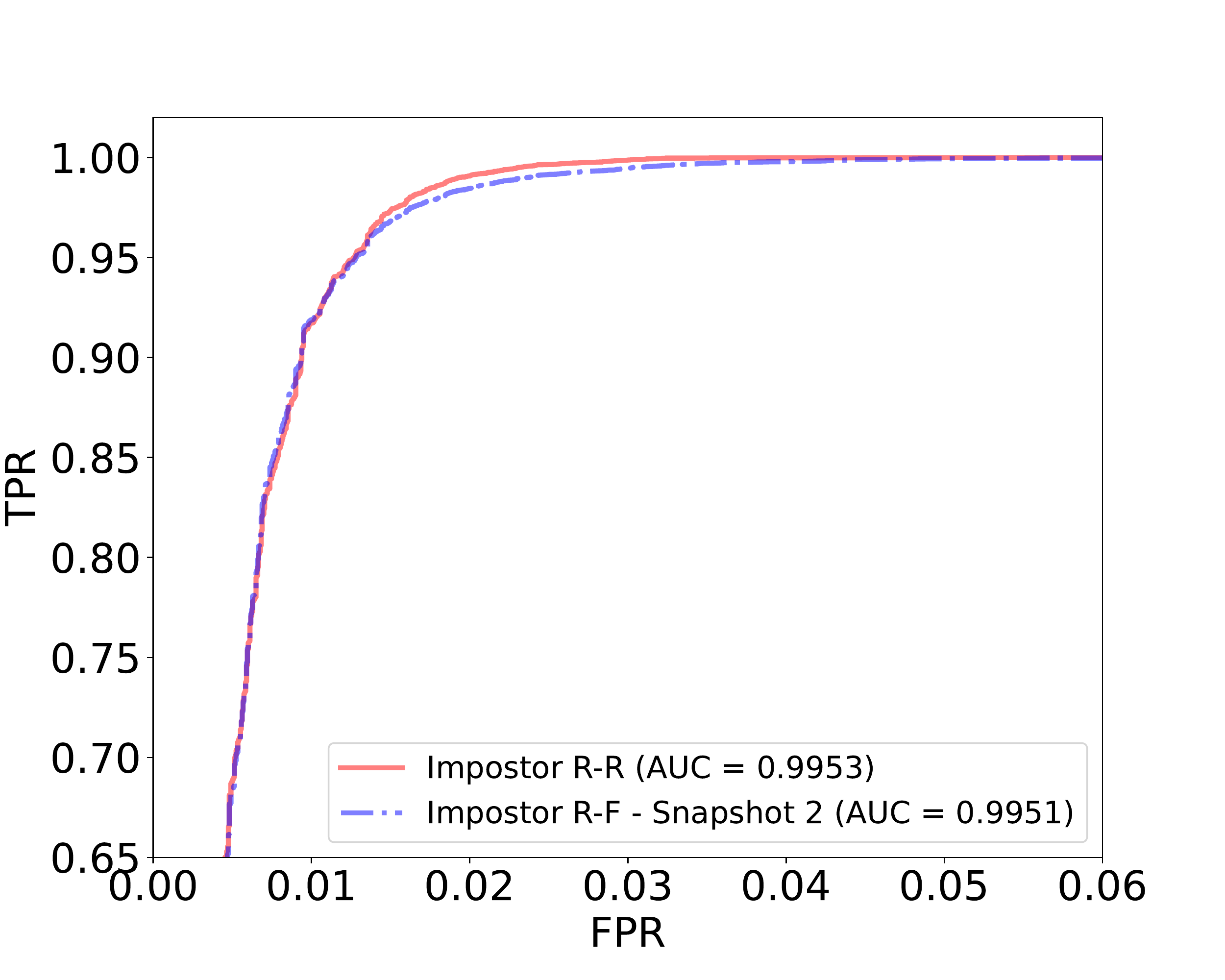}
        \caption{HDBSIF}
        \label{fig:hdbsif-snap14-roc}
    \end{subfigure}
    \begin{subfigure}{.3\textwidth}
        \centering
        \includegraphics[width=\textwidth]{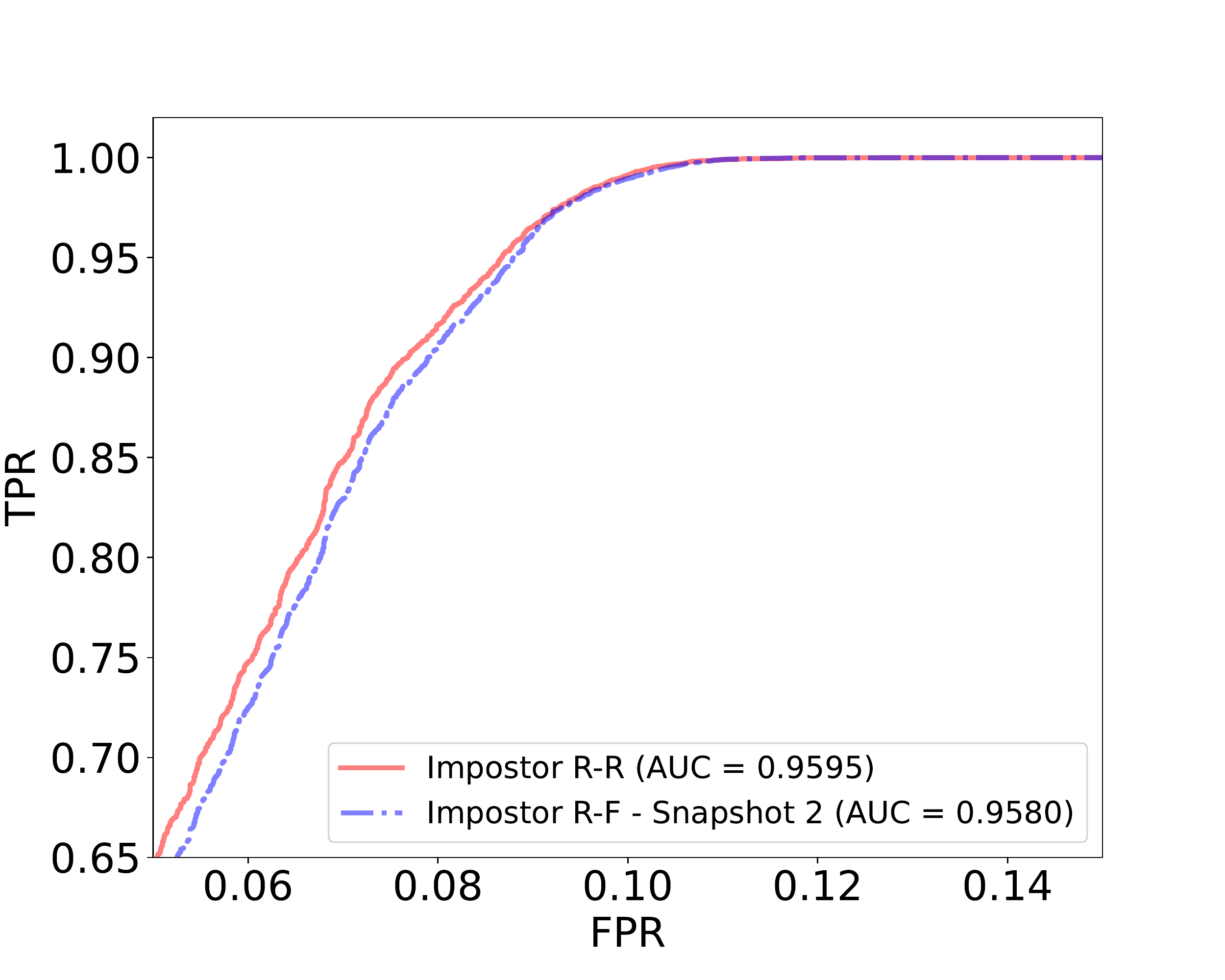}
        \caption{USIT3}
        \label{fig:usit-snap14-roc}
    \end{subfigure}
    \begin{subfigure}{0.3\textwidth}
        \centering
        \includegraphics[width=\textwidth]{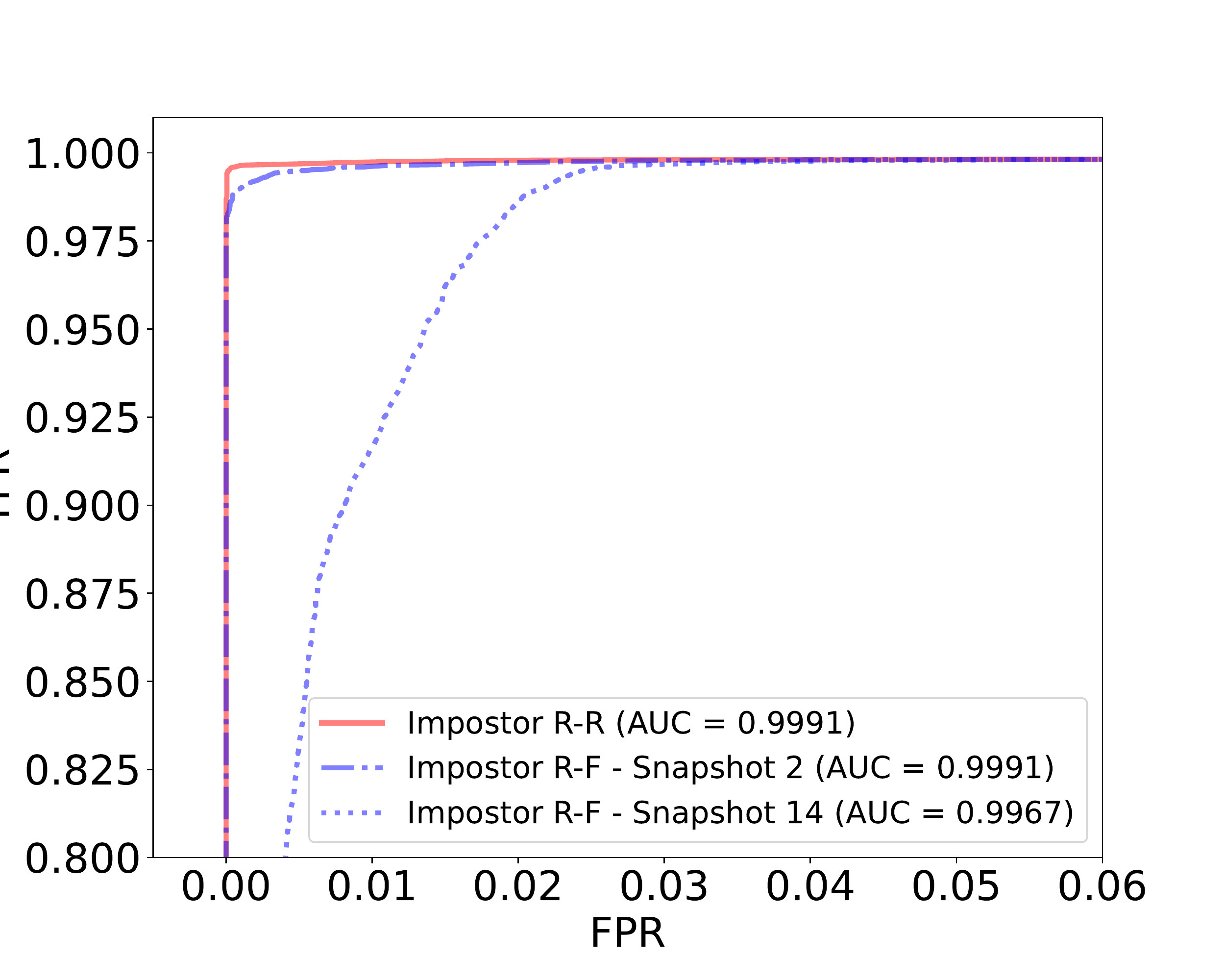}
        \caption{VeriEye}
        \label{fig:verieye-snap14-roc}
    \end{subfigure}
    \caption{Receiver operating characteristic (ROC) curves for impostor comparison scores across three matchers: HDBSIF, USIT3, and VeriEye. FPR = False Positive Rate. TPR = True Positive Rate. AUC = Area Under ROC Curve. While the curves obtained for authentic samples and those generated by lightly-trained ({\it snapshot 2}) StyleGAN3 model are similar, the ROC in case of using synthetic irises generated by well-trained ({\it snapshot 14}) StyleGAN3 suggests significant decrease in performance. The latter is caused by more synthetic samples being matched with genuine samples used previously in training (= identity leakage).}
    \label{fig:rocs}
\end{figure*}

\subsection{Snapshot 2: After 400k Training Iterations}
\label{sec:snap2}

At {\it snapshot 2}, we observe greater agreement between the R-R and R-F impostor distributions. At this point, synthetic images pass the quality assessment performed by the VeriEye matcher, and it appears that the model (after ingesting 400,000 images) has also learned iris photo-realism. The agreement between the R-R and R-F distributions suggests that the majority of synthetic samples do not match samples in the training data, and can thus be treated as privacy-preserving, ``safe-to-use'' synthetic irises, e.g., for the sake of dataset augmentation.

However, we see slightly {\bf longer tails toward the direction of the genuine score distribution}: left for HDBSIF and USIT3, and right for VeriEye. These tails suggest that there are fake-real image pairs that would be considered a match and thus there is a degree of identity leakage. That is, close-to-training samples may be appearing amongst the ensemble of synthetic images.

For the HDBSIF and USIT3 matchers, we see a general shift to the \textit{right} in the impostor F-F distributions (light-blue), relative to those from \textit{snapshot 1} (Figure \ref{fig:snap1}). This shift in impostor F-F scores towards impostor R-R scores suggests that our model is producing more realistic irises with defined textural information that can be used to discerned different identities, which was not the case in \textit{snapshot 1}. Additionally, in all three matchers, genuine-leaning tails are apparent in the impostor F-F distribution, meaning that the textural information being learned is similar across some synthetic samples.

\subsection{Snapshot 14: After 4.56M Training Iterations}
\label{sec:snap4}

At {\it snapshot 14}, there is also a high degree of overlap between the R-R and R-F impostor distributions. As in {\it snapshot 2}, this overlap suggests most fake samples do not divulge identity-salient information. However, the genuine-leaning tails of the \textbf{Impostor R-F} distribution are now extremely pronounced, displaying evidence of severe identity leakage in the case of a few synthetic samples, as highlighted and zoomed in on Fig. \ref{fig:snap14}.

Figure \ref{fig:rocs} shows receiver operating characteristic (ROC) curves for each matcher for snapshots 2 and 14; ROC curves for outstanding snapshots are included in the Supplementary Materials. As seen in the figure, the matchers show a decrease in discriminatory performance as model training progresses as a result of overfitting during the training process.

\begin{figure}[t]
    \centering
    \includegraphics[width=\columnwidth]{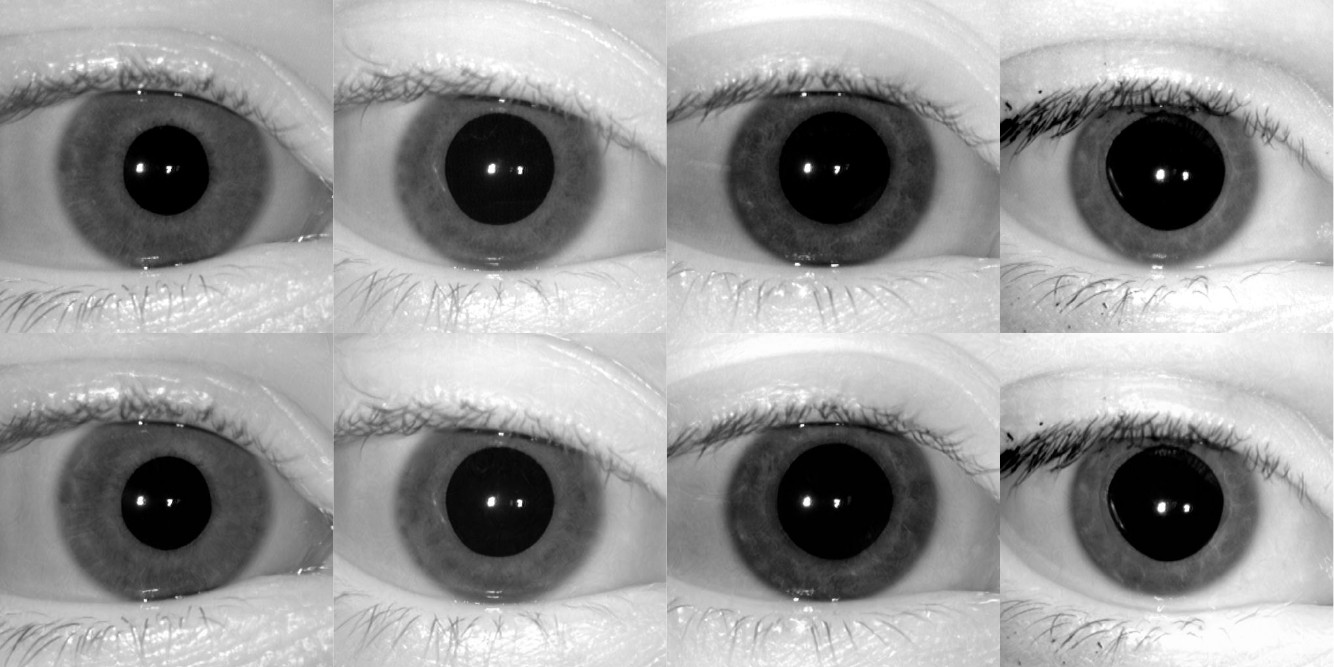}
    \caption{Instances of near identical training (real) and synthesised samples. The top row shows real images from the training data, and the bottom row shows fake images, generated by the model at {\it snapshot} 14. All three matchers judged the image pair as a bona fide match.}
    \label{fig:grid8}
\end{figure}

In the case of novel face generation it has been shown that identity leakage is evident, though matcher-dependent~\cite{tinsley2021face}. Qualitative analysis of real-fake images that falsely matched contained faces that were similar in race, gender, and even age. However, these images were demonstrably different images \ie the synthetic samples were \textit{similar} to the training samples, but were not \textit{exact} replicas of the training samples. In the case of novel iris generation with the StyleGAN3 model, when identity leakage does occur, it occurs as near-exact image re-construction of select training samples; this is much more concerning in the context of privacy than feature leakage in face synthesis reported previously in literature. Figure \ref{fig:grid8} shows several real-fake image pairs that all matchers judged as bona fide matches with high degree of confidence.

\begin{figure}[!htb]
    \centering
    \includegraphics[width=\columnwidth]{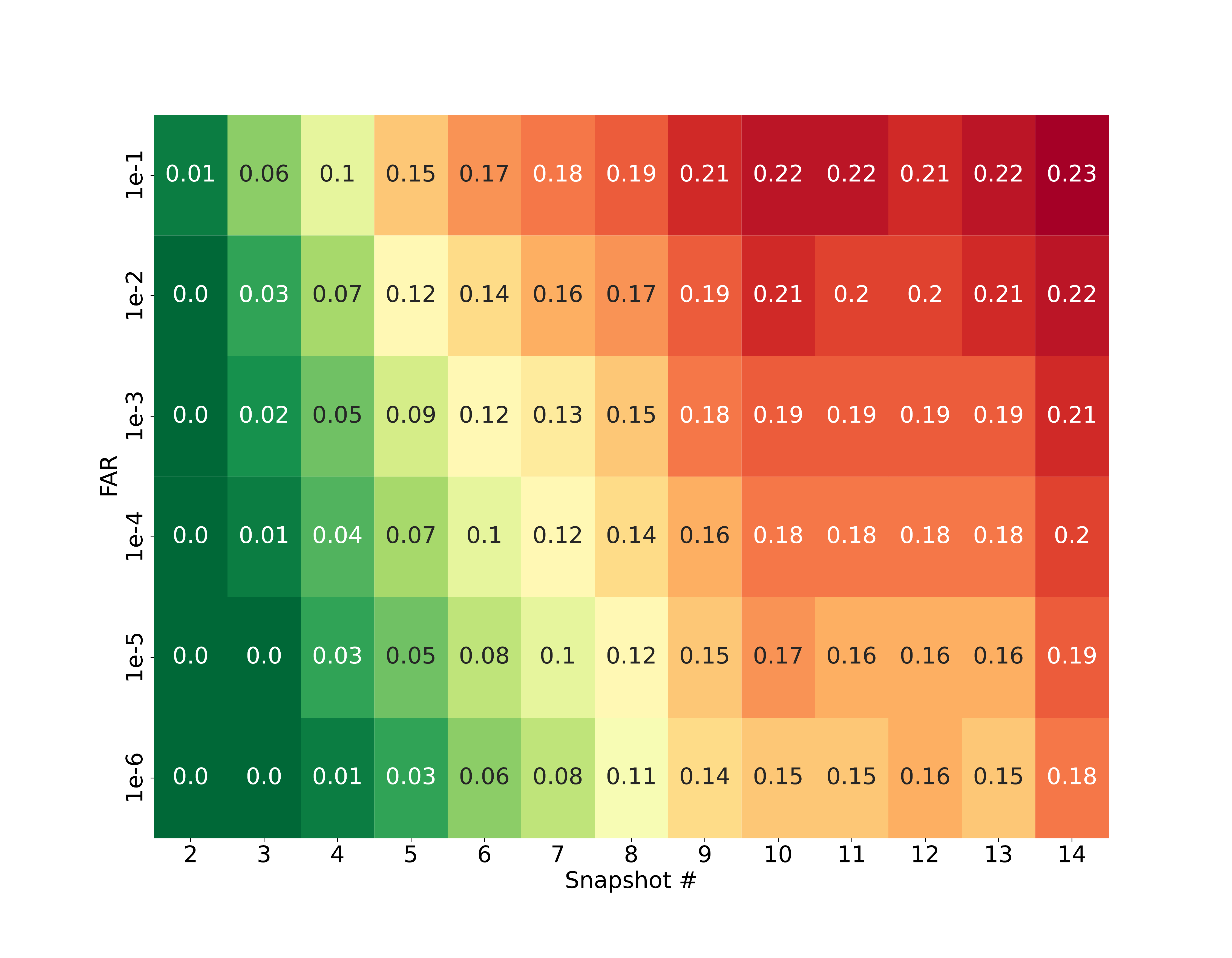}
    \caption{Percentage of pairs that falsely match between real and synthetic samples at NeuroTechnology-defined false acceptance rates (rows) per snapshot number.}
    \label{fig:heatmap}
\end{figure}

As a frame of reference, the developers of the VeriEye matcher claim that a match score greater than 96 maintains a false acceptance rate (FAR) of $1\mathrm{e}{-6}$, or 1 in 1,000,000; there are 706 such pairs that register a score greater than 96. Figure \ref{fig:heatmap} presents a heatmap of real-fake matching percentage at varying NeuroTechnology-defined false acceptance rates.

At this late stage in the training process, it appears the model has learned a generalized distribution of irises, but shows a tendency to remember a handful of samples from the training data. We believe that one of the factors responsible for this overfitting and associated leakage may be the video-based nature of the dataset in addition to the low number of different identities. The pre-trained face synthesis model was trained on 70,000 images in the FFHQ dataset, each representing a distinct identity. In the case of our iris synthesizer, the data comes from 25 fps video, meaning training images show minimal variety on a per iris basis and correspond to only one of 52 distinct irises. That is, the training data contains many close-to-replica images, saved during the periods of dilation and contraction. As a consequence, our generator processes massively redundant information, and ``learns'' to recreate the few base samples that have been seen several times. This is an important observation that may suggest that iris-specific generative models should include much more diversified training data, either by reducing the number of images per subject/iris or by increasing the number of distinct irises represented in the training data.

Beyond modifying the composition of the training data, changes to the StyleGAN3 architecture itself might mitigate the dangers of identity leakage. The size of the StyleGAN3's latent space is an inherent parameter in the synthesis procedure, meaning it can potentially be a source of identity leakage. And the latent space is vast (16 $\times$ 512 or 18 $\times$ 512 dimensions) compared to the experimentally-estimated number of degrees of freedom of the binomial distribution modeling the number of disagreeing bits of iris codes of different eyes, usually estimated to be around 200~\cite{Daugman_TIFS_2016}. Thus, an analysis of the latent space, for instance detecting salient dimensions, would hopefully allow for succinct, privacy-preserving iris synthesis.

Identity-leaking models, such as the model from {\it snapshot} 14, pose a significant threat against training subjects' privacy. For instance, if an overfit generative model were to be compromised, training samples could be recreated in order to steal a training subject's iris information. Additionally, since image generation from pre-trained models only takes seconds per image on a GPU, training subjects can be found rather quickly.

\section{Conclusions} \label{Conclusions}

In this work, we explore possibilities but also manage expectations when training generative models for the task of iris image synthesis. It appears that even with a relatively small dataset that features few subjects, StyleGAN3 is capable of producing photo-realistic irises, a common biometric technique in personal and societal security. However, given the vast amount of identity information that spills out of the model at nearly every point of the training process, there is a resounding call to exercise caution in choice and curation of training data, especially in biometrics. With privacy as a central pillar in the growing field of trustworthy AI, a great deal of scrutiny is required at all levels of model development and deployment process, including the associated training data.

We are not aware of any previous works investigating the problem of identity leakage observed for modern synthetic iris generators, such as StyleGAN3. With a recent quest for training data (both for designing iris recognition and presentation attack detection methods), and decreasing barriers of fast adoption of pre-trained deep learning-based models, this paper opens a new research area related to privacy of iris biometrics data generators. To facilitate further work in this area, we make all the snapshots of the StyleGAN3 model trained for this paper available to other researchers.

{\small
\bibliographystyle{ieee}
\bibliography{main}
}

\clearpage
\appendix
\onecolumn
\section*{Supplementary Materials:\\Image Samples and ROC Curves for Snapshots 2 -- 14}
\vskip2cm

\begin{figure*}[h]
    \centering
    \includegraphics[width=0.9\textwidth]{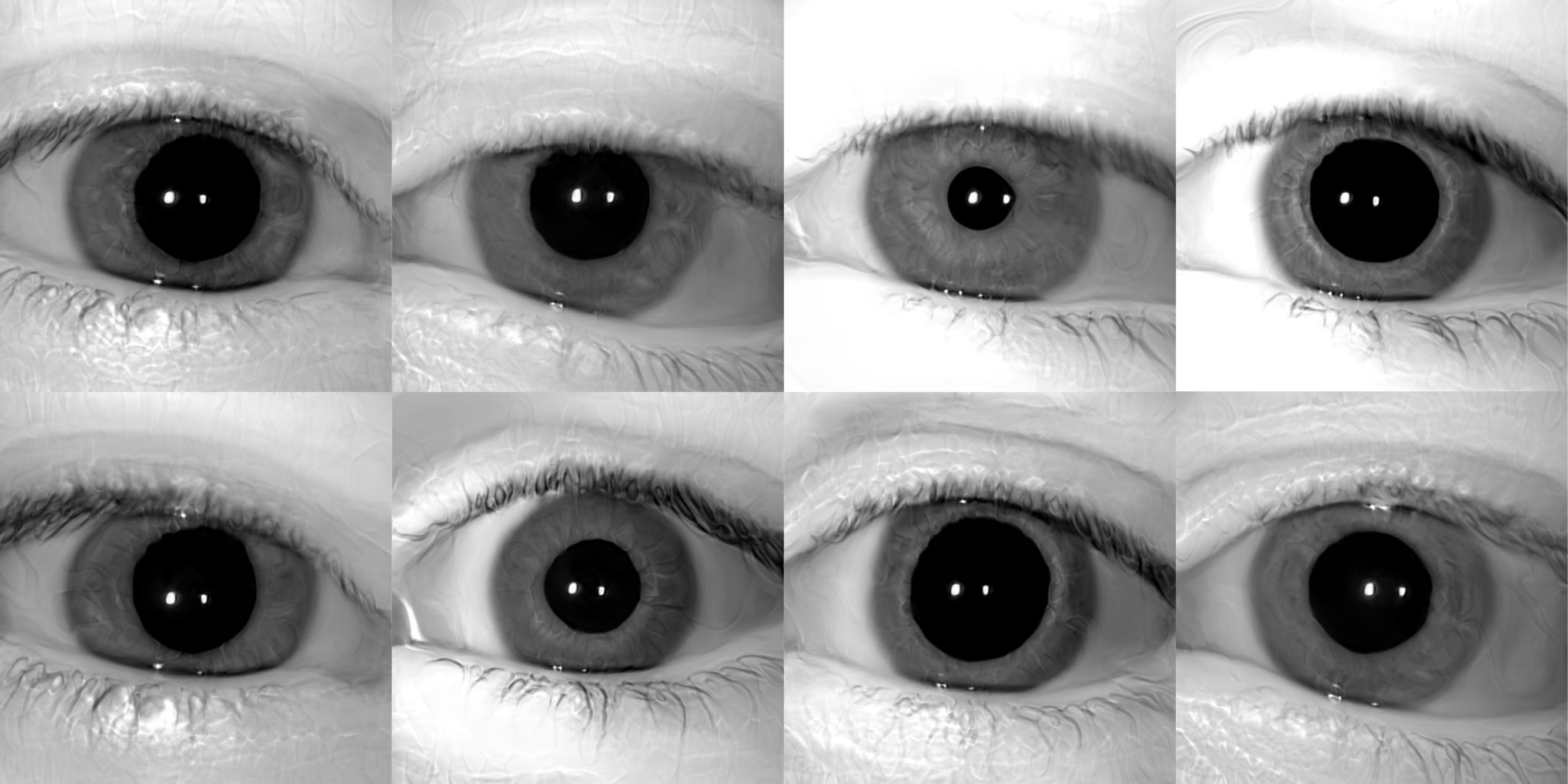}
    \caption{Image samples generated by the model at \textit{snapshot 2}.}
\end{figure*}
\begin{figure*}[h]
    \centering
    \begin{subfigure}{.31\textwidth}
        \centering
        \includegraphics[width=\textwidth]{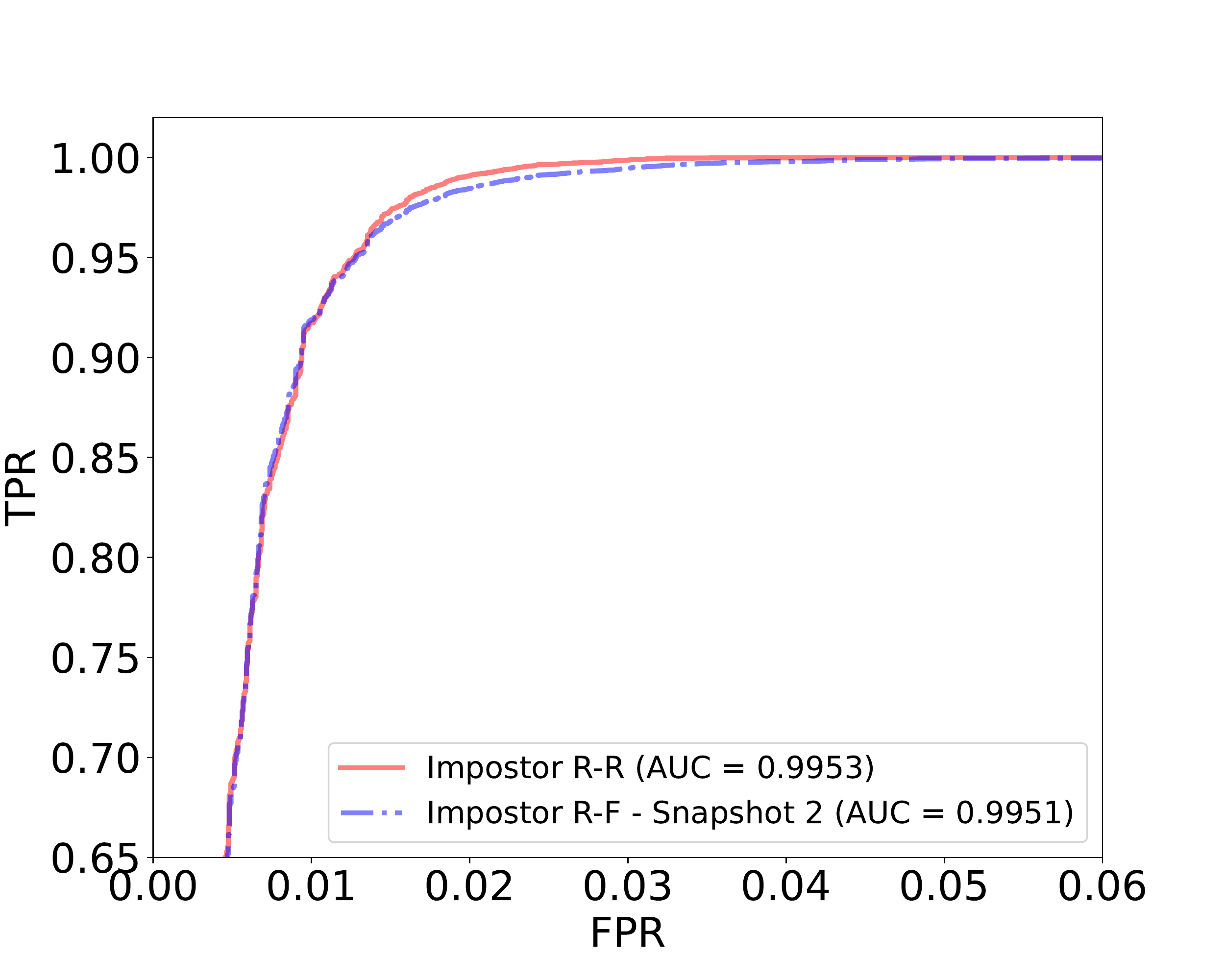}
        \caption{HDBSIF}
    \end{subfigure}%
    \begin{subfigure}{.31\textwidth}
        \centering
        \includegraphics[width=\textwidth]{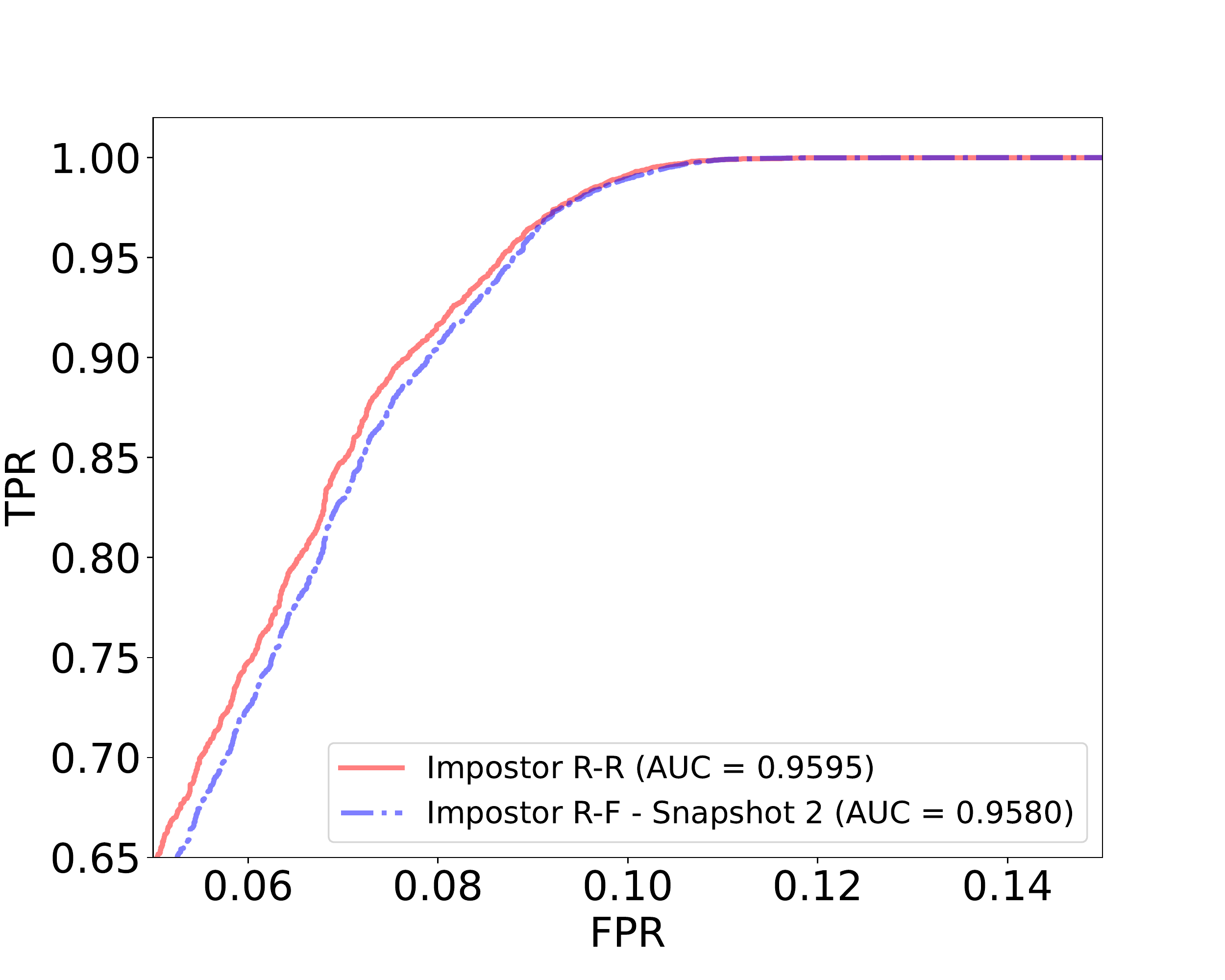}
        \caption{USIT3}
    \end{subfigure}%
    \begin{subfigure}{0.31\textwidth}
        \centering
        \includegraphics[width=\textwidth]{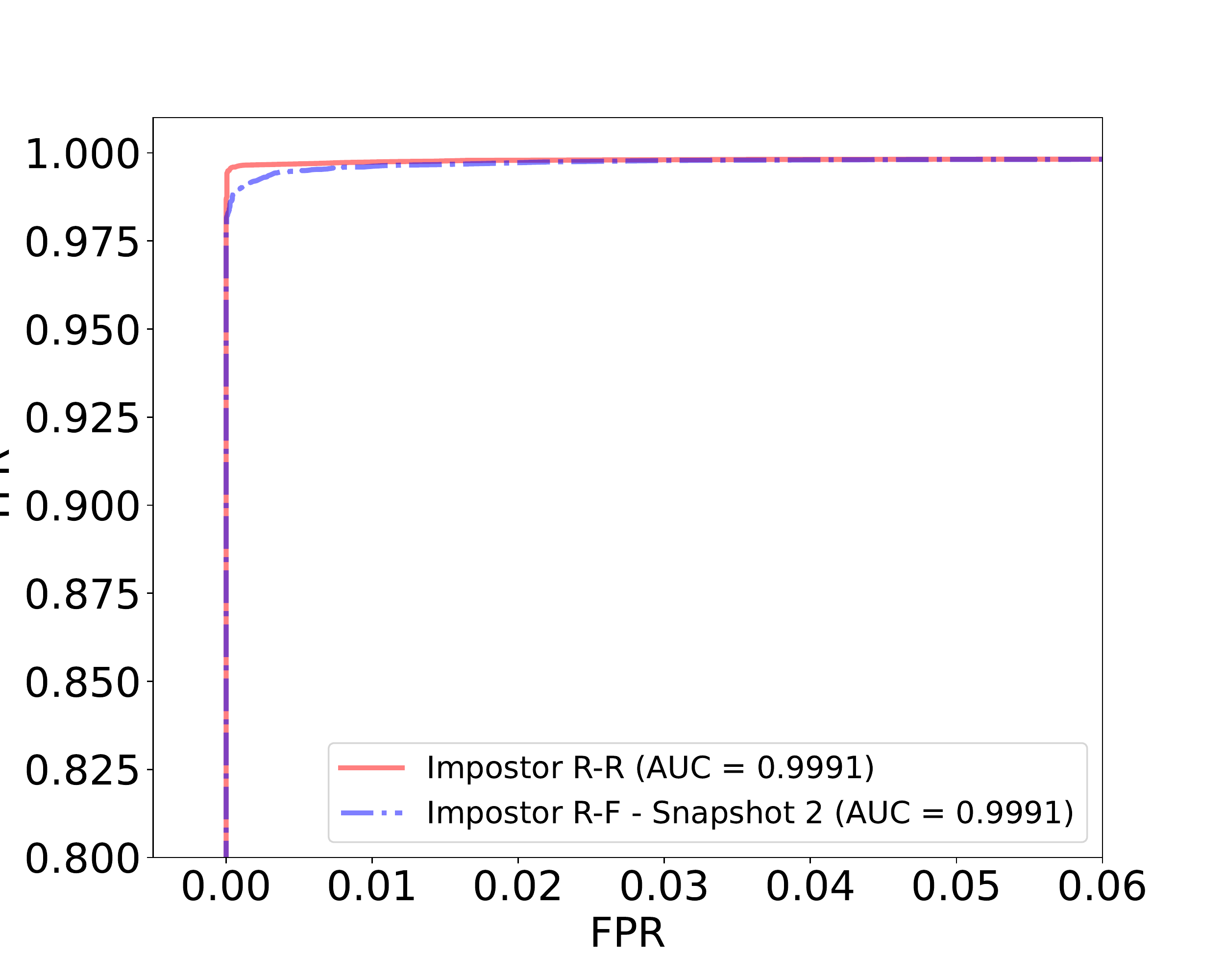}
        \caption{VeriEye}
    \end{subfigure}%
    \caption{Match score distributions for HDBSIF, USIT3, and VeriEye results for images generated at \textit{snapshot 2}.}
\end{figure*}

\begin{figure*}[h]
    \centering
    \includegraphics[width=0.9\textwidth]{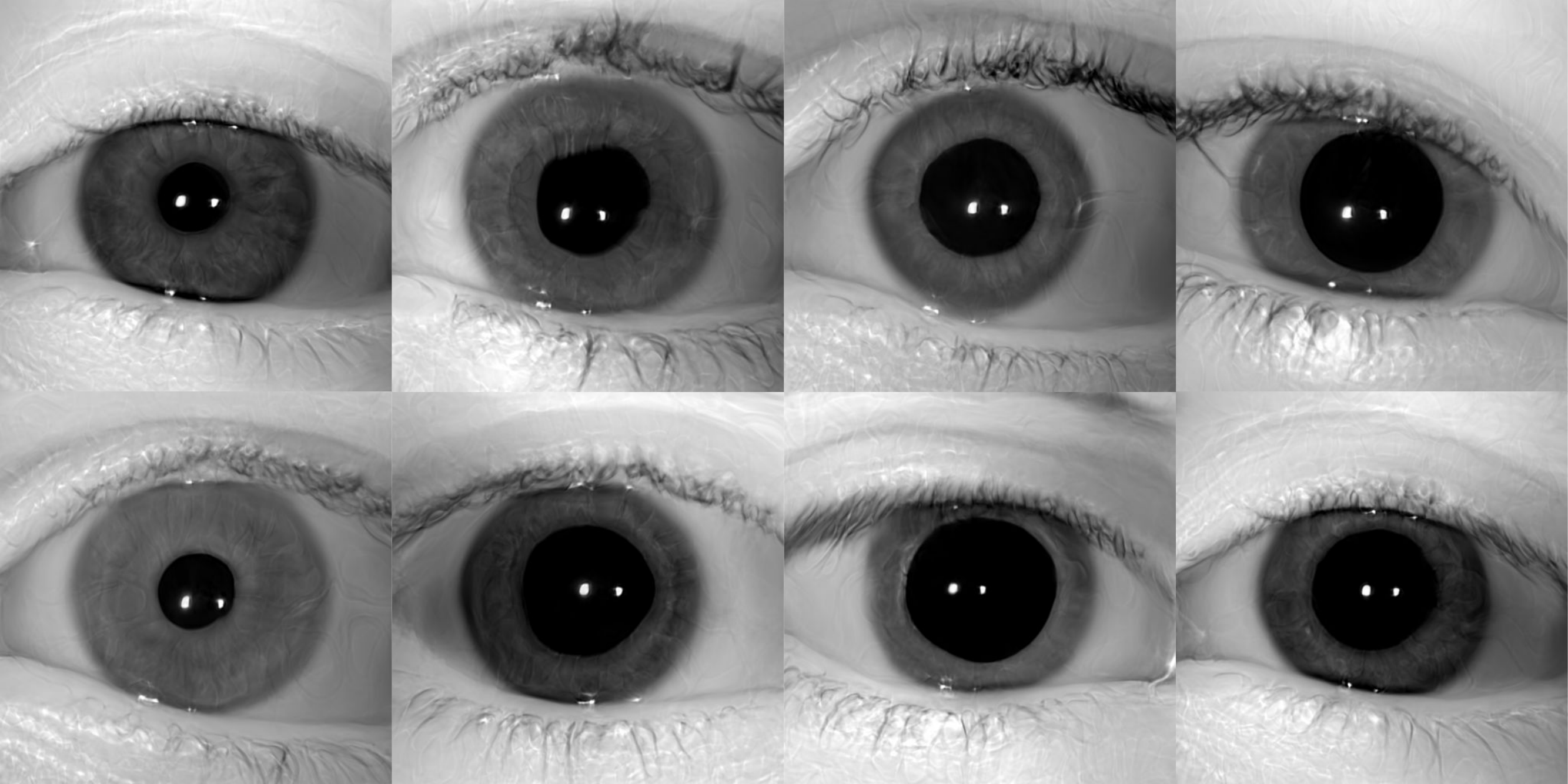}
    \caption{Image samples generated by the model at \textit{snapshot 3}.}
\end{figure*}
\begin{figure*}[h]
    \centering
    \begin{subfigure}{.31\textwidth}
        \centering
        \includegraphics[width=\textwidth]{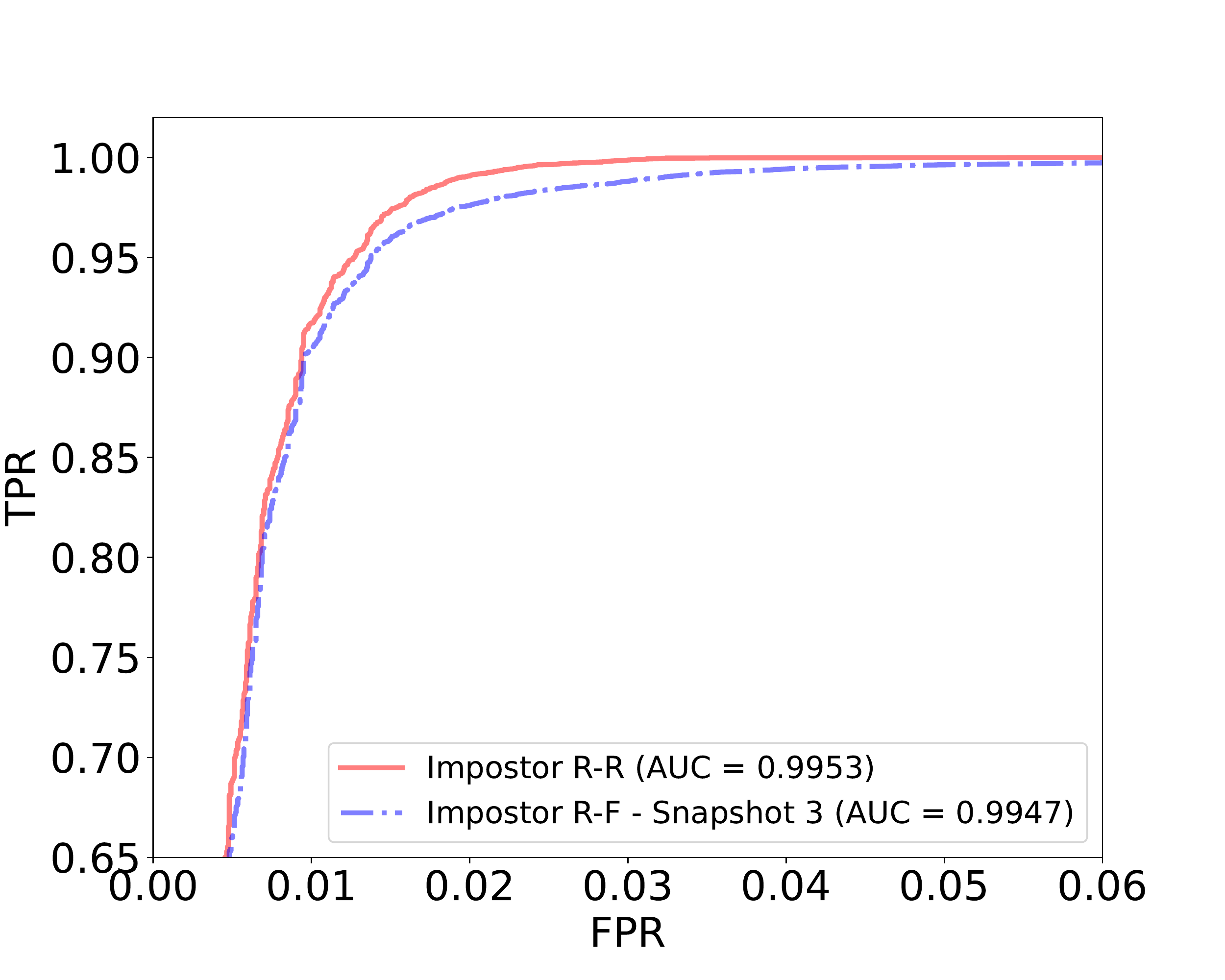}
        \caption{HDBSIF}
    \end{subfigure}%
    \begin{subfigure}{.31\textwidth}
        \centering
        \includegraphics[width=\textwidth]{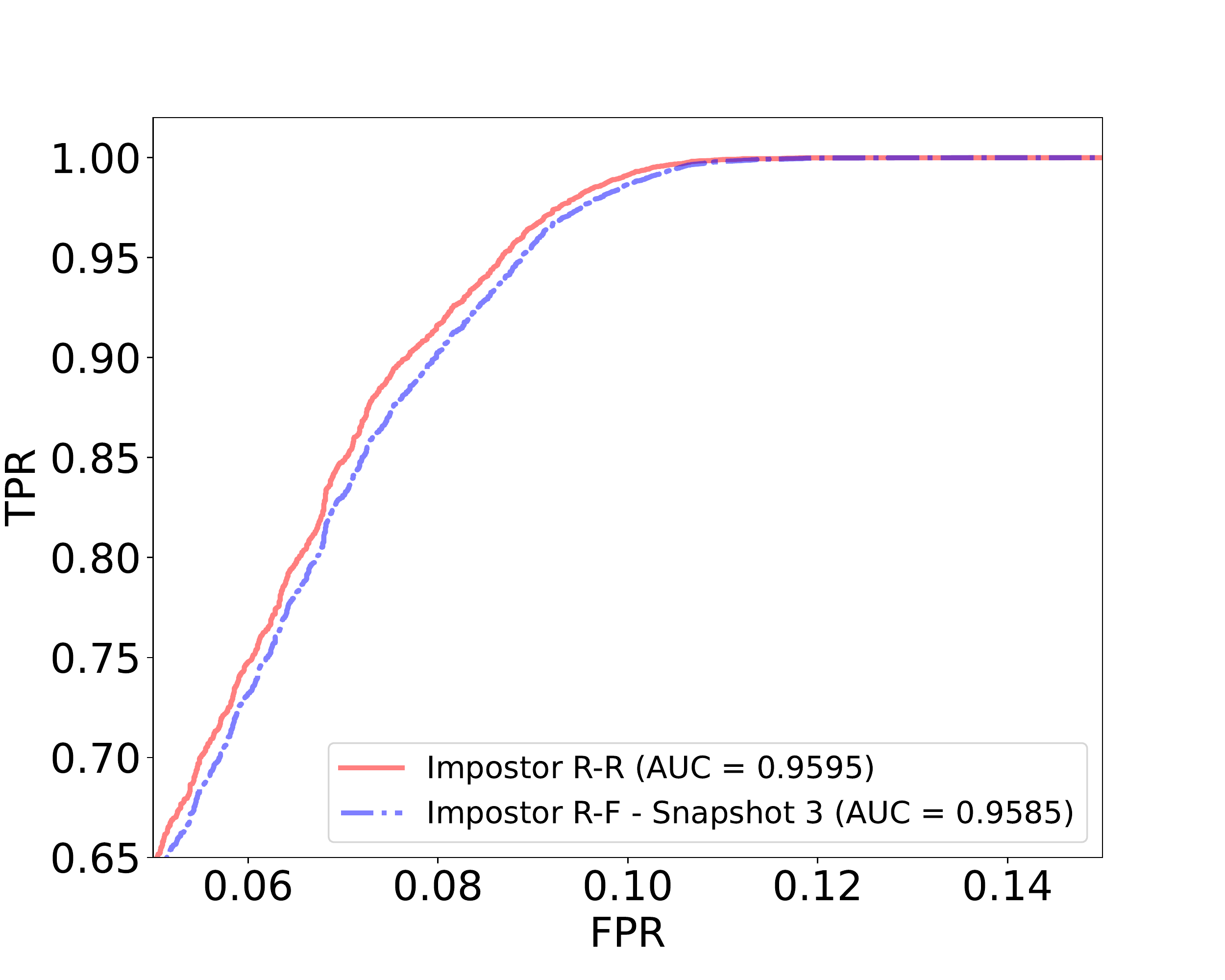}
        \caption{USIT3}
    \end{subfigure}%
    \begin{subfigure}{0.31\textwidth}
        \centering
        \includegraphics[width=\textwidth]{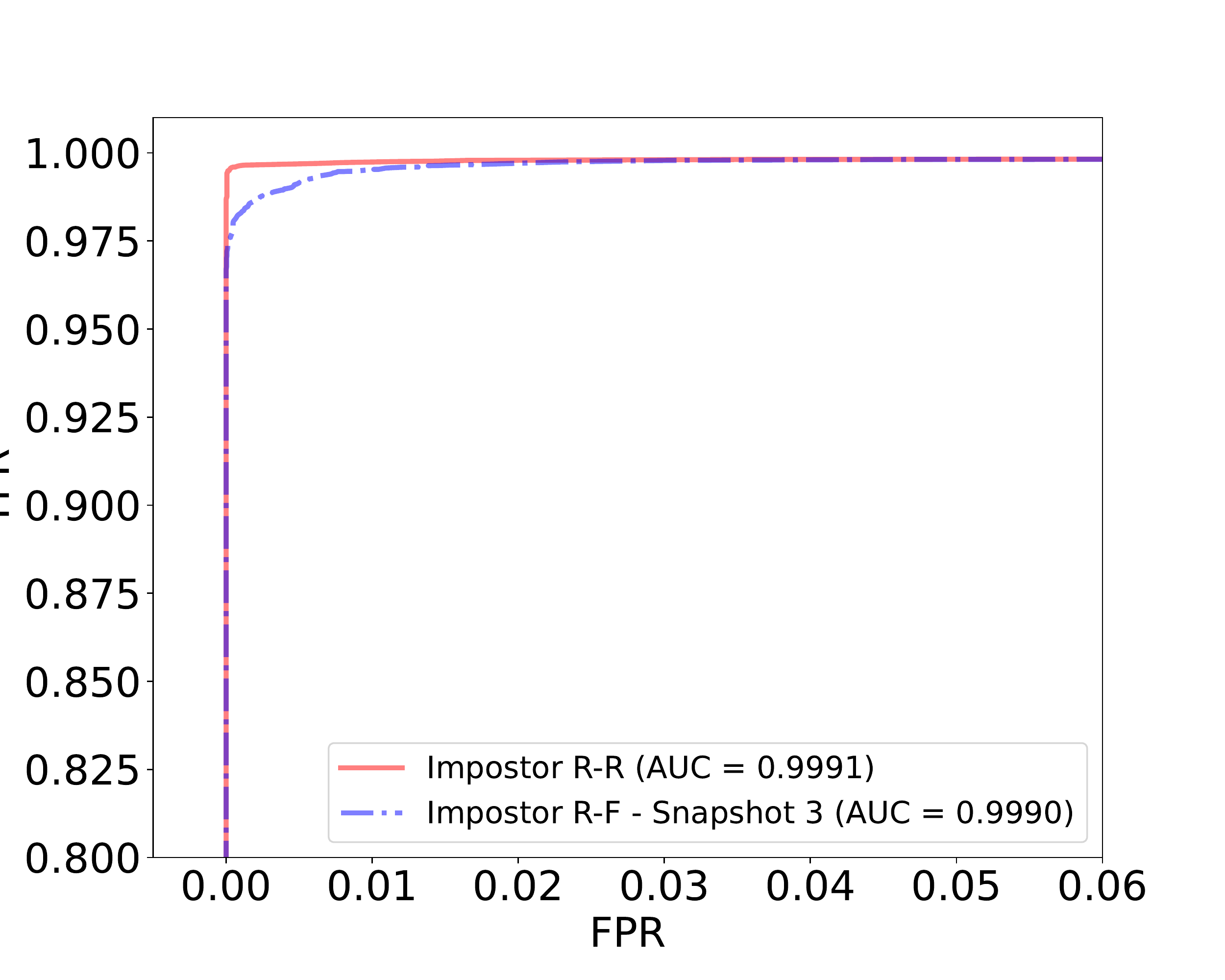}
        \caption{VeriEye}
    \end{subfigure}%
    \caption{Match score distributions for HDBSIF, USIT3, and VeriEye results for images generated at \textit{snapshot 3}.}
\end{figure*}

\begin{figure*}[h]
    \centering
    \includegraphics[width=0.9\textwidth]{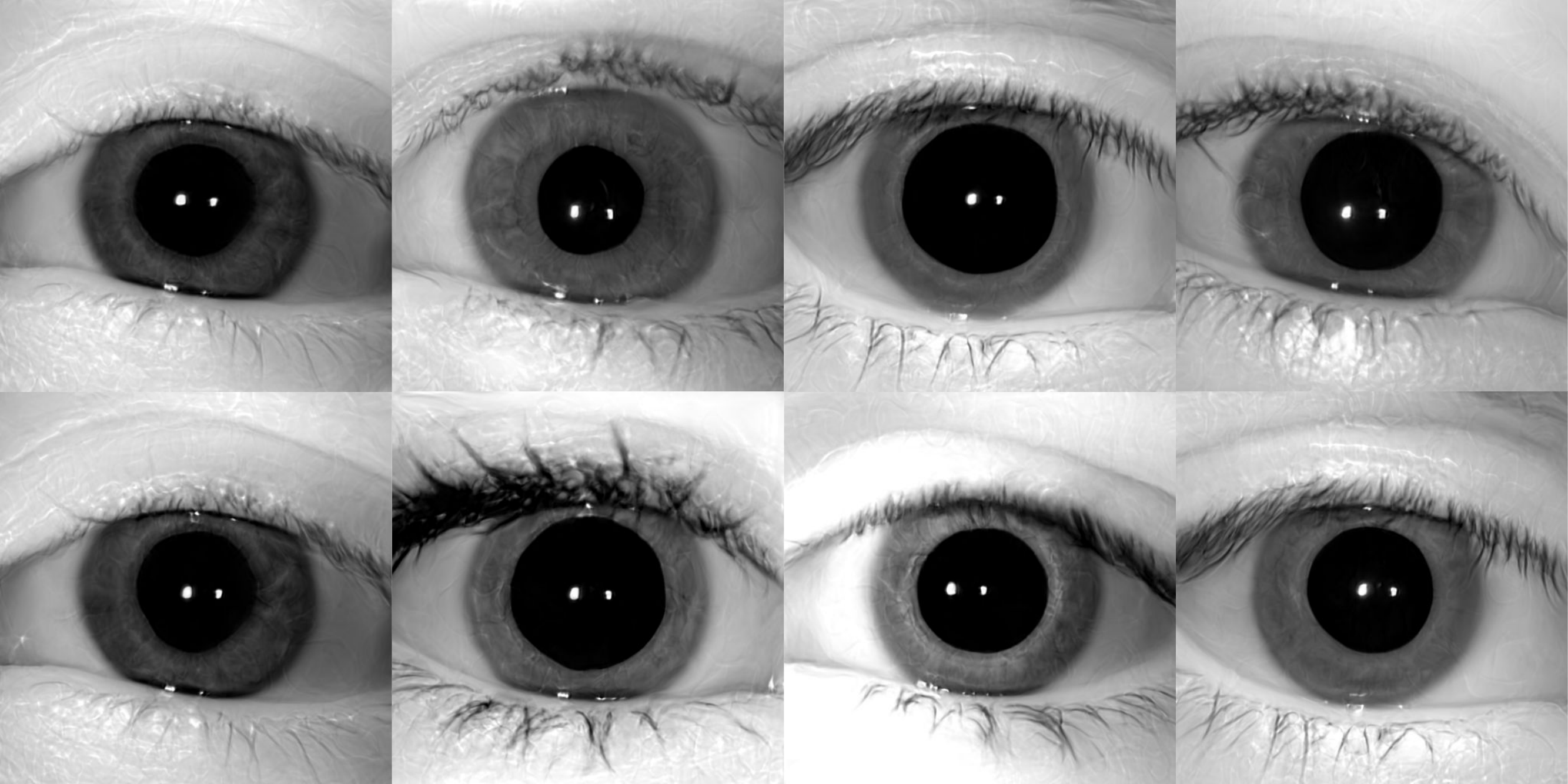}
    \caption{Image samples generated by the model at \textit{snapshot 4}.}
\end{figure*}
\begin{figure*}[h]
    \centering
    \begin{subfigure}{.31\textwidth}
        \centering
        \includegraphics[width=\textwidth]{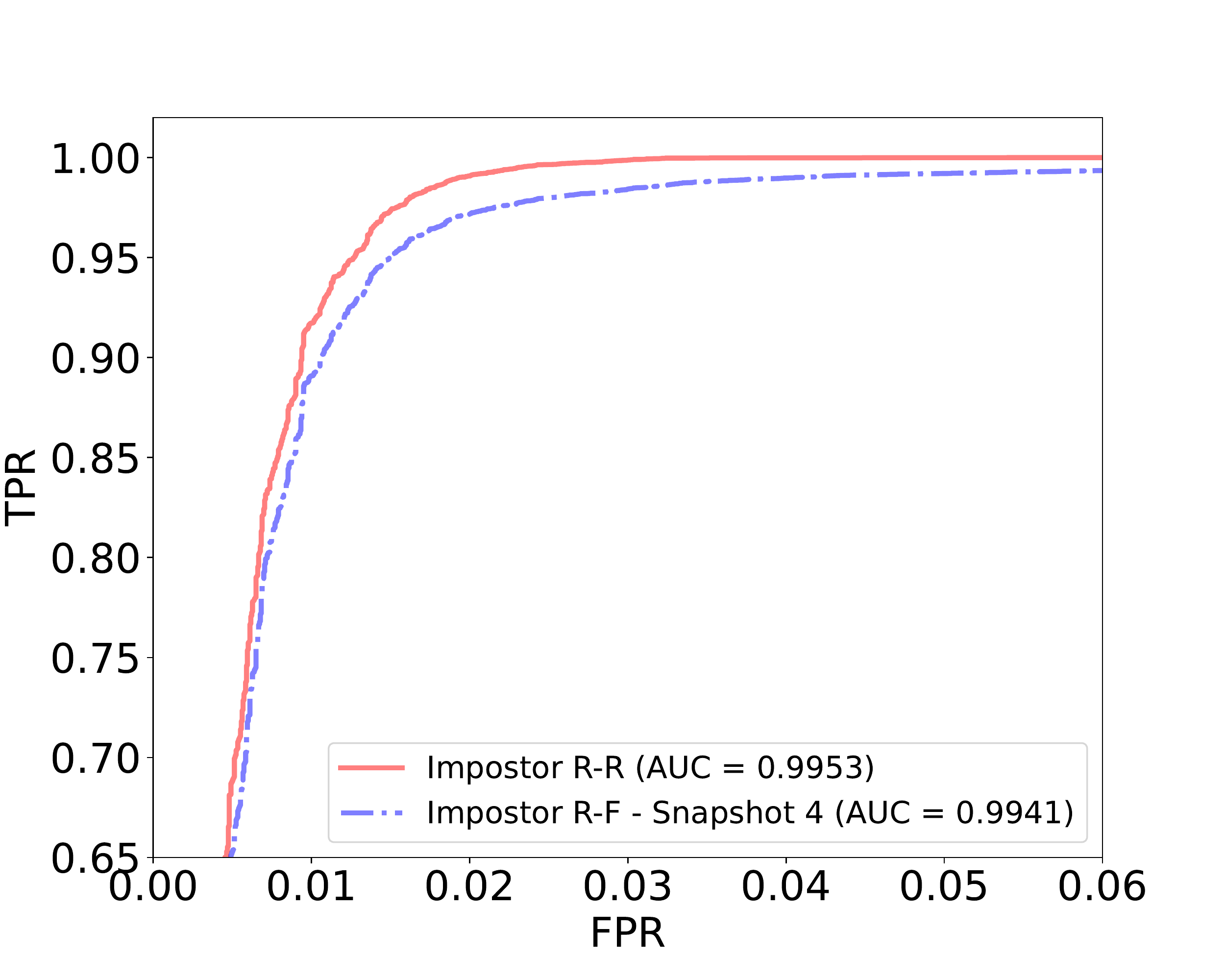}
        \caption{HDBSIF}
    \end{subfigure}%
    \begin{subfigure}{.31\textwidth}
        \centering
        \includegraphics[width=\textwidth]{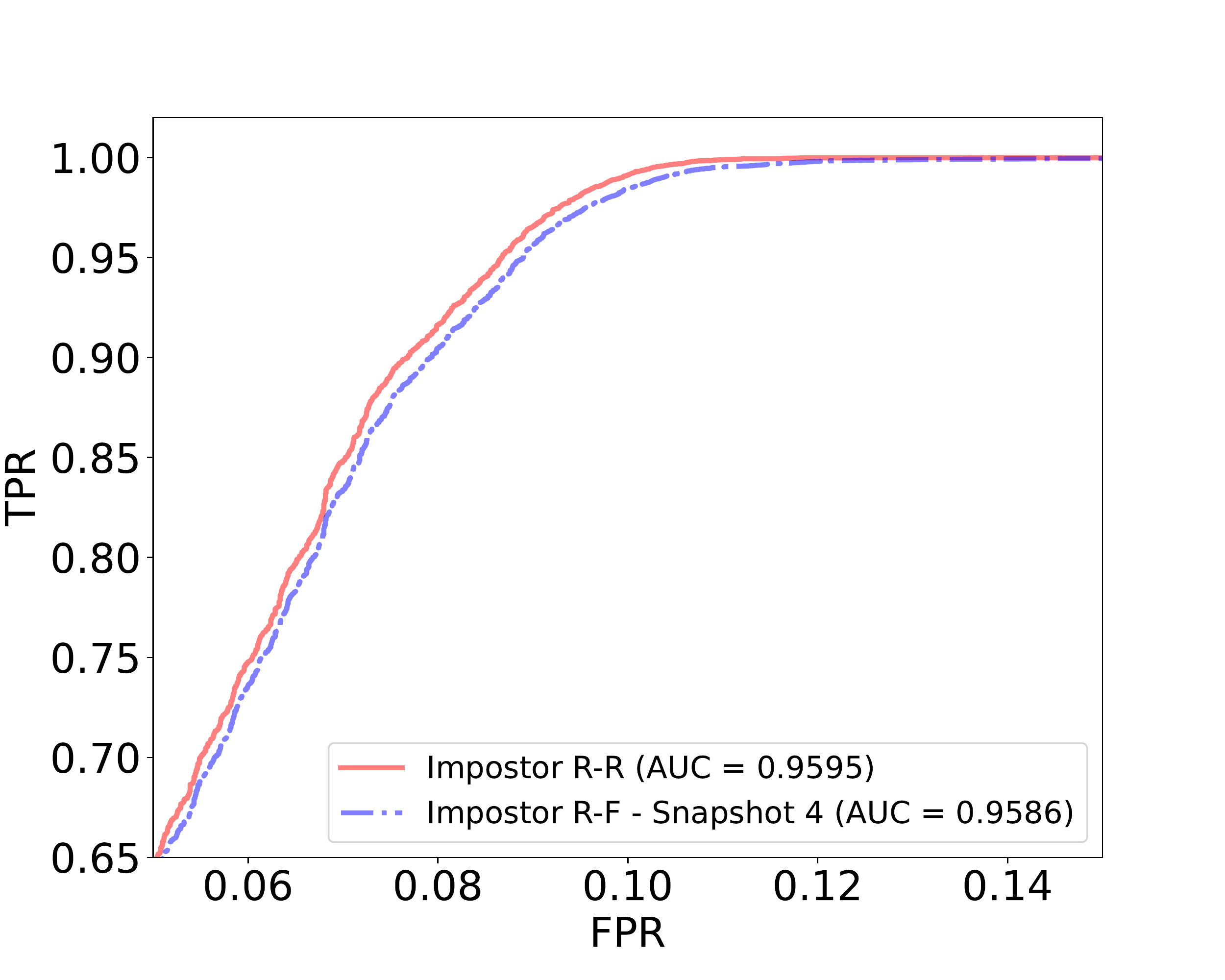}
        \caption{USIT3}
    \end{subfigure}%
    \begin{subfigure}{0.31\textwidth}
        \centering
        \includegraphics[width=\textwidth]{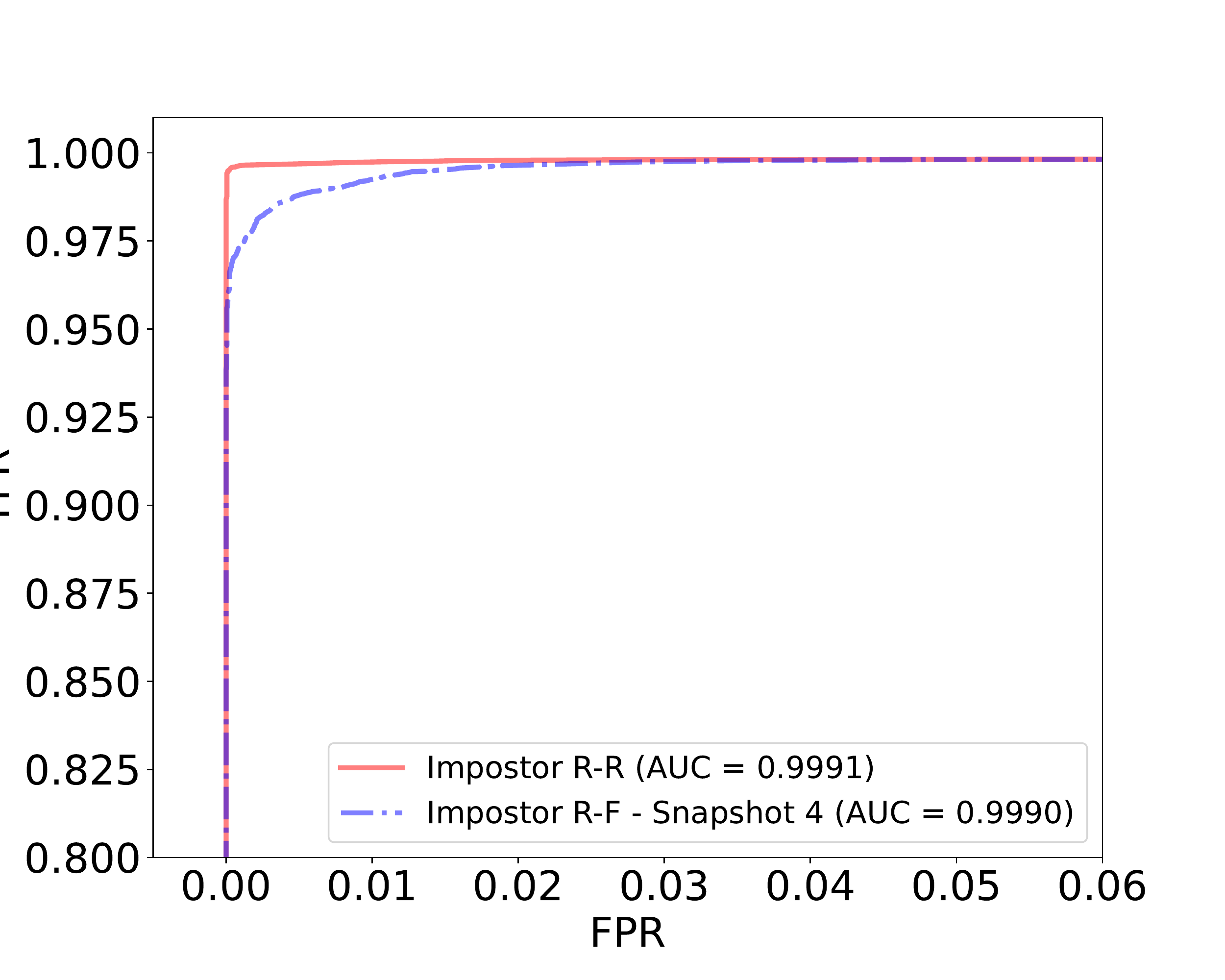}
        \caption{VeriEye}
    \end{subfigure}%
    \caption{Match score distributions for HDBSIF, USIT3, and VeriEye results for images generated at \textit{snapshot 4}.}
\end{figure*}

\begin{figure*}[h]
    \centering
    \includegraphics[width=0.9\textwidth]{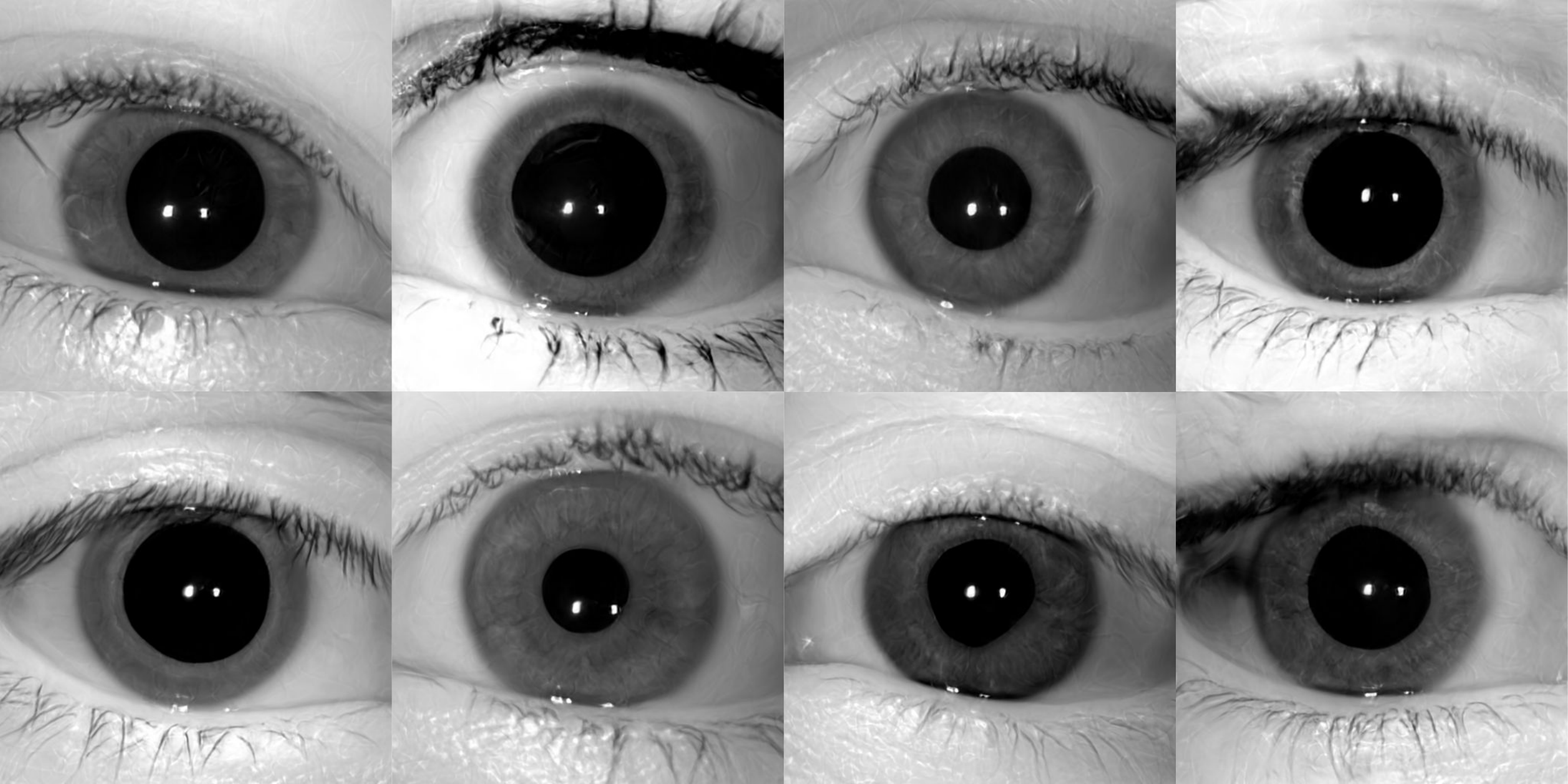}
    \caption{Image samples generated by the model at \textit{snapshot 5}.}
\end{figure*}
\begin{figure*}[h]
    \centering
    \begin{subfigure}{.31\textwidth}
        \centering
        \includegraphics[width=\textwidth]{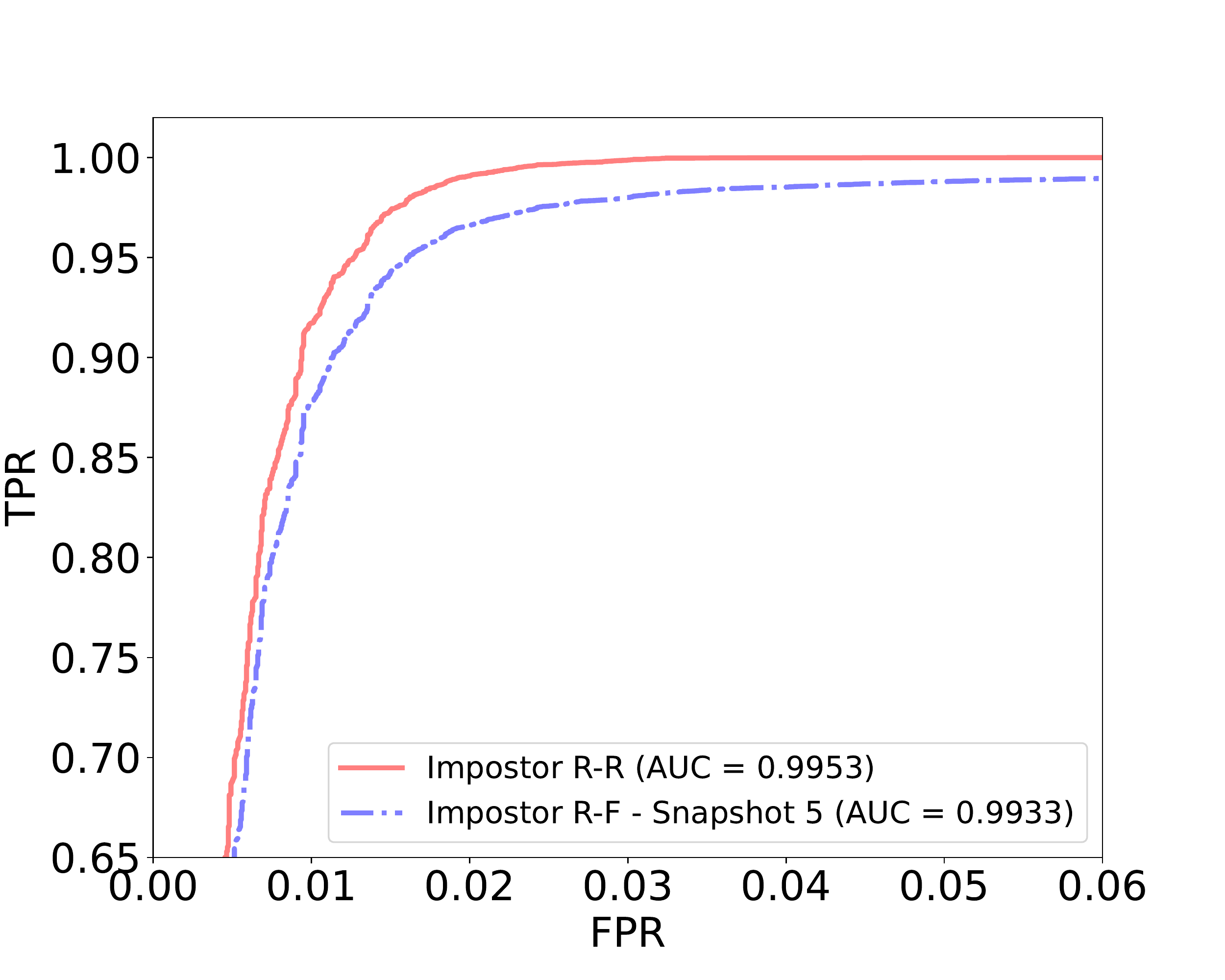}
        \caption{HDBSIF}
    \end{subfigure}%
    \begin{subfigure}{.31\textwidth}
        \centering
        \includegraphics[width=\textwidth]{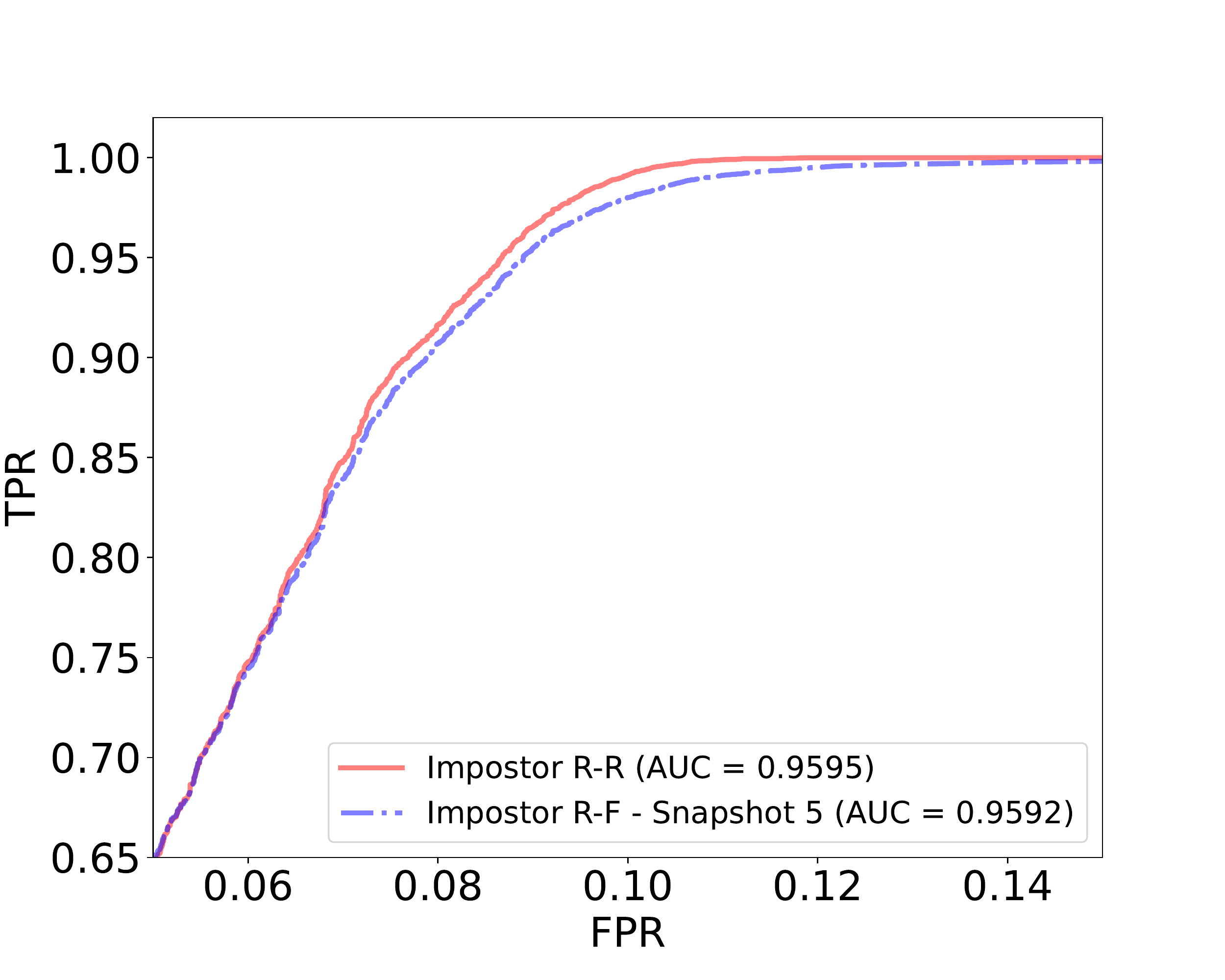}
        \caption{USIT3}
    \end{subfigure}%
    \begin{subfigure}{0.31\textwidth}
        \centering
        \includegraphics[width=\textwidth]{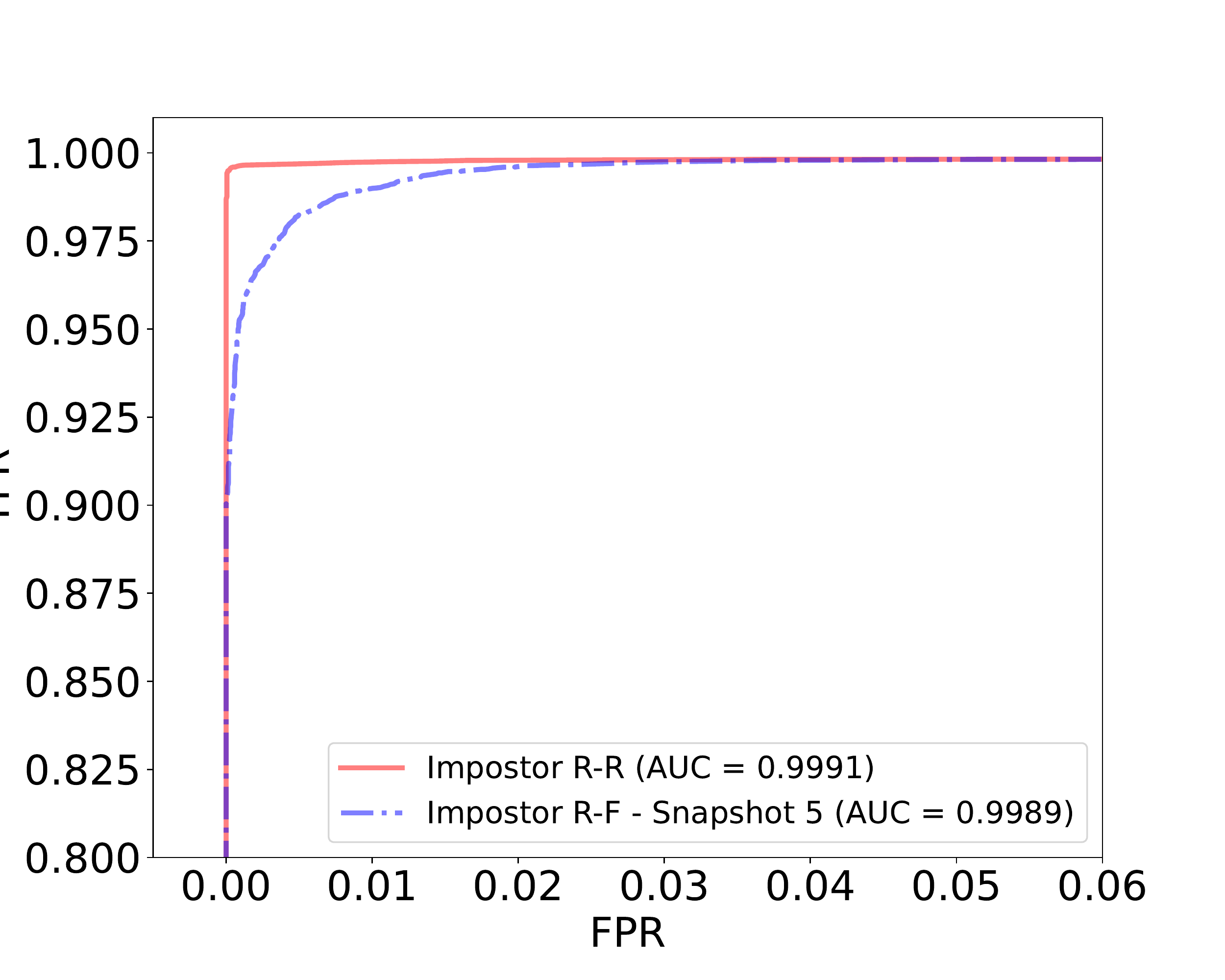}
        \caption{VeriEye}
    \end{subfigure}%
    \caption{Match score distributions for HDBSIF, USIT3, and VeriEye results for images generated at \textit{snapshot 5}.}
\end{figure*}

\begin{figure*}[h]
    \centering
    \includegraphics[width=0.9\textwidth]{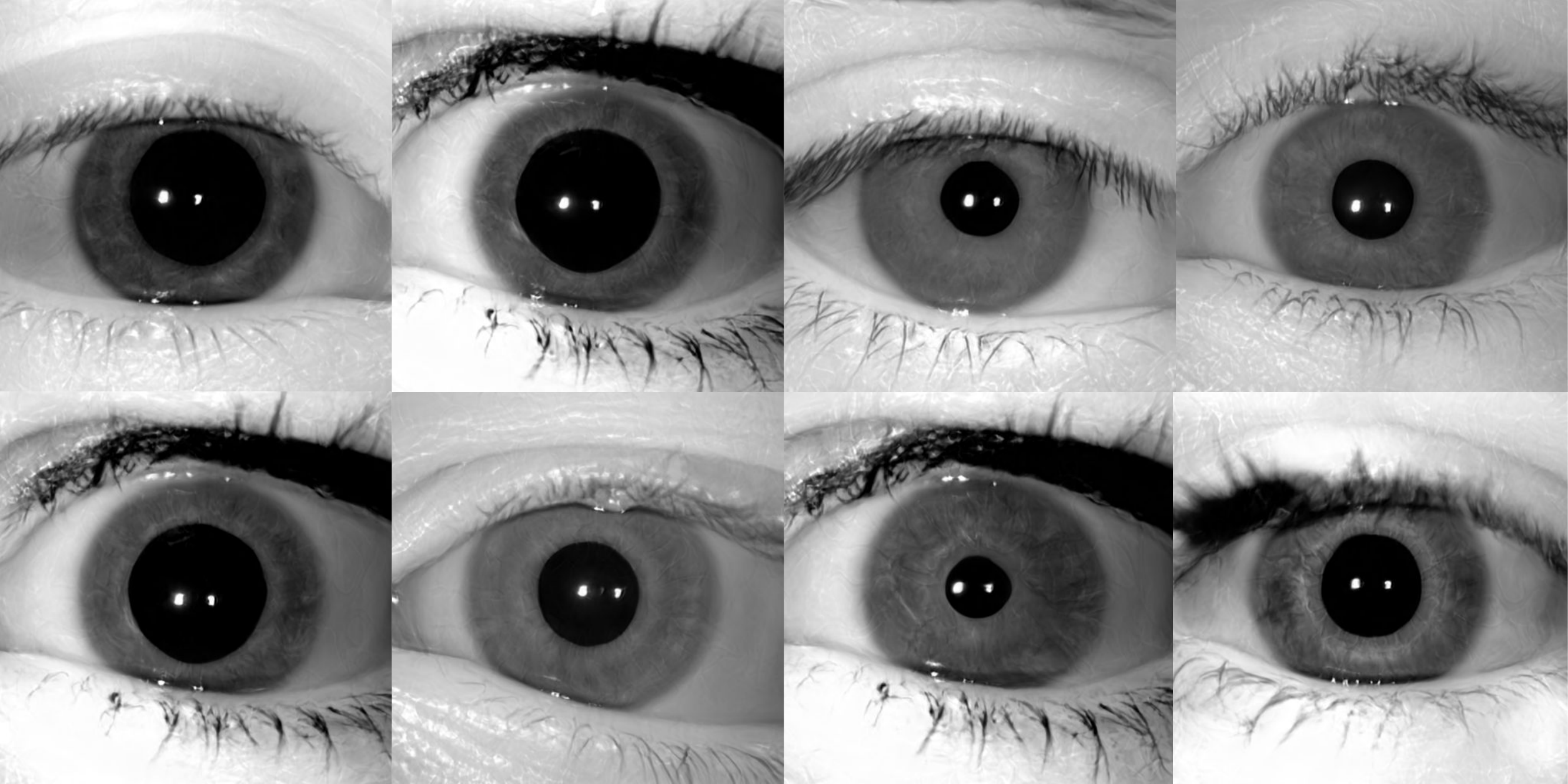}
    \caption{Image samples generated by the model at \textit{snapshot 6}.}
\end{figure*}
\begin{figure*}[h]
    \centering
    \begin{subfigure}{.31\textwidth}
        \centering
        \includegraphics[width=\textwidth]{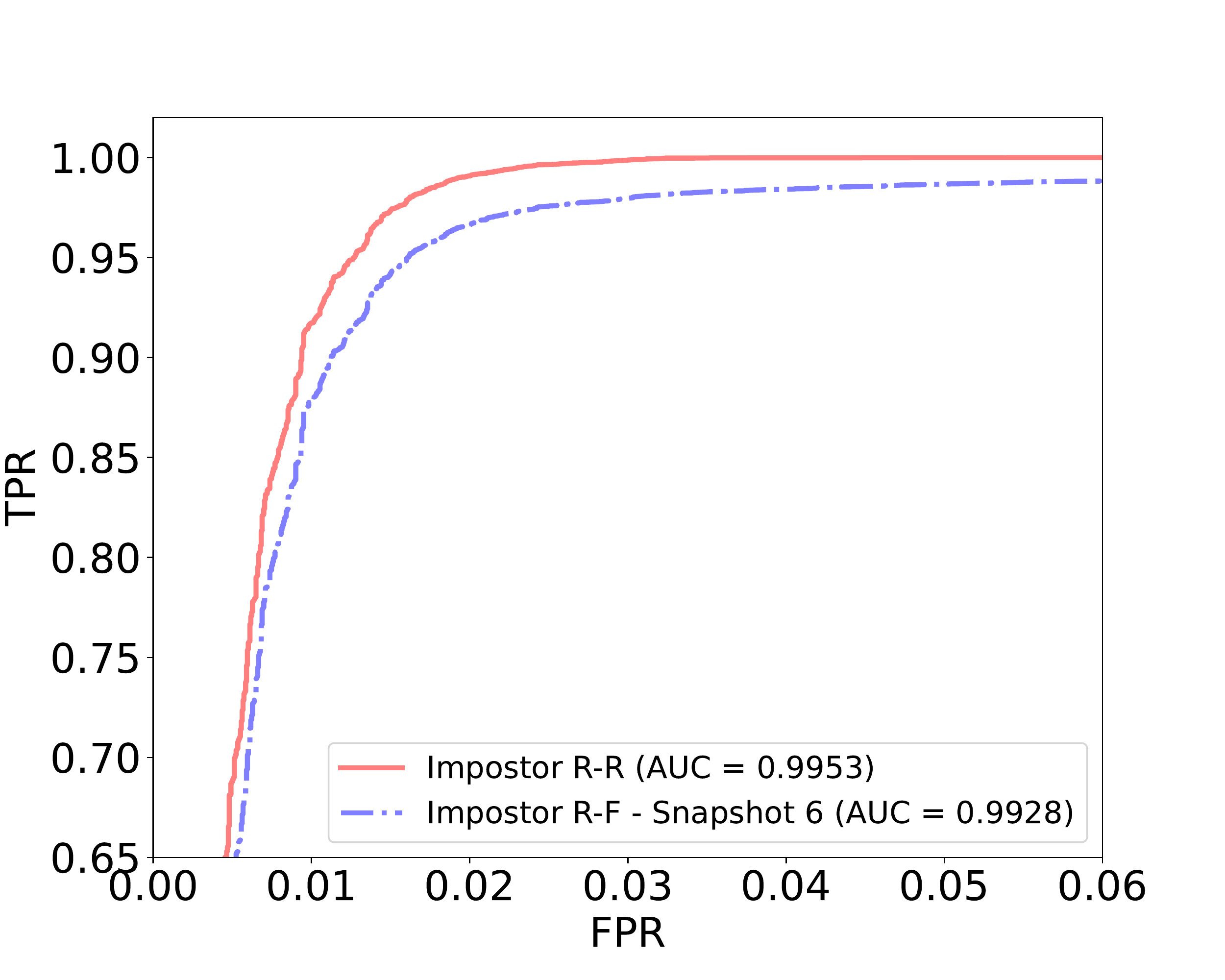}
        \caption{HDBSIF}
    \end{subfigure}%
    \begin{subfigure}{.31\textwidth}
        \centering
        \includegraphics[width=\textwidth]{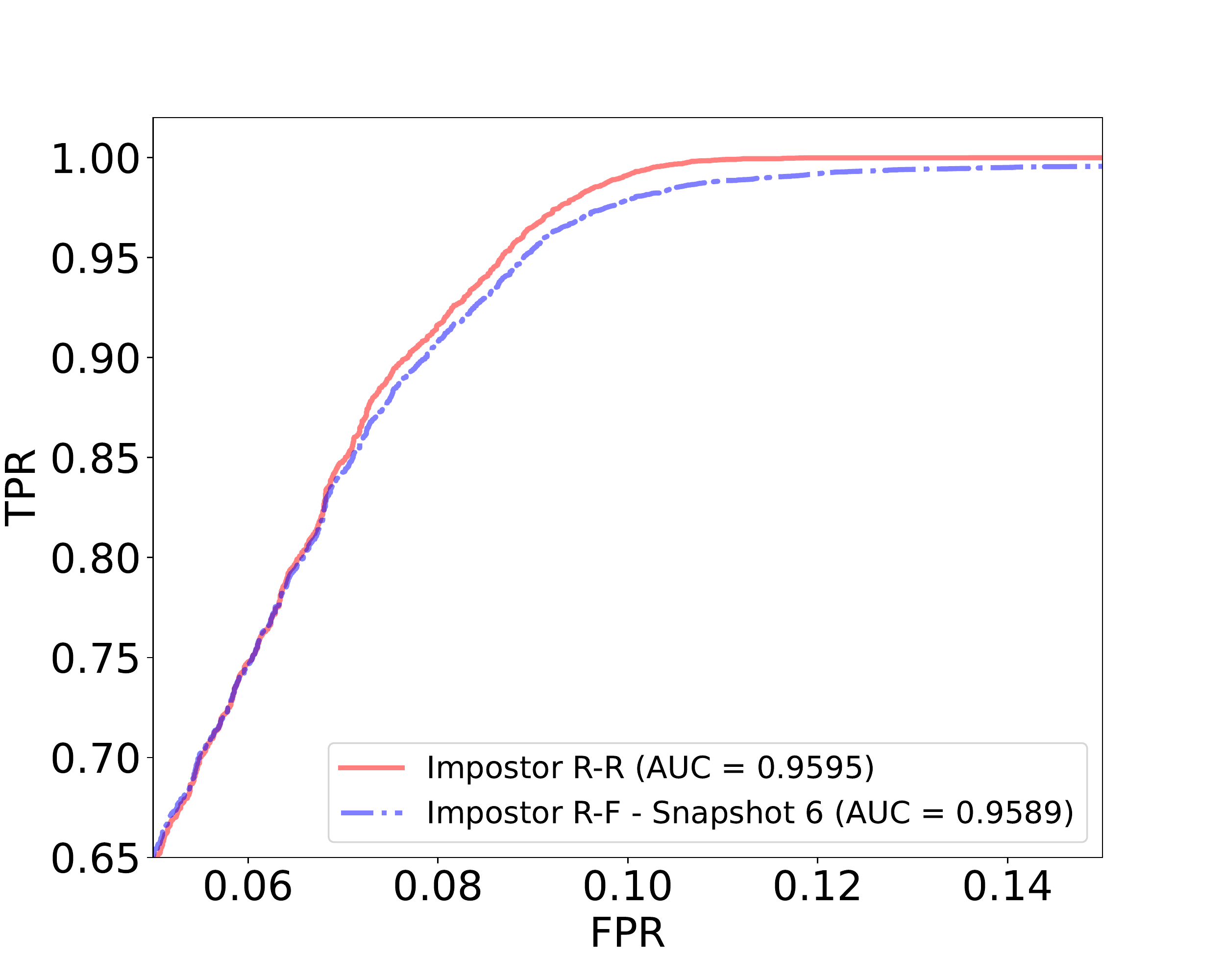}
        \caption{USIT3}
    \end{subfigure}%
    \begin{subfigure}{0.31\textwidth}
        \centering
        \includegraphics[width=\textwidth]{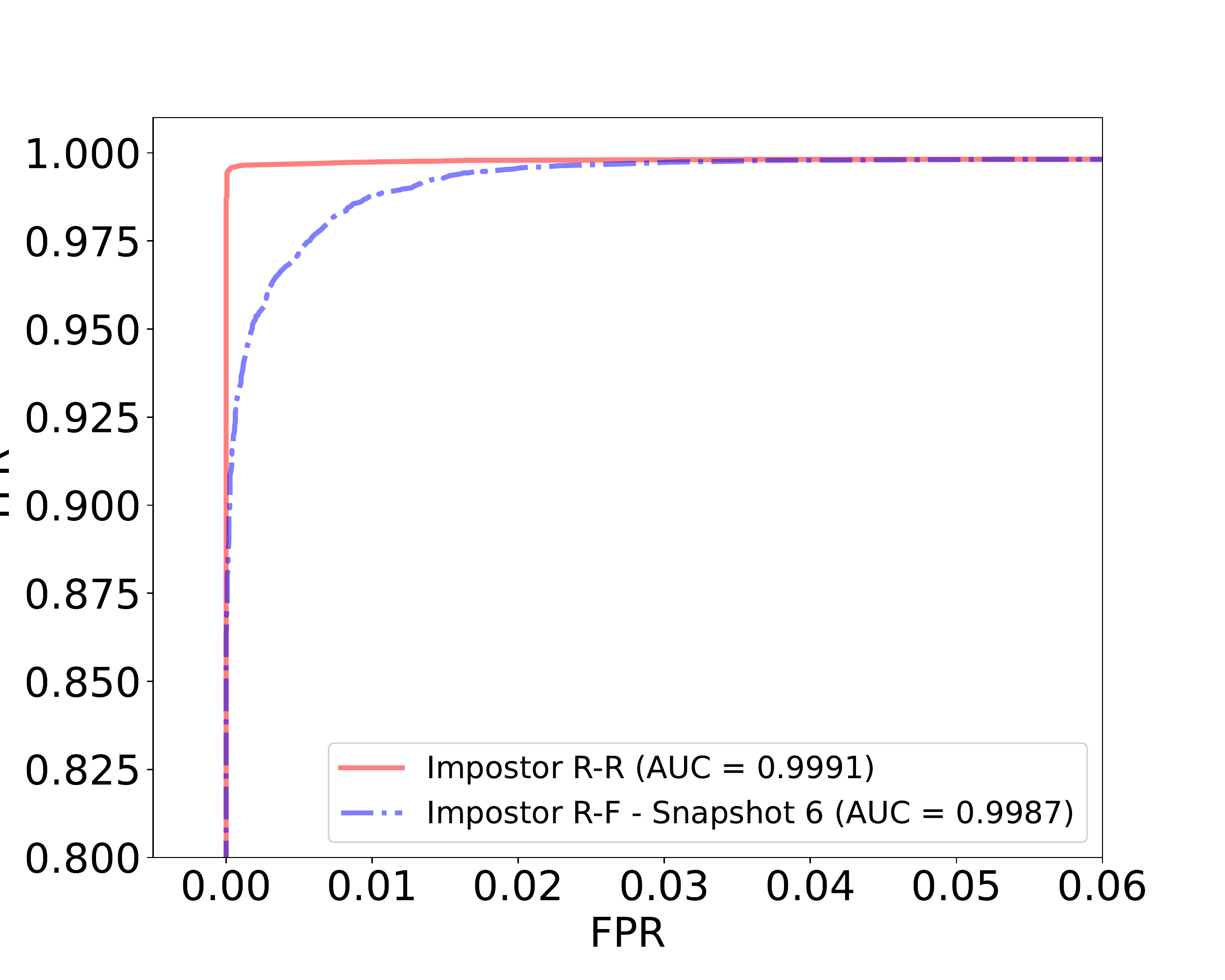}
        \caption{VeriEye}
    \end{subfigure}%
    \caption{Match score distributions for HDBSIF, USIT3, and VeriEye results for images generated at \textit{snapshot 6}.}
\end{figure*}

\begin{figure*}[h]
    \centering
    \includegraphics[width=0.9\textwidth]{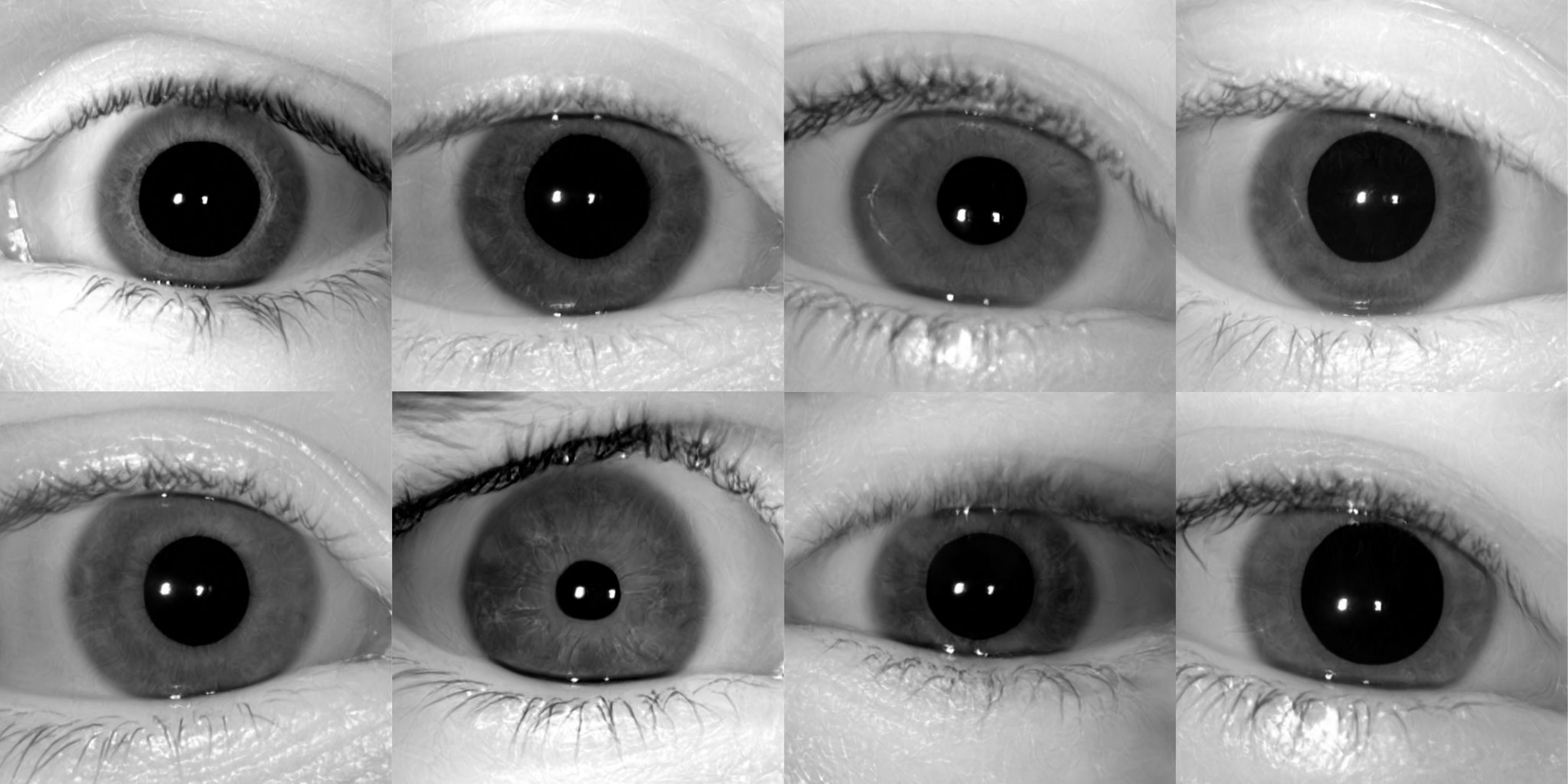}
    \caption{Image samples generated by the model at \textit{snapshot 7}.}
\end{figure*}
\begin{figure*}[h]
    \centering
    \begin{subfigure}{.31\textwidth}
        \centering
        \includegraphics[width=\textwidth]{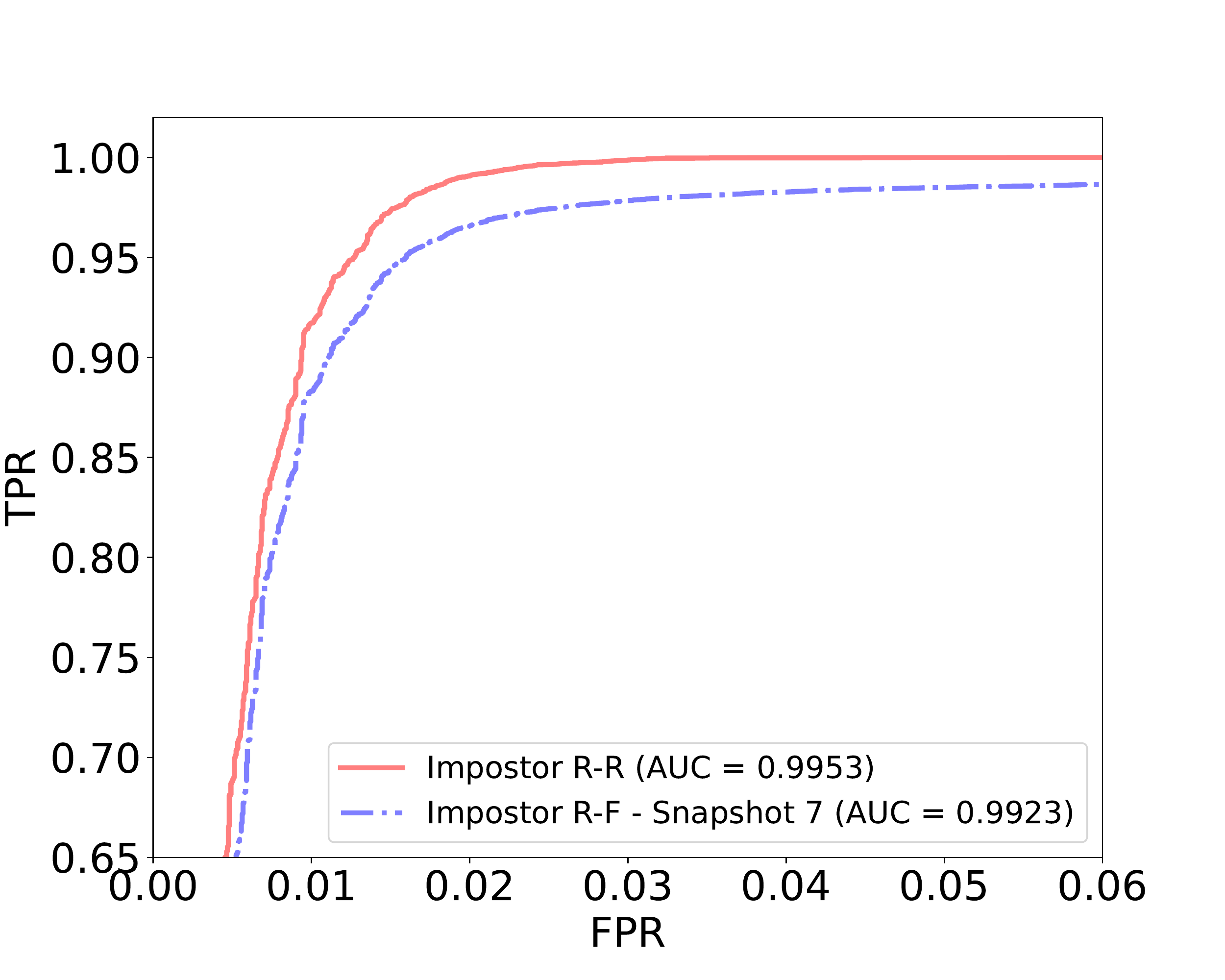}
        \caption{HDBSIF}
    \end{subfigure}%
    \begin{subfigure}{.31\textwidth}
        \centering
        \includegraphics[width=\textwidth]{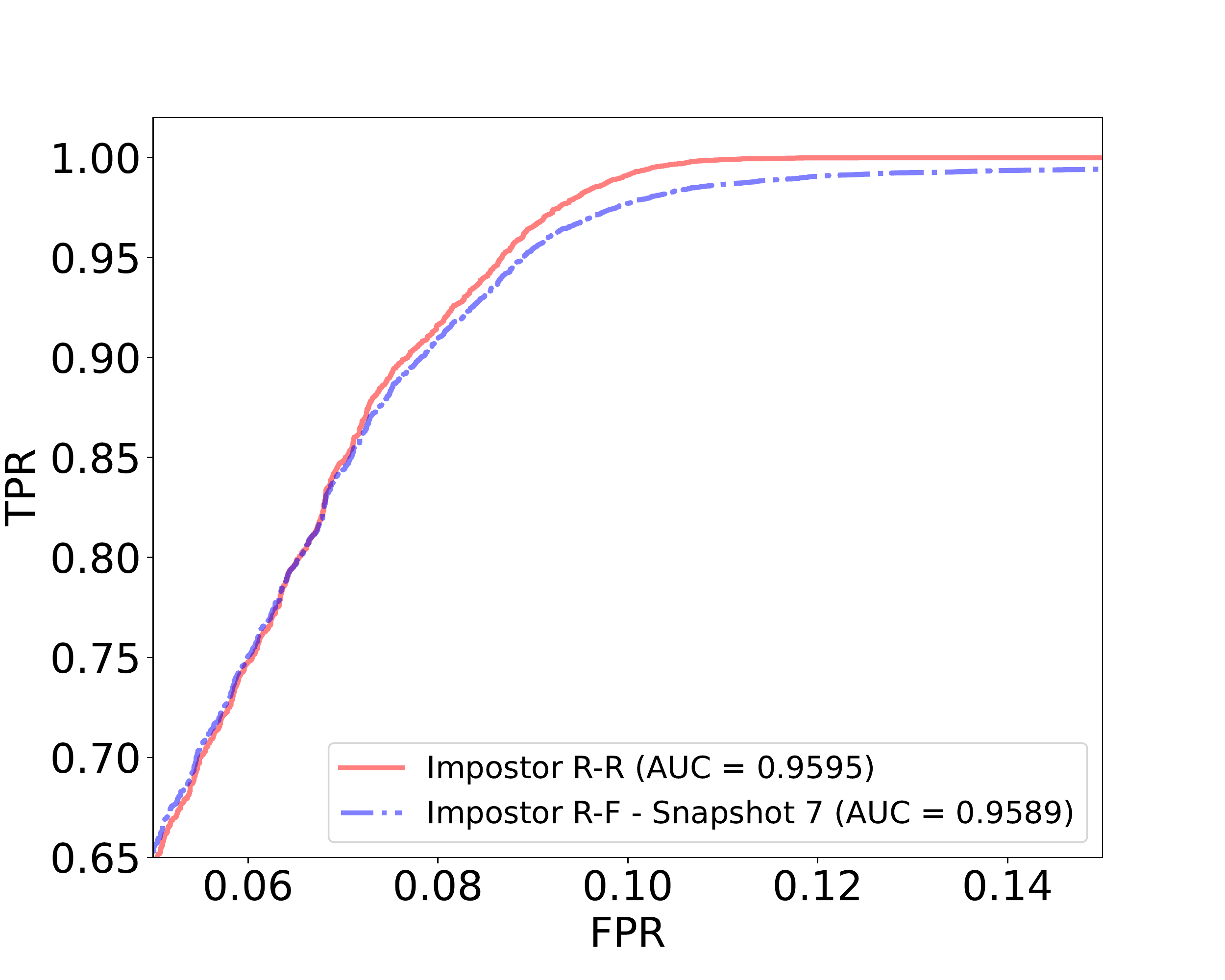}
        \caption{USIT3}
    \end{subfigure}%
    \begin{subfigure}{0.31\textwidth}
        \centering
        \includegraphics[width=\textwidth]{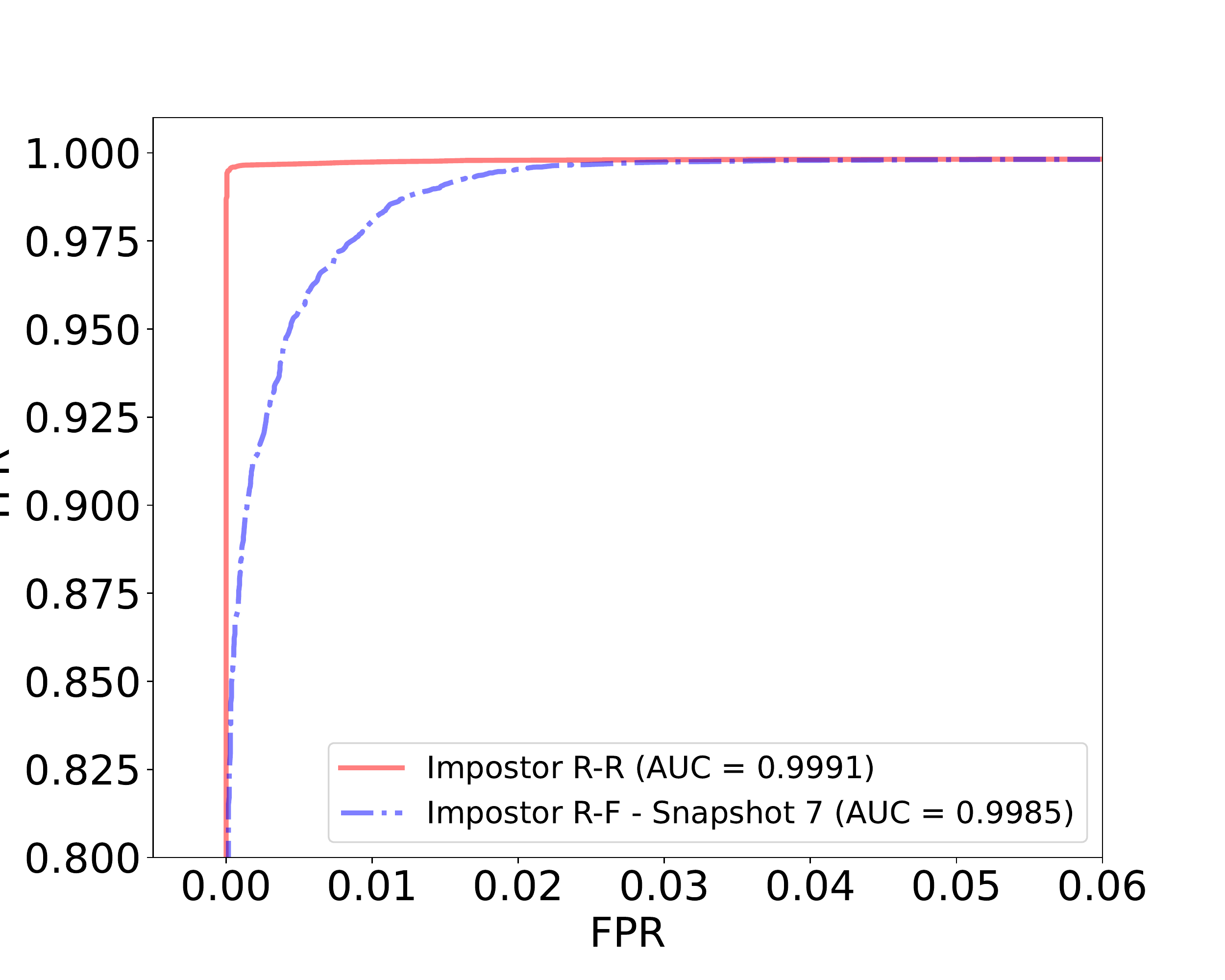}
        \caption{VeriEye}
    \end{subfigure}%
    \caption{Match score distributions for HDBSIF, USIT3, and VeriEye results for images generated at \textit{snapshot 7}.}
\end{figure*}

\begin{figure*}[h]
    \centering
    \includegraphics[width=0.9\textwidth]{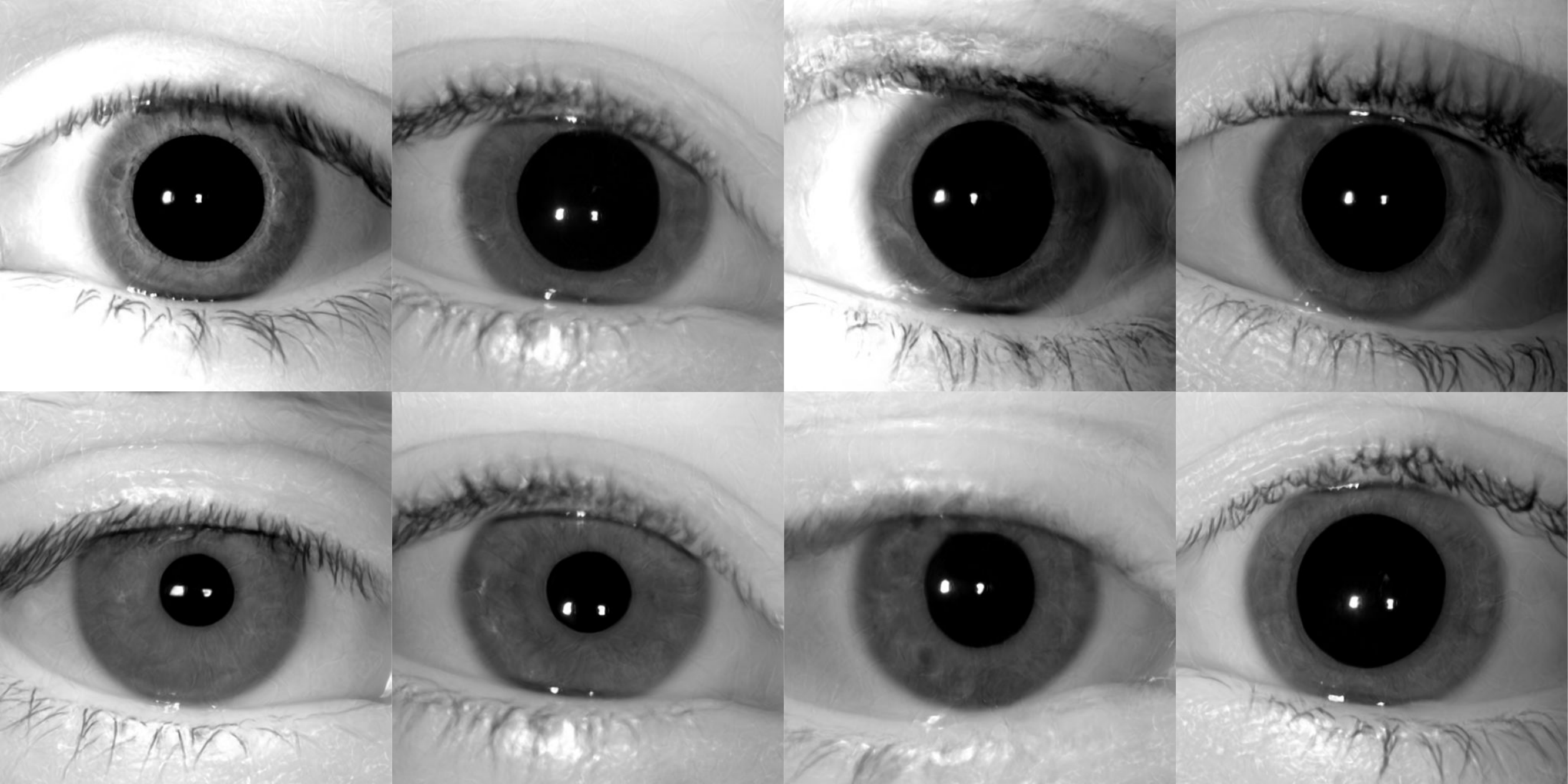}
    \caption{Image samples generated by the model at \textit{snapshot 8}.}
\end{figure*}
\begin{figure*}[h]
    \centering
    \begin{subfigure}{.31\textwidth}
        \centering
        \includegraphics[width=\textwidth]{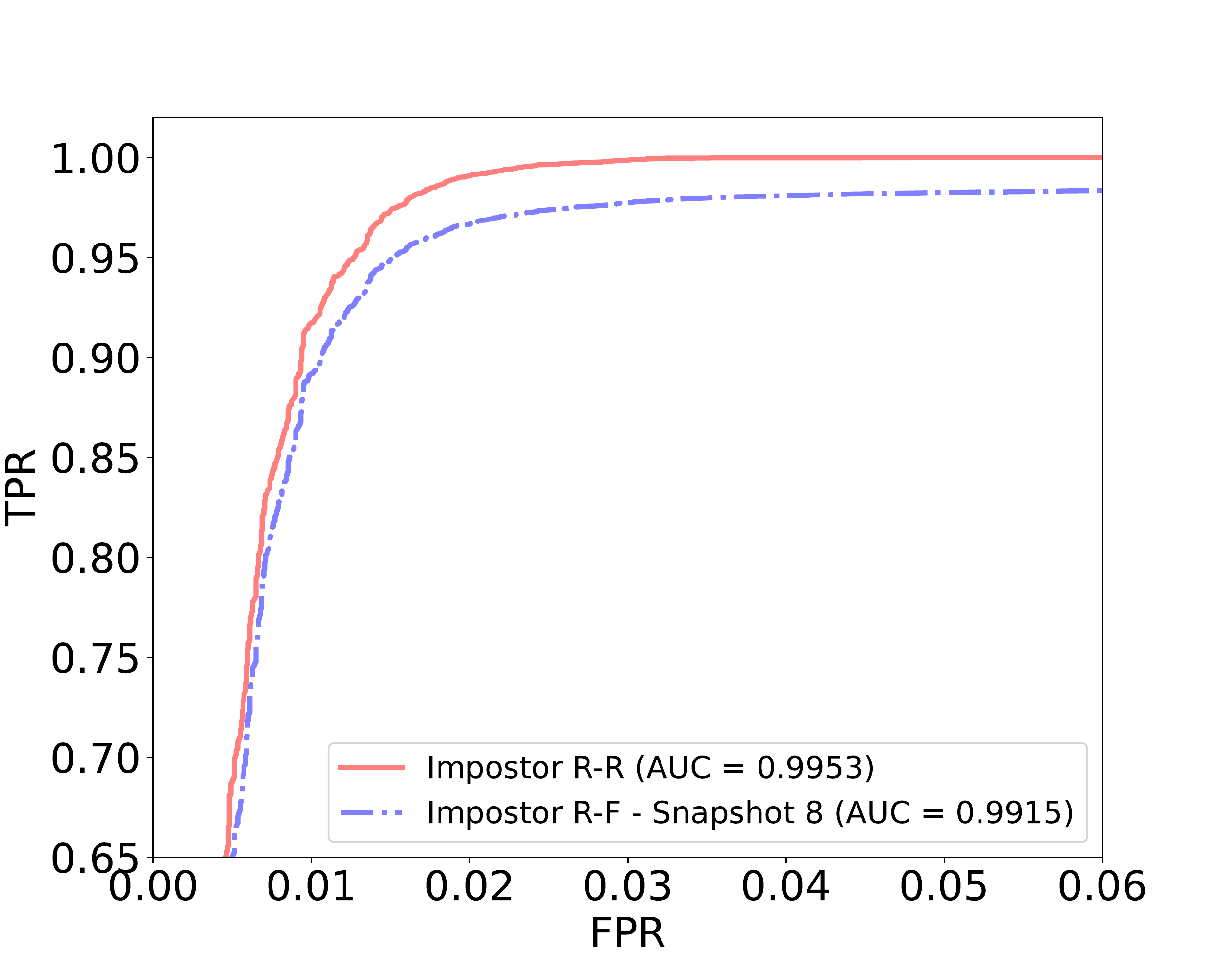}
        \caption{HDBSIF}
    \end{subfigure}%
    \begin{subfigure}{.31\textwidth}
        \centering
        \includegraphics[width=\textwidth]{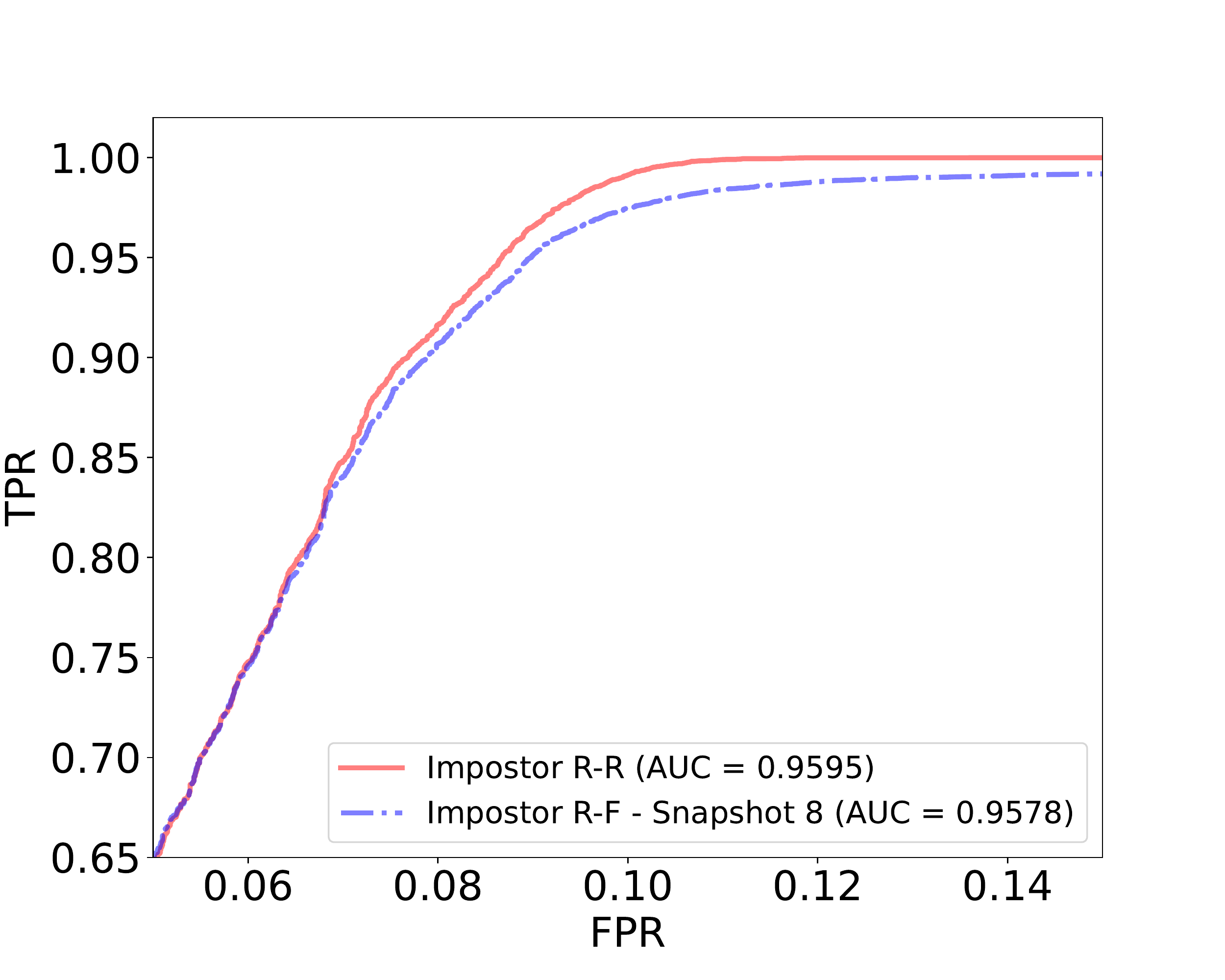}
        \caption{USIT3}
    \end{subfigure}%
    \begin{subfigure}{0.31\textwidth}
        \centering
        \includegraphics[width=\textwidth]{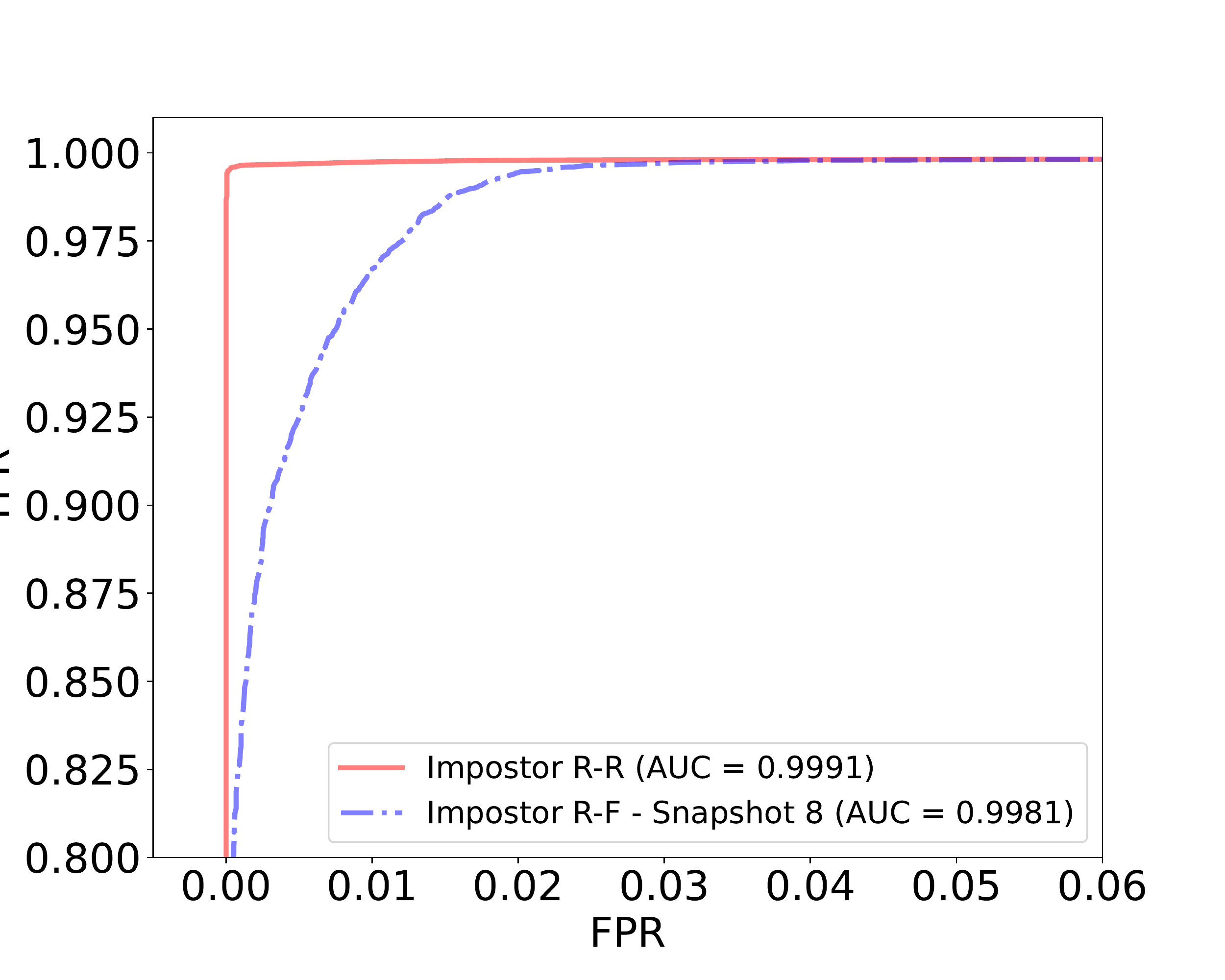}
        \caption{VeriEye}
    \end{subfigure}%
    \caption{Match score distributions for HDBSIF, USIT3, and VeriEye results for images generated at \textit{snapshot 8}.}
\end{figure*}

\begin{figure*}[h]
    \centering
    \includegraphics[width=0.9\textwidth]{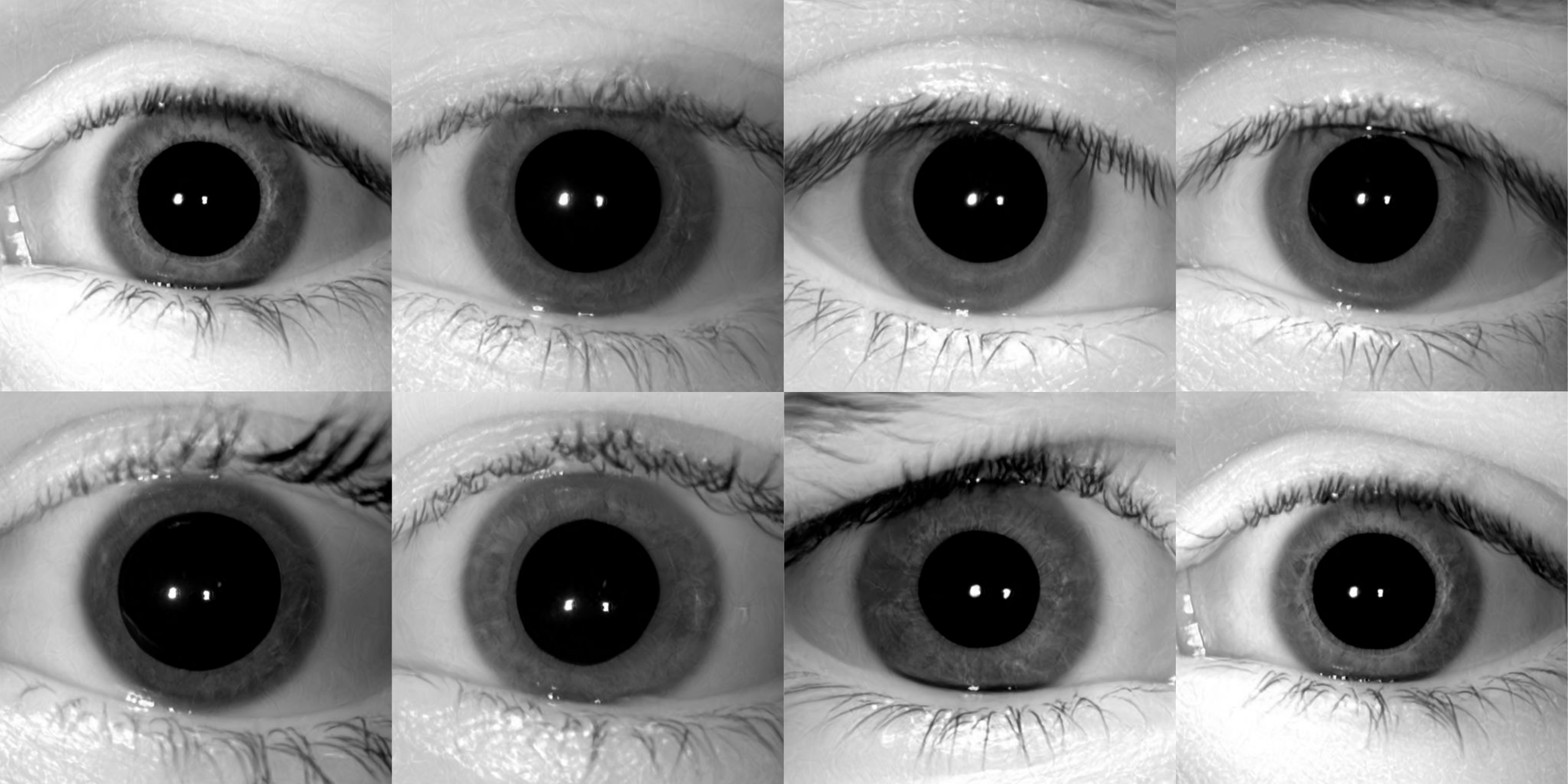}
    \caption{Image samples generated by the model at \textit{snapshot 9}.}
\end{figure*}
\begin{figure*}[h]
    \centering
    \begin{subfigure}{.31\textwidth}
        \centering
        \includegraphics[width=\textwidth]{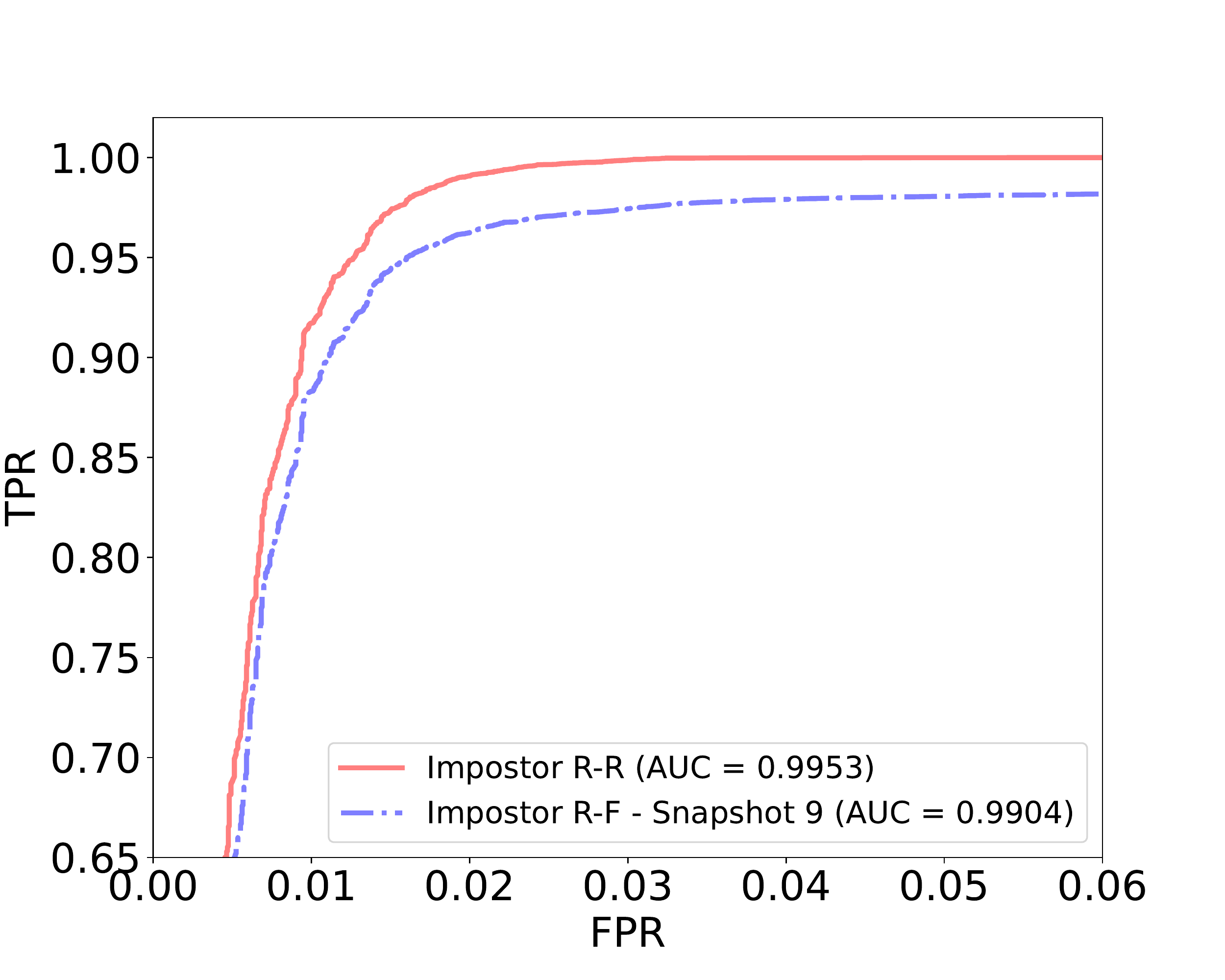}
        \caption{HDBSIF}
    \end{subfigure}%
    \begin{subfigure}{.31\textwidth}
        \centering
        \includegraphics[width=\textwidth]{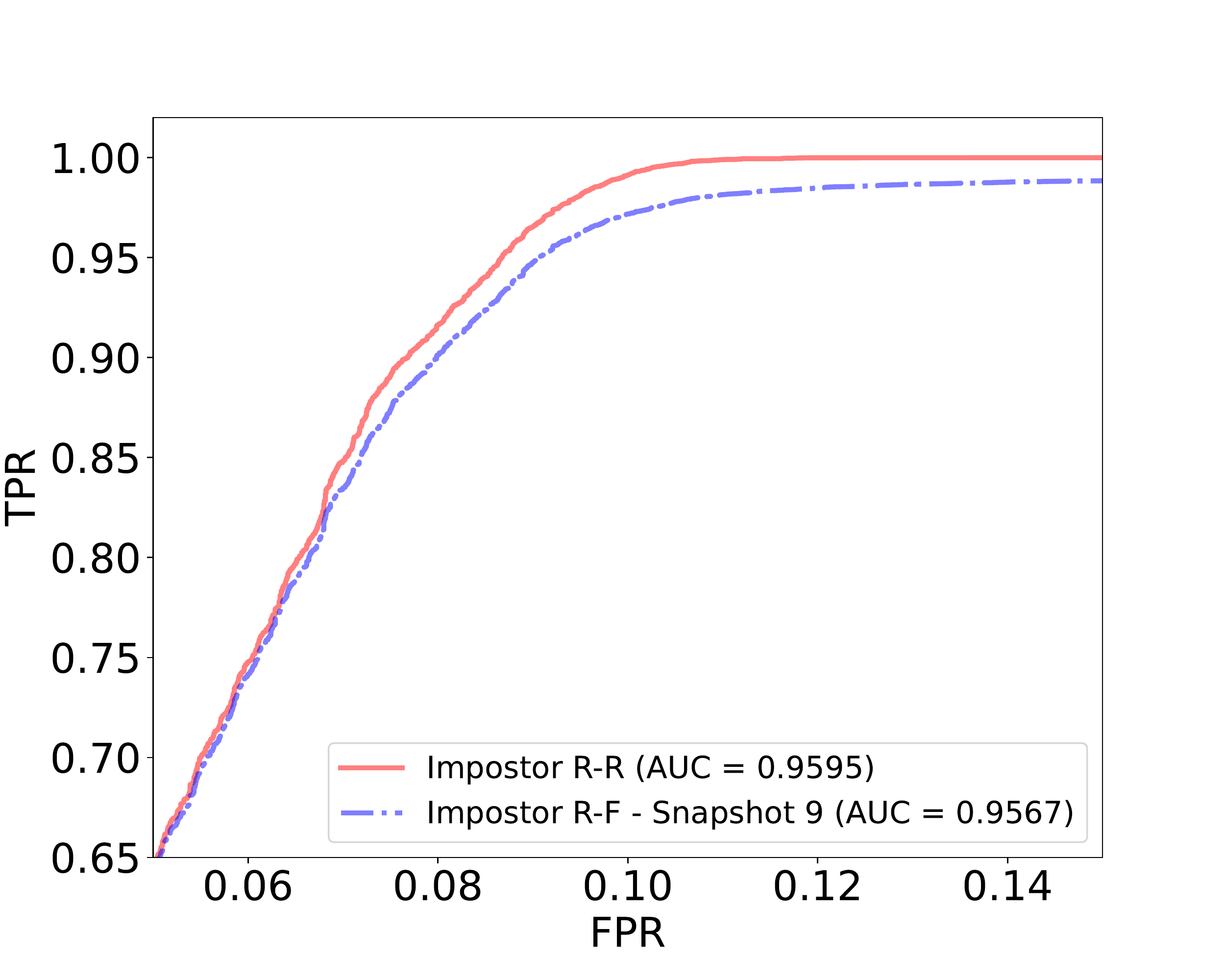}
        \caption{USIT3}
    \end{subfigure}%
    \begin{subfigure}{0.31\textwidth}
        \centering
        \includegraphics[width=\textwidth]{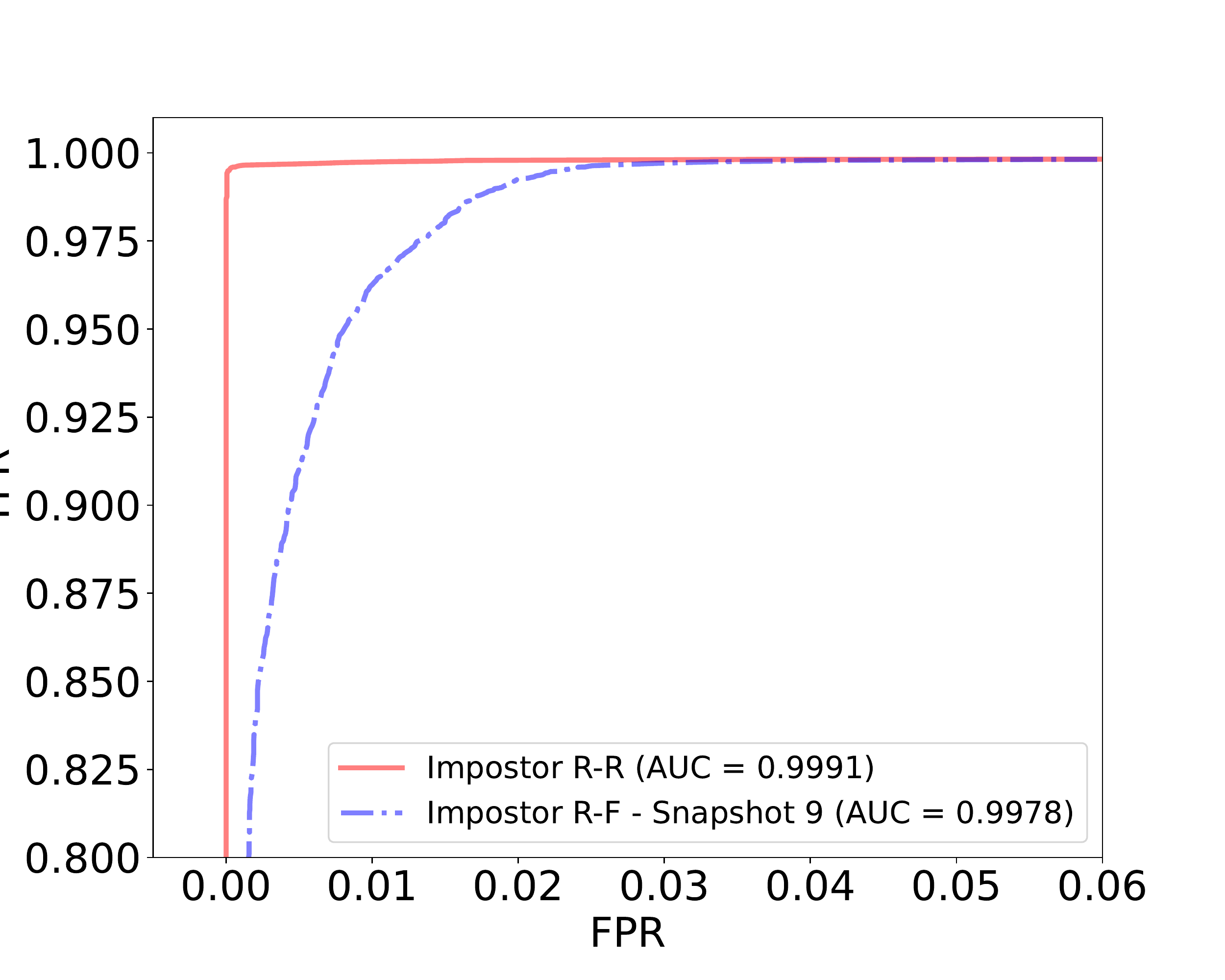}
        \caption{VeriEye}
    \end{subfigure}%
    \caption{Match score distributions for HDBSIF, USIT3, and VeriEye results for images generated at \textit{snapshot 9}.}
\end{figure*}

\begin{figure*}[h]
    \centering
    \includegraphics[width=0.9\textwidth]{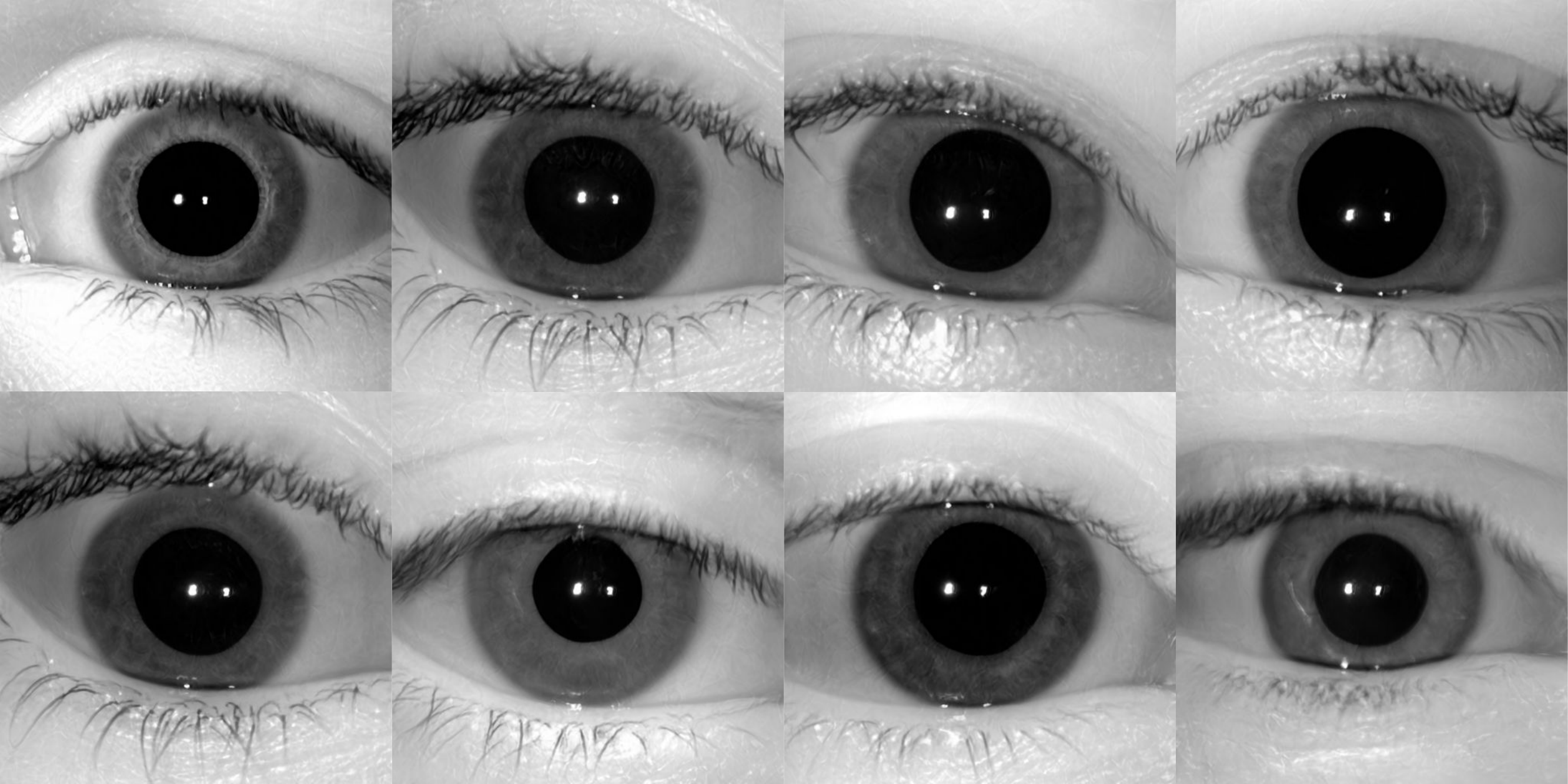}
    \caption{Image samples generated by the model at \textit{snapshot 10}.}
\end{figure*}
\begin{figure*}[h]
    \centering
    \begin{subfigure}{.31\textwidth}
        \centering
        \includegraphics[width=\textwidth]{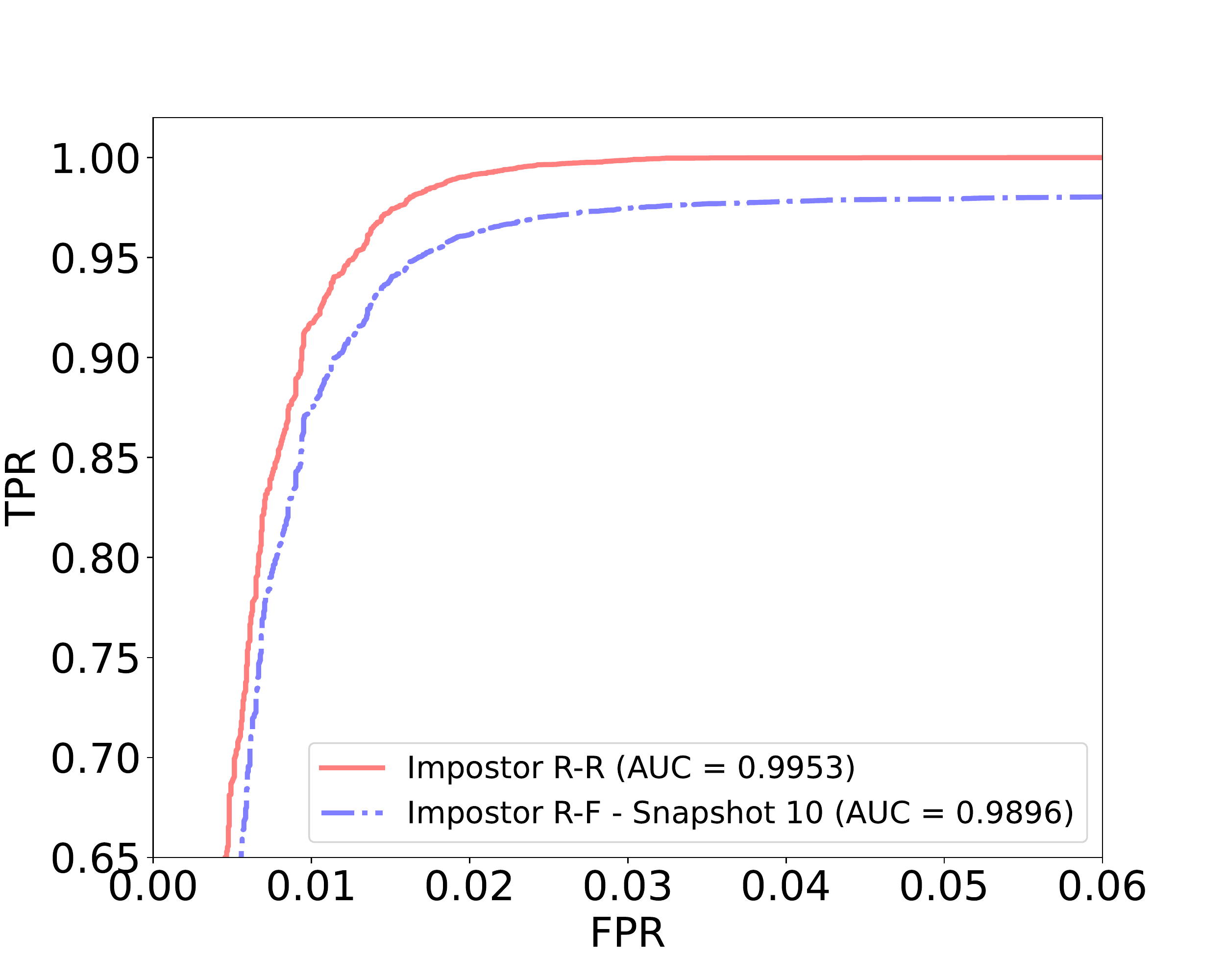}
        \caption{HDBSIF}
    \end{subfigure}%
    \begin{subfigure}{.31\textwidth}
        \centering
        \includegraphics[width=\textwidth]{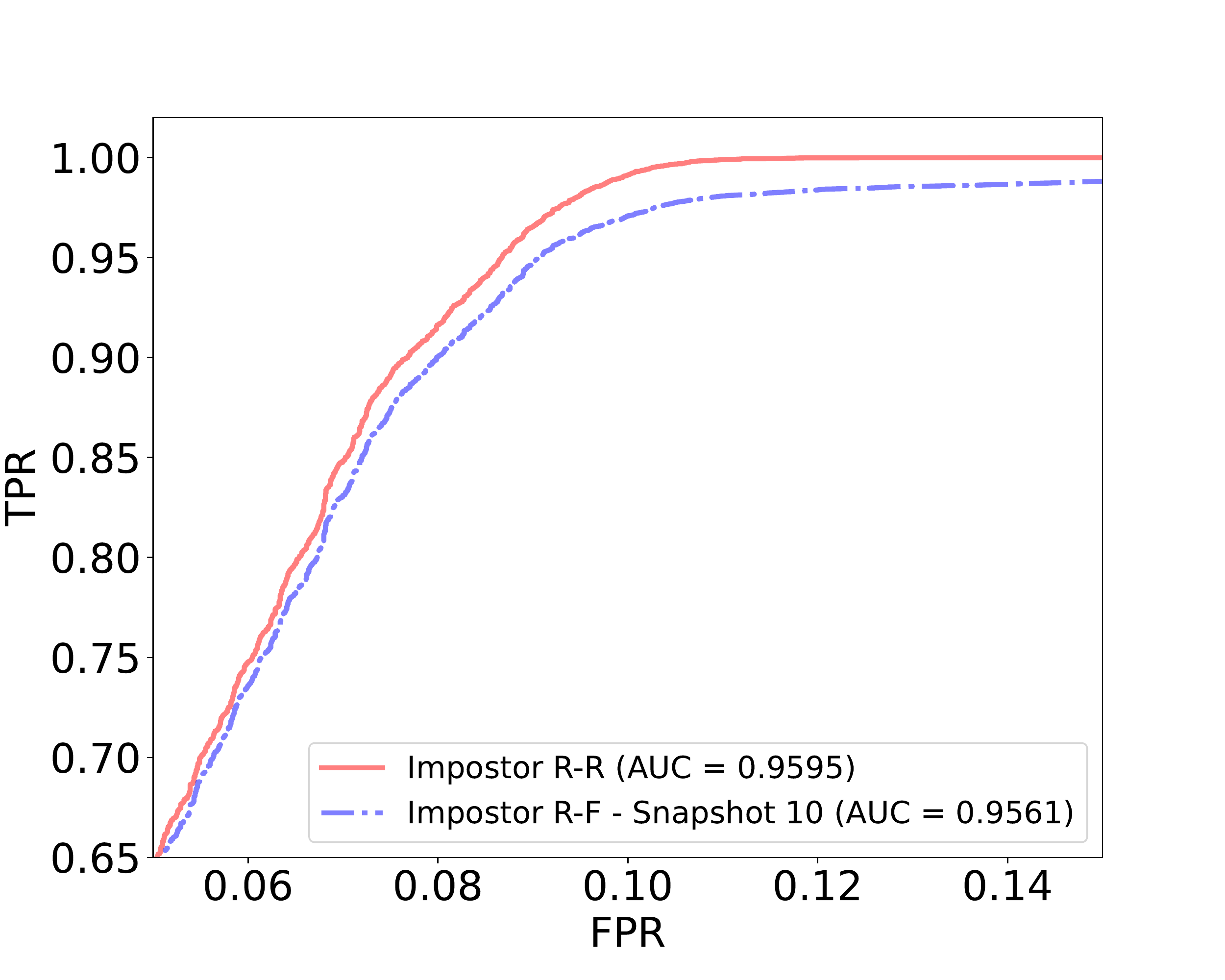}
        \caption{USIT3}
    \end{subfigure}%
    \begin{subfigure}{0.31\textwidth}
        \centering
        \includegraphics[width=\textwidth]{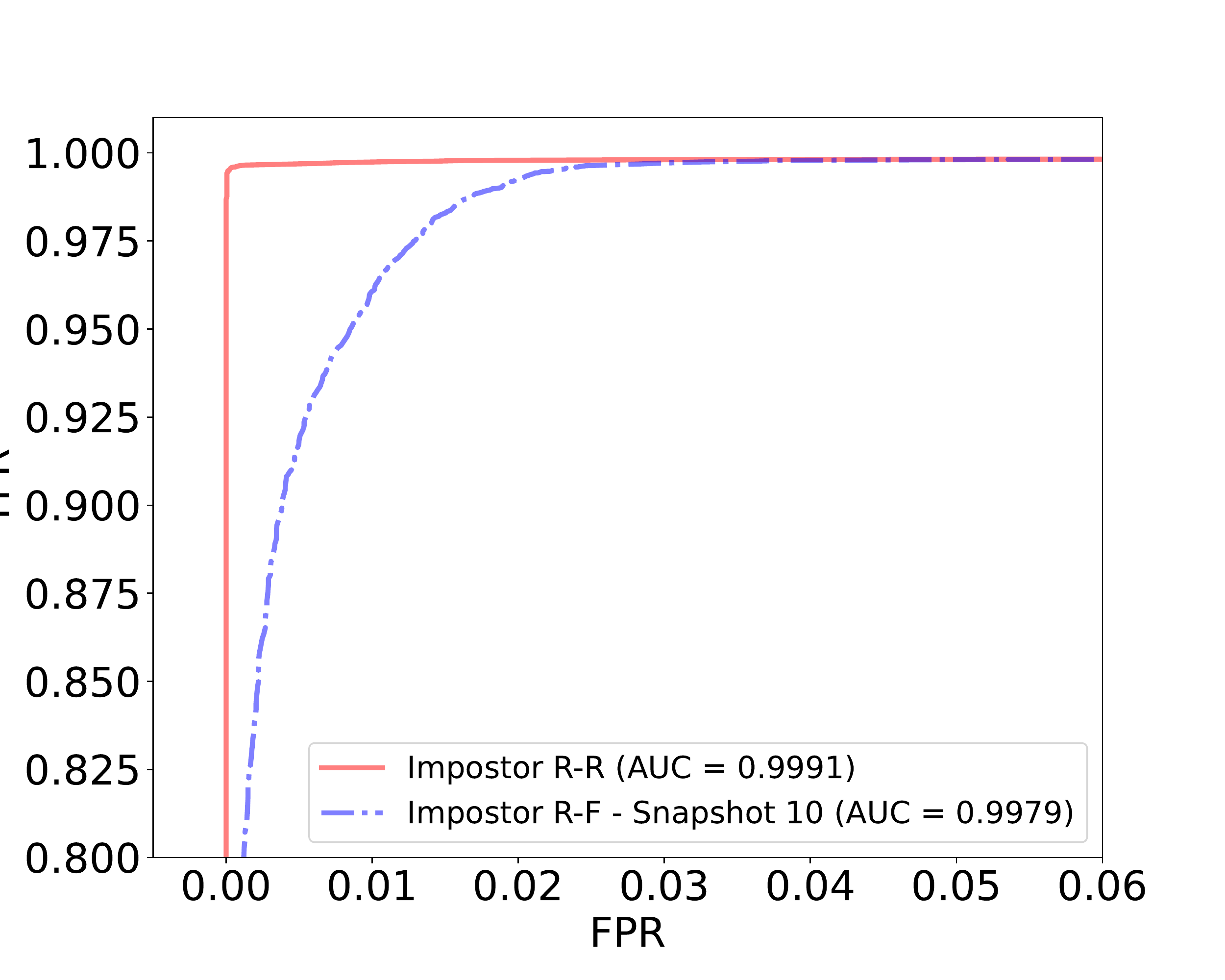}
        \caption{VeriEye}
    \end{subfigure}%
    \caption{Match score distributions for HDBSIF, USIT3, and VeriEye results for images generated at \textit{snapshot 10}.}
\end{figure*}

\begin{figure*}[h]
    \centering
    \includegraphics[width=0.9\textwidth]{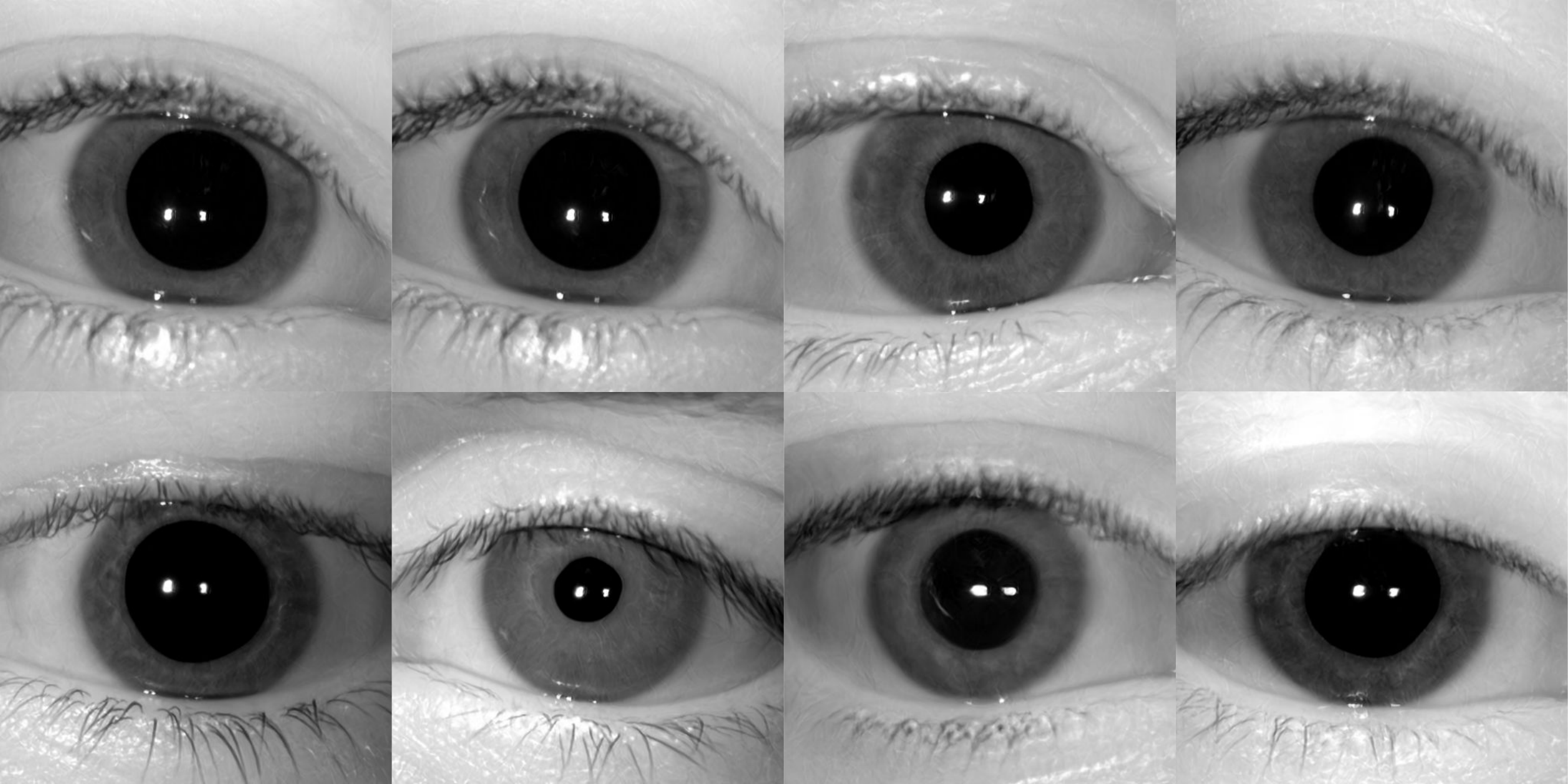}
    \caption{Image samples generated by the model at \textit{snapshot 11}.}
\end{figure*}
\begin{figure*}[h]
    \centering
    \begin{subfigure}{.31\textwidth}
        \centering
        \includegraphics[width=\textwidth]{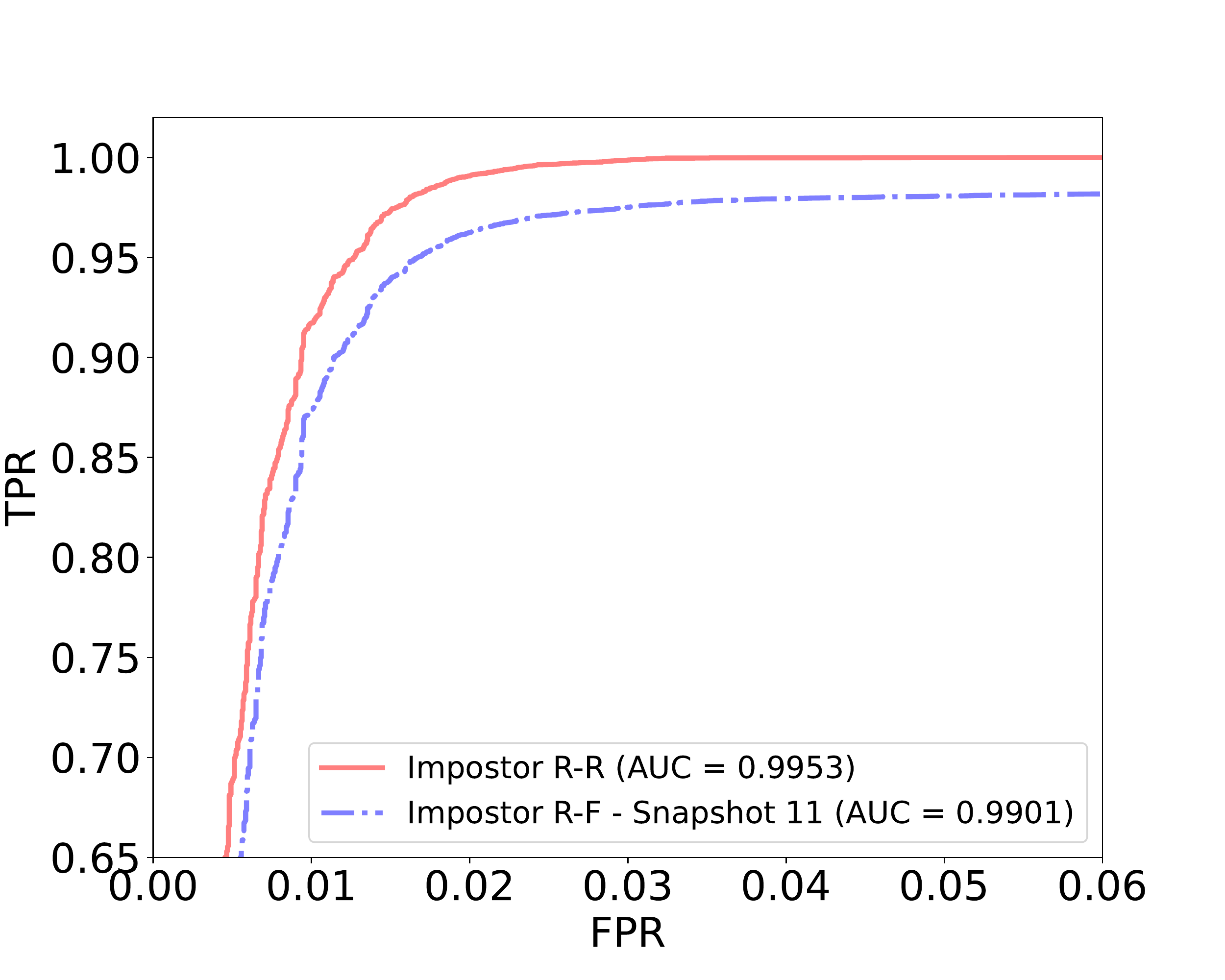}
        \caption{HDBSIF}
    \end{subfigure}%
    \begin{subfigure}{.31\textwidth}
        \centering
        \includegraphics[width=\textwidth]{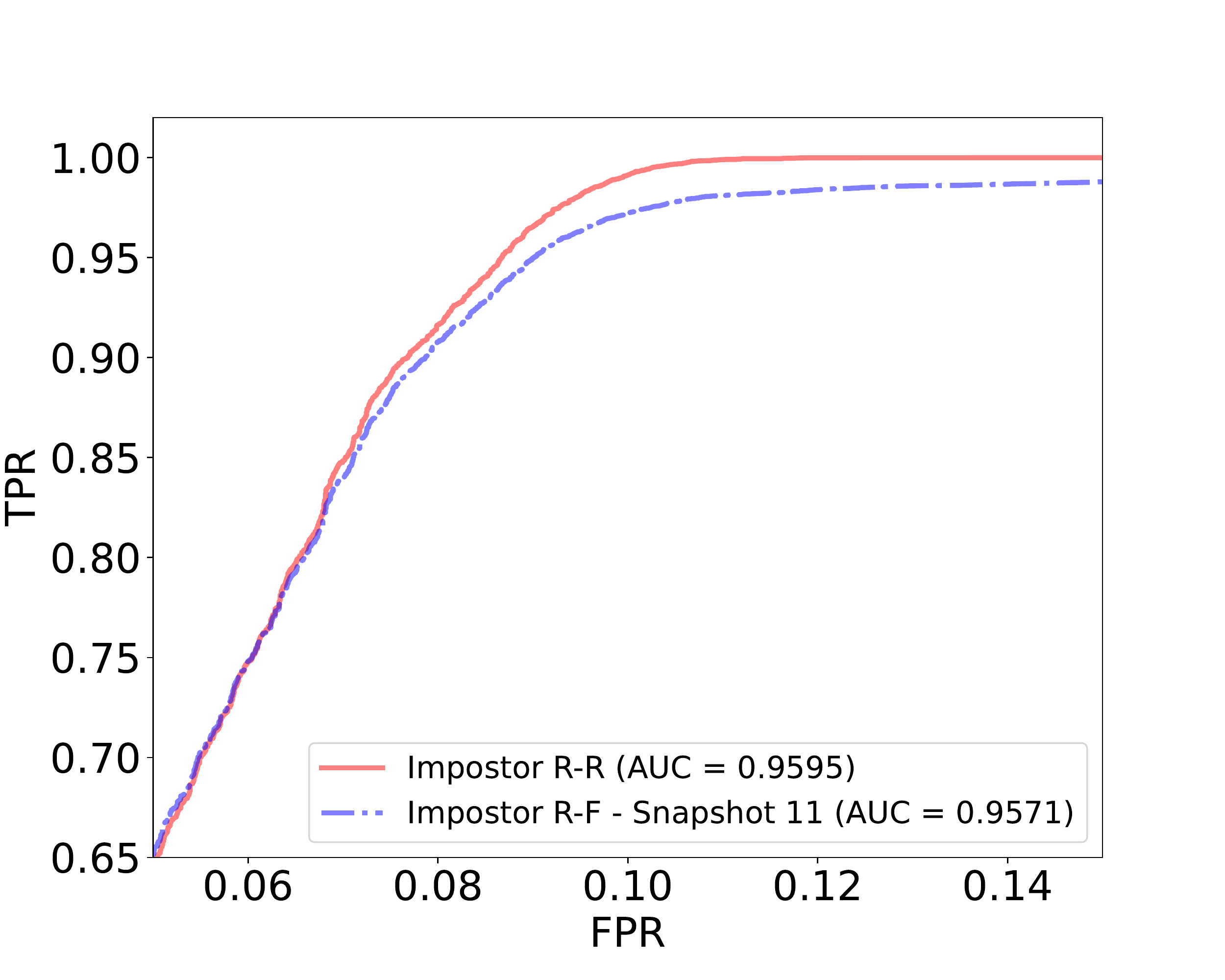}
        \caption{USIT3}
    \end{subfigure}%
    \begin{subfigure}{0.31\textwidth}
        \centering
        \includegraphics[width=\textwidth]{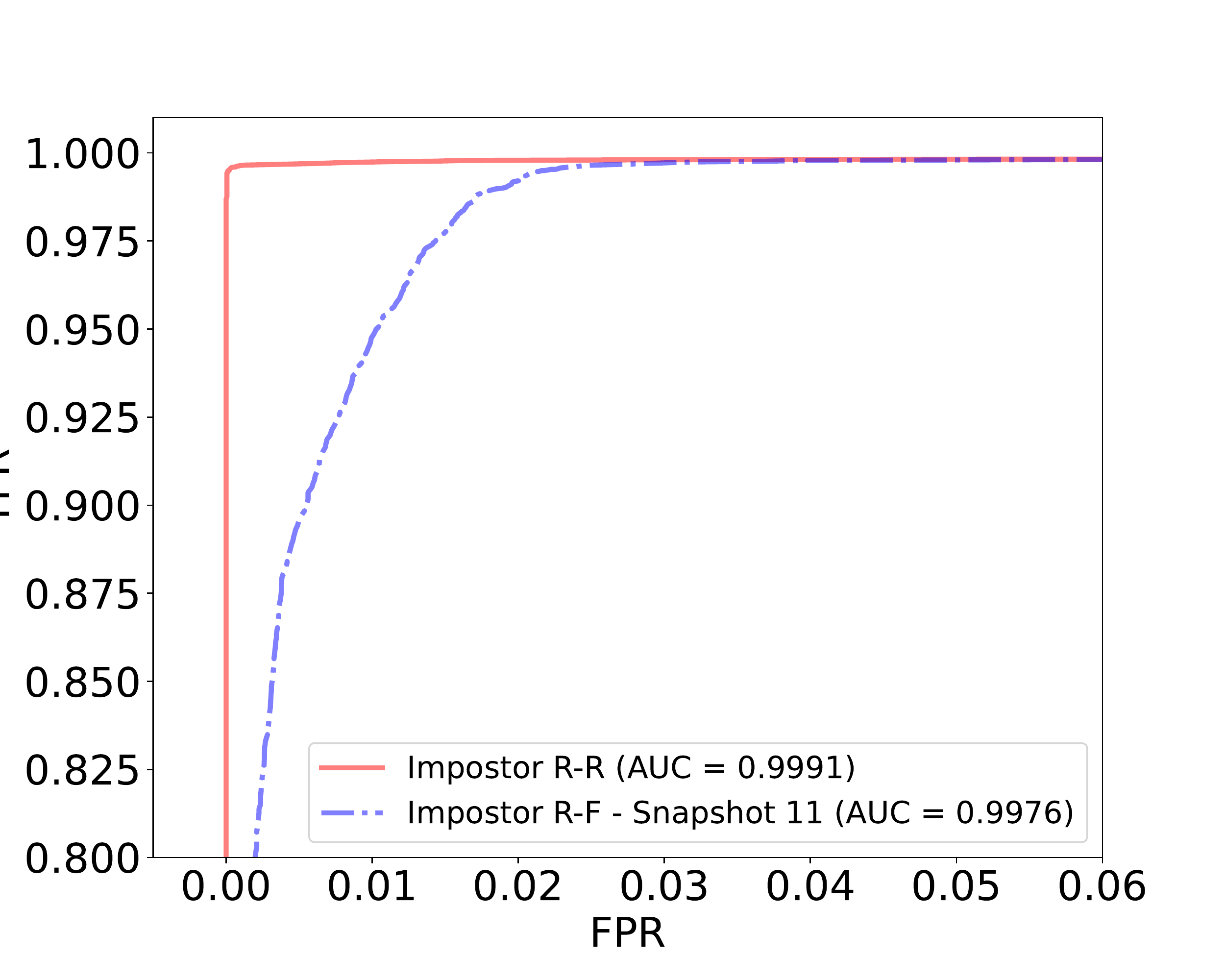}
        \caption{VeriEye}
    \end{subfigure}%
    \caption{Match score distributions for HDBSIF, USIT3, and VeriEye results for images generated at \textit{snapshot 11}.}
\end{figure*}

\begin{figure*}[h]
    \centering
    \includegraphics[width=0.9\textwidth]{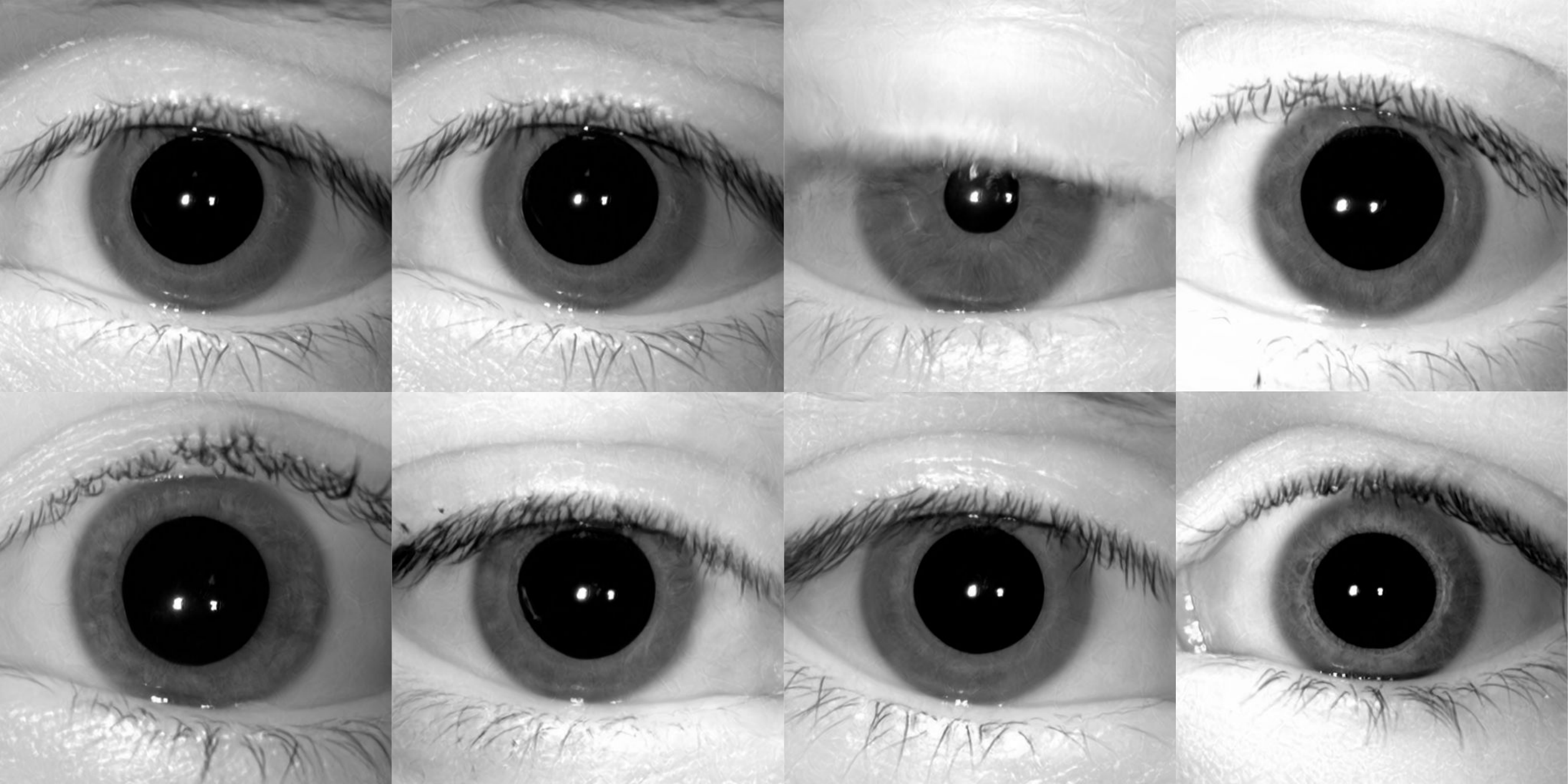}
    \caption{Image samples generated by the model at \textit{snapshot 12}.}
\end{figure*}
\begin{figure*}[h]
    \centering
    \begin{subfigure}{.31\textwidth}
        \centering
        \includegraphics[width=\textwidth]{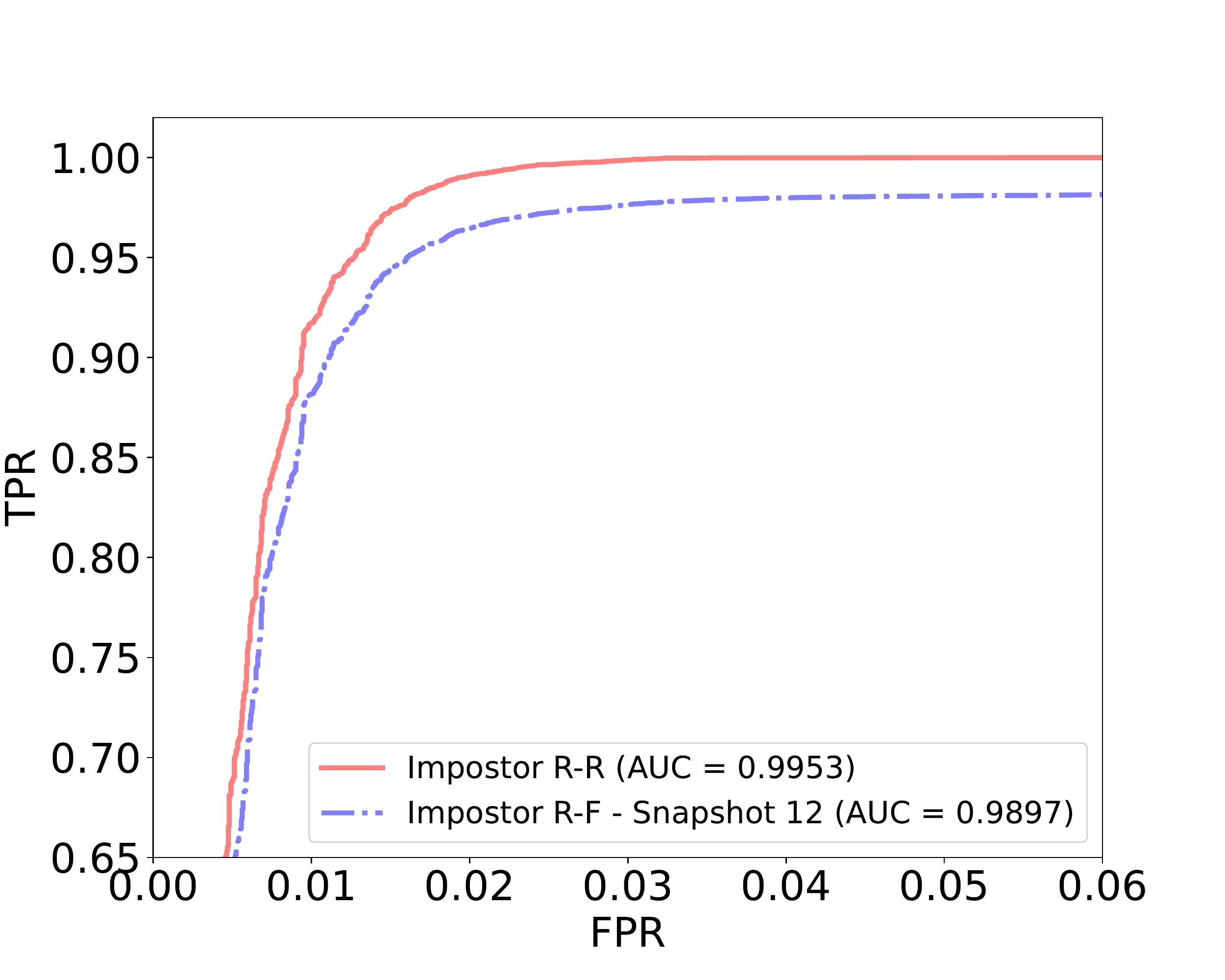}
        \caption{HDBSIF}
    \end{subfigure}%
    \begin{subfigure}{.31\textwidth}
        \centering
        \includegraphics[width=\textwidth]{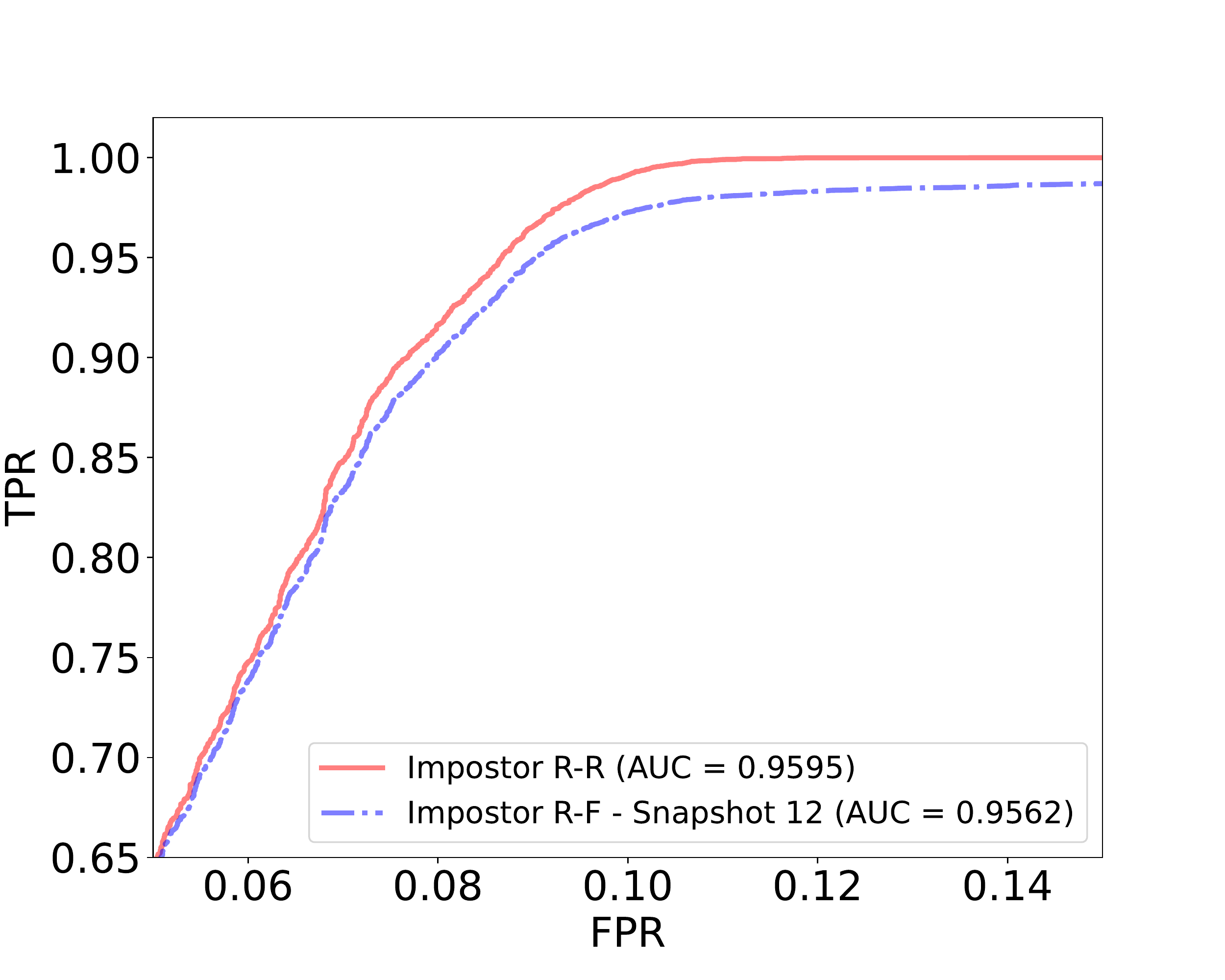}
        \caption{USIT3}
    \end{subfigure}%
    \begin{subfigure}{0.31\textwidth}
        \centering
        \includegraphics[width=\textwidth]{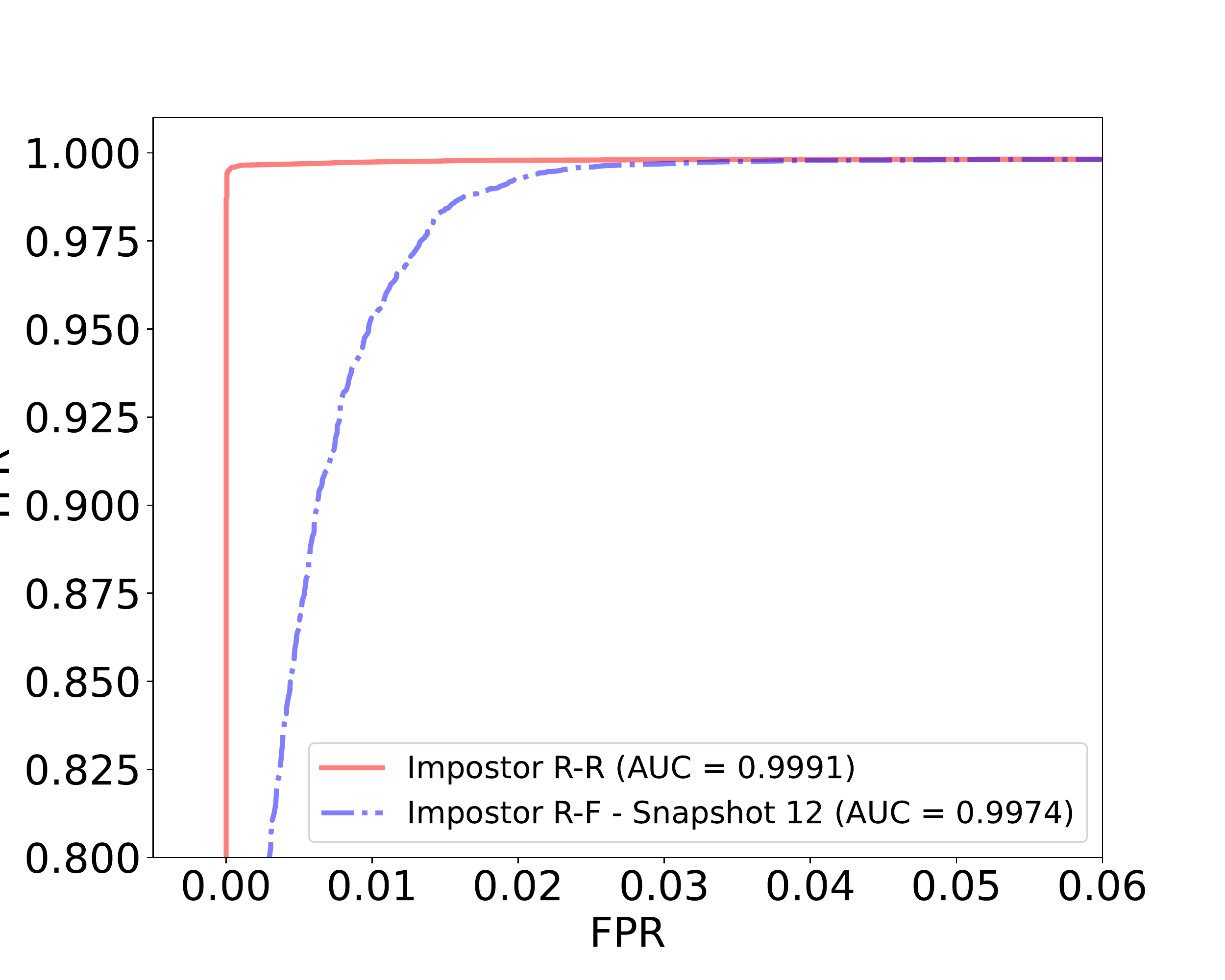}
        \caption{VeriEye}
    \end{subfigure}%
    \caption{Match score distributions for HDBSIF, USIT3, and VeriEye results for images generated at \textit{snapshot 12}.}
\end{figure*}

\begin{figure*}[h]
    \centering
    \includegraphics[width=0.9\textwidth]{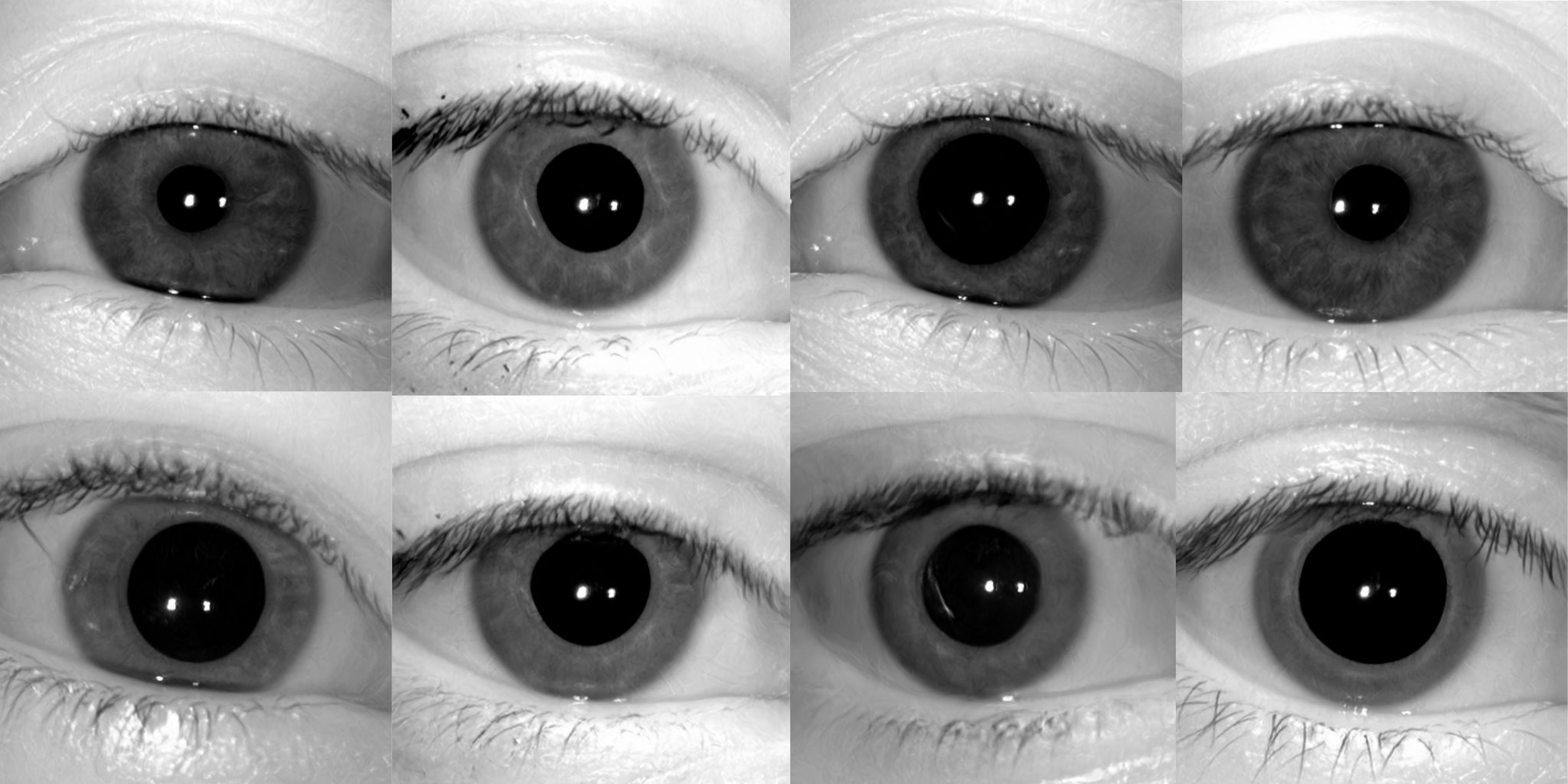}
    \caption{Image samples generated by the model at \textit{snapshot 13}.}
\end{figure*}
\begin{figure*}[h]
    \centering
    \begin{subfigure}{.31\textwidth}
        \centering
        \includegraphics[width=\textwidth]{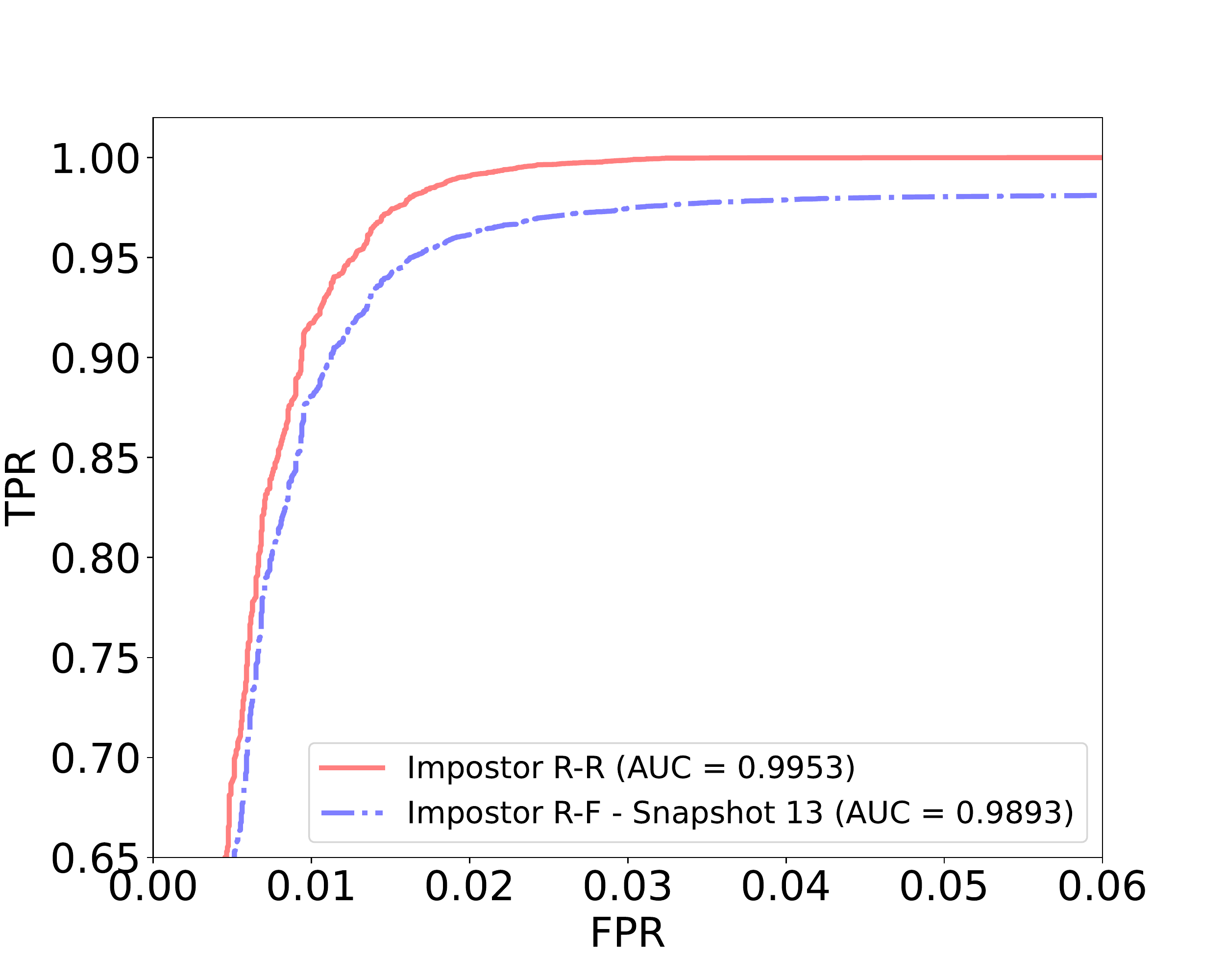}
        \caption{HDBSIF}
    \end{subfigure}%
    \begin{subfigure}{.31\textwidth}
        \centering
        \includegraphics[width=\textwidth]{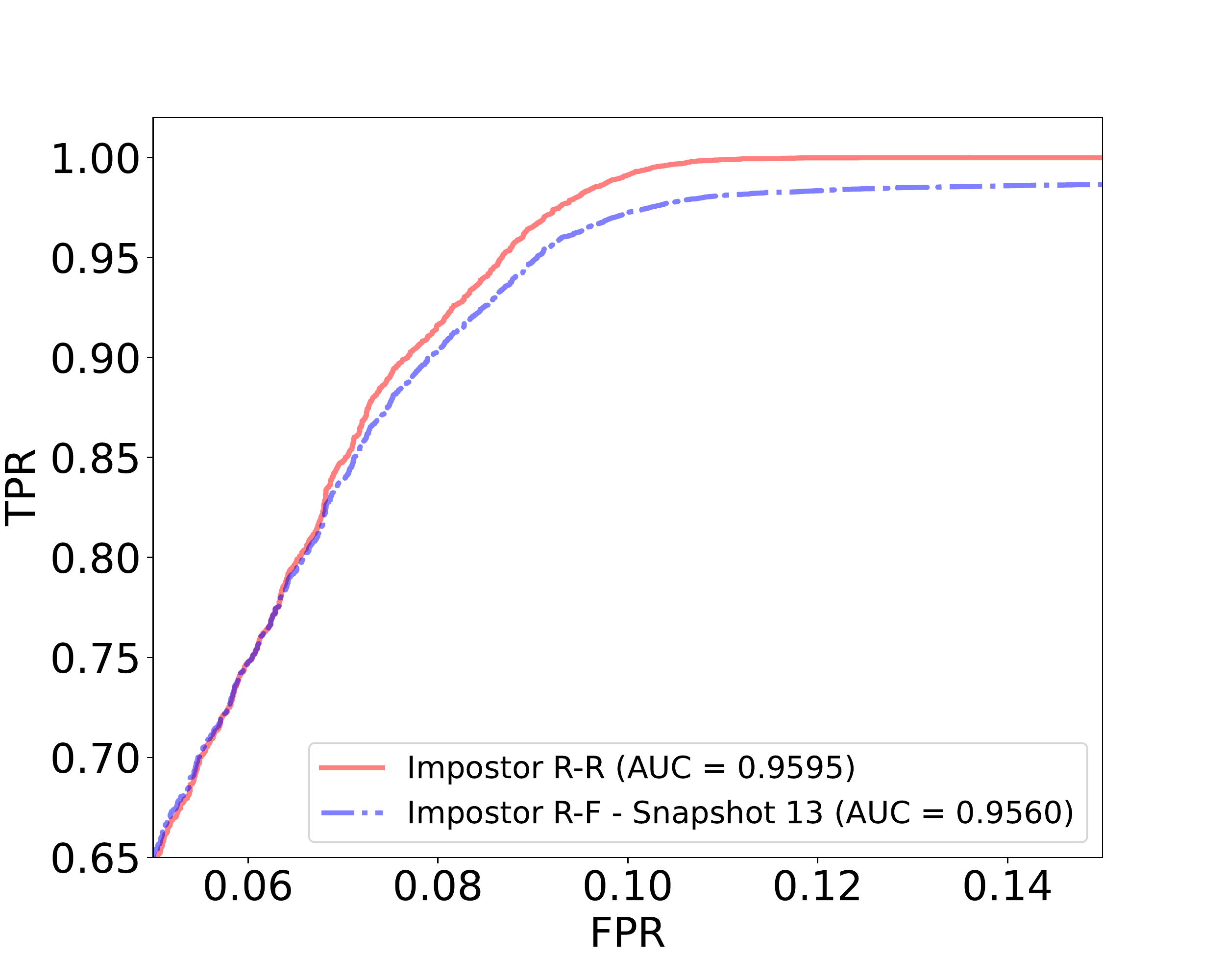}
        \caption{USIT3}
    \end{subfigure}%
    \begin{subfigure}{0.31\textwidth}
        \centering
        \includegraphics[width=\textwidth]{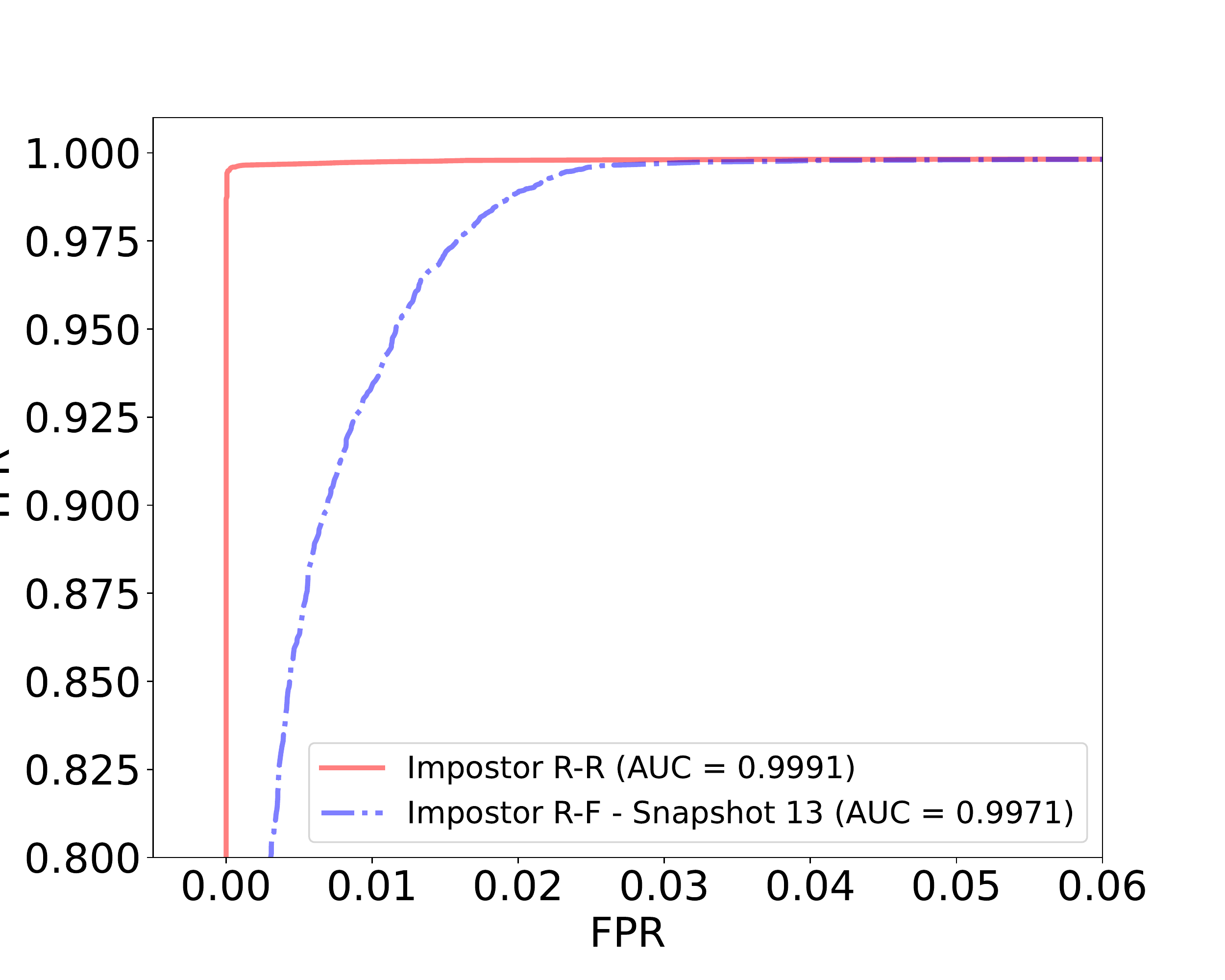}
        \caption{VeriEye}
    \end{subfigure}%
    \caption{Match score distributions for HDBSIF, USIT3, and VeriEye results for images generated at \textit{snapshot 13}.}
\end{figure*}

\begin{figure*}[h]
    \centering
    \includegraphics[width=0.9\textwidth]{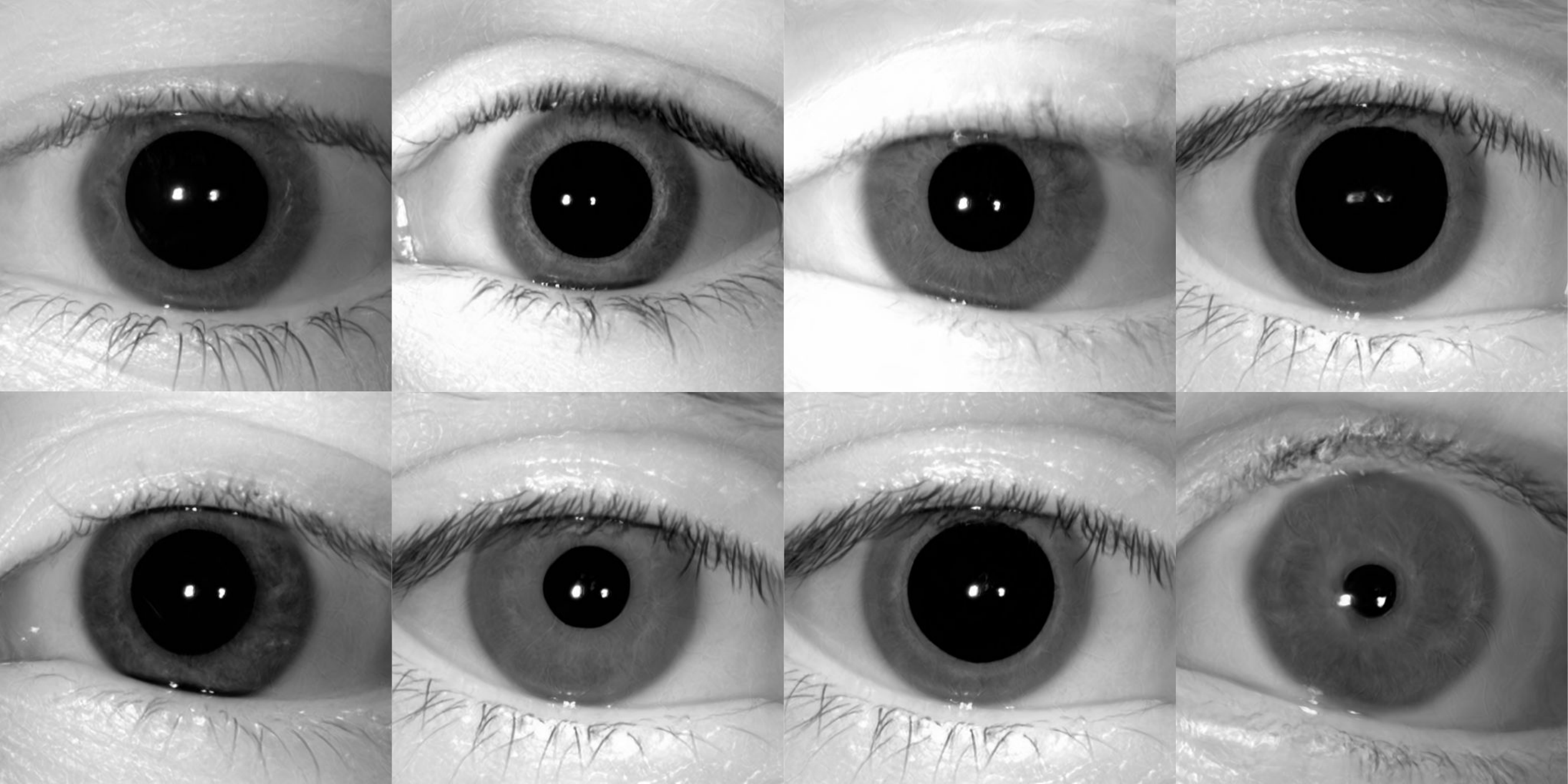}
    \caption{Image samples generated by the model at \textit{snapshot 14}.}
\end{figure*}
\begin{figure*}[h]
    \centering
    \begin{subfigure}{.31\textwidth}
        \centering
        \includegraphics[width=\textwidth]{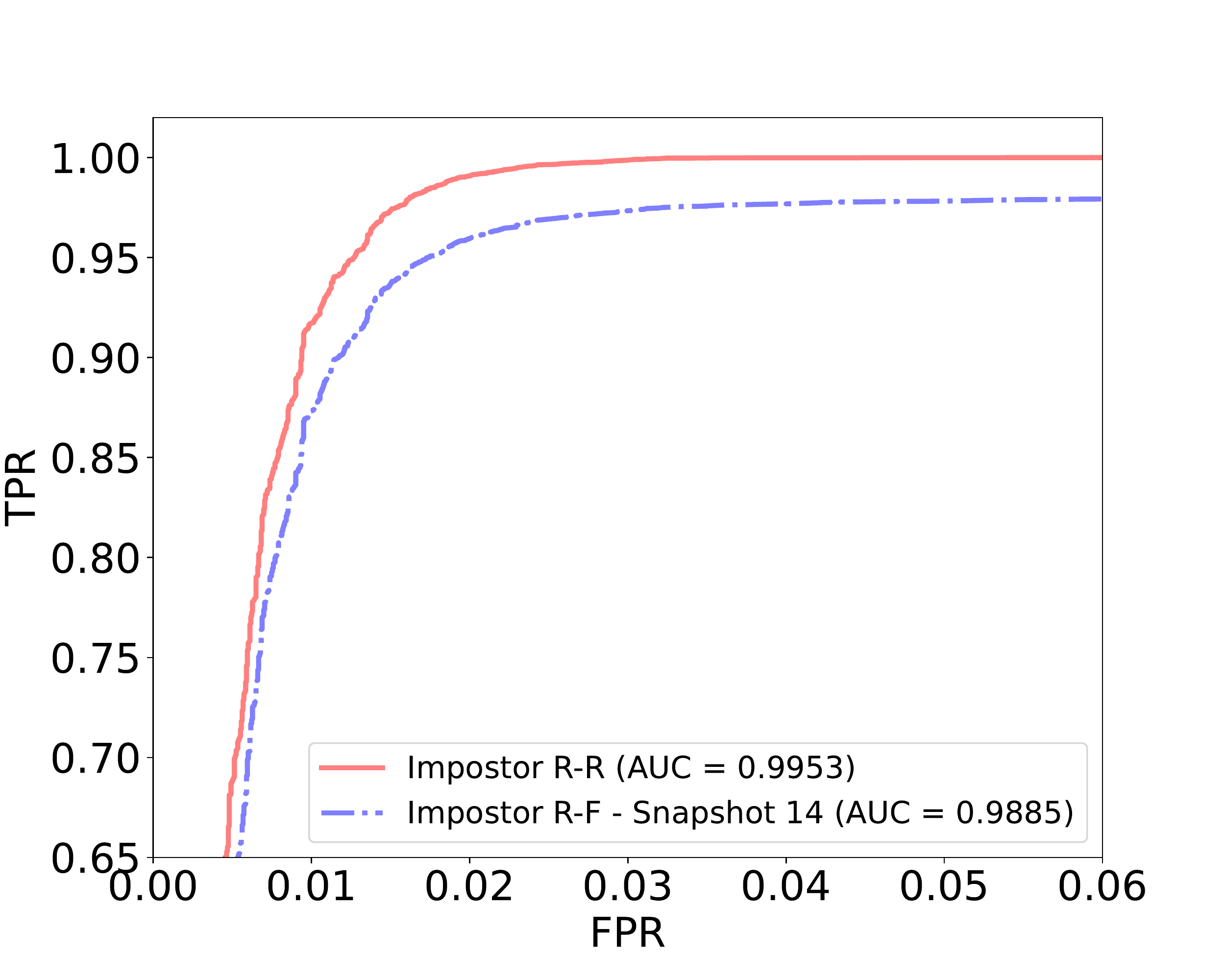}
        \caption{HDBSIF}
    \end{subfigure}%
    \begin{subfigure}{.31\textwidth}
        \centering
        \includegraphics[width=\textwidth]{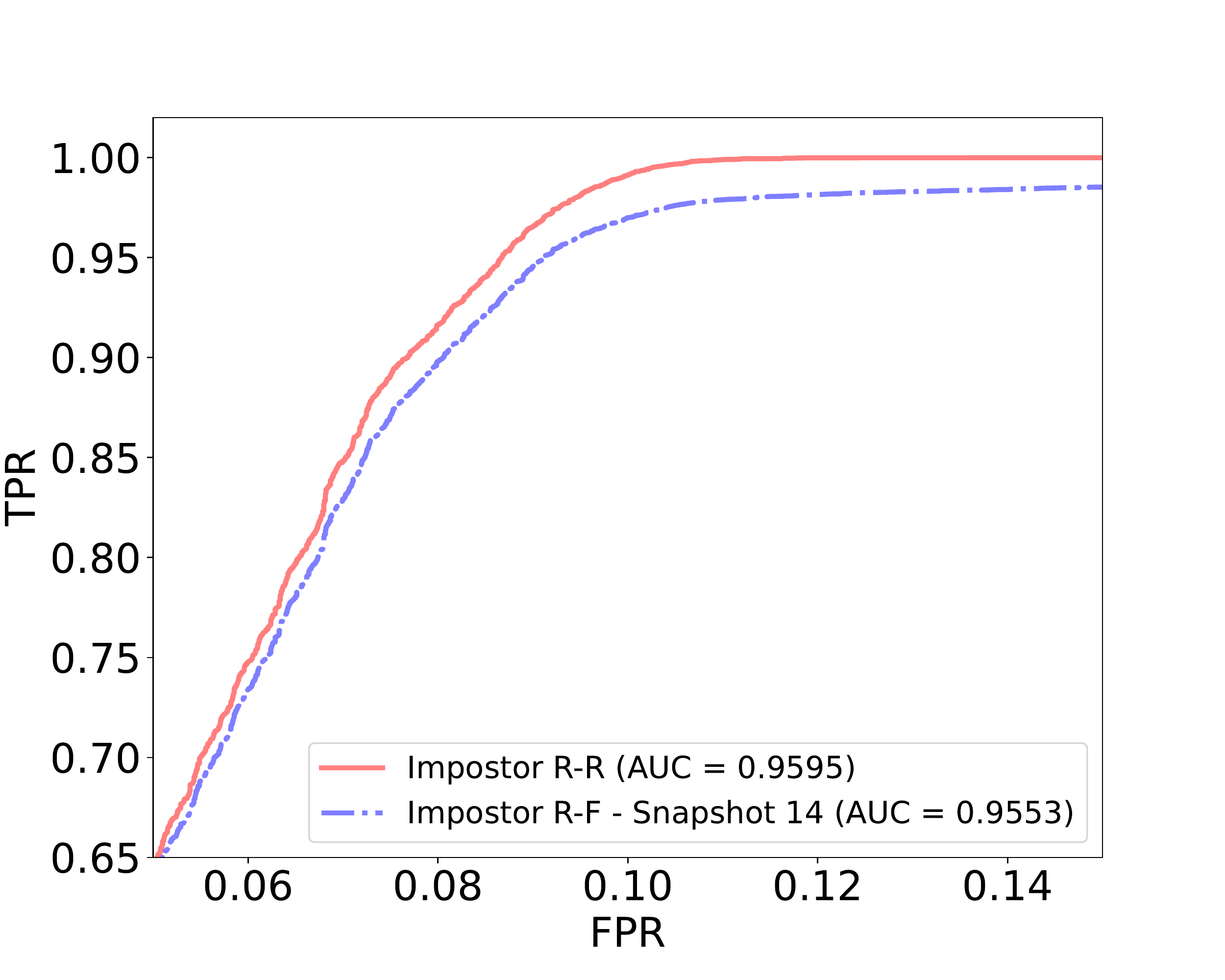}
        \caption{USIT3}
    \end{subfigure}%
    \begin{subfigure}{0.31\textwidth}
        \centering
        \includegraphics[width=\textwidth]{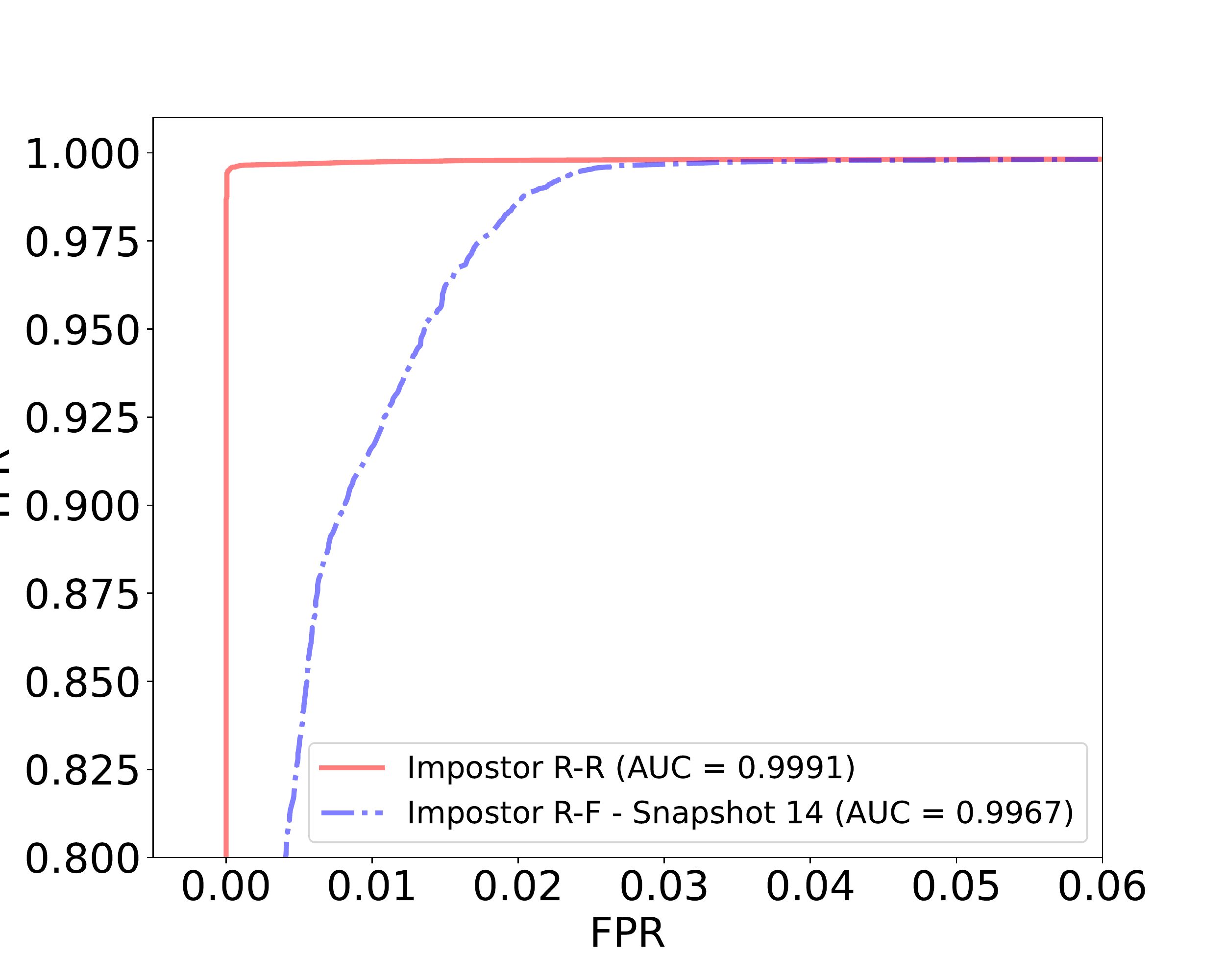}
        \caption{VeriEye}
    \end{subfigure}%
    \caption{Match score distributions for HDBSIF, USIT3, and VeriEye results for images generated at \textit{snapshot 14}.}
\end{figure*}

\end{document}